\pdfoutput=1
\documentclass[letterpaper, 10 pt, journal, twoside]{ieeetran}
\let\labelindent\relax
\usepackage[numbers]{natbib}
\usepackage{url}
\usepackage{enumitem}
\usepackage[font={small}]{caption}
\usepackage{subcaption}
\usepackage{textcomp}
\usepackage{mathtools, nccmath}
\usepackage{graphicx}
\usepackage{amsfonts}
\usepackage{amsmath}
\usepackage{amssymb}
\usepackage{algorithm}
\usepackage{algorithmic}
\usepackage{tikz}
\usepackage{arydshln}
\usepackage{multirow}
\usepackage{bm}
\usepackage{epstopdf}
\usepackage{balance}
\usepackage{etoolbox}

\newtheorem{remark}{Remark}

\newcommand*\circled[1]{\tikz[baseline=(char.base)]{
            \node[shape=circle,draw,inner sep=0.25pt] (char) {#1};}}

\newcommand{\insertfig}{%
  \makebox[0pt]{\includegraphics[width=\linewidth]{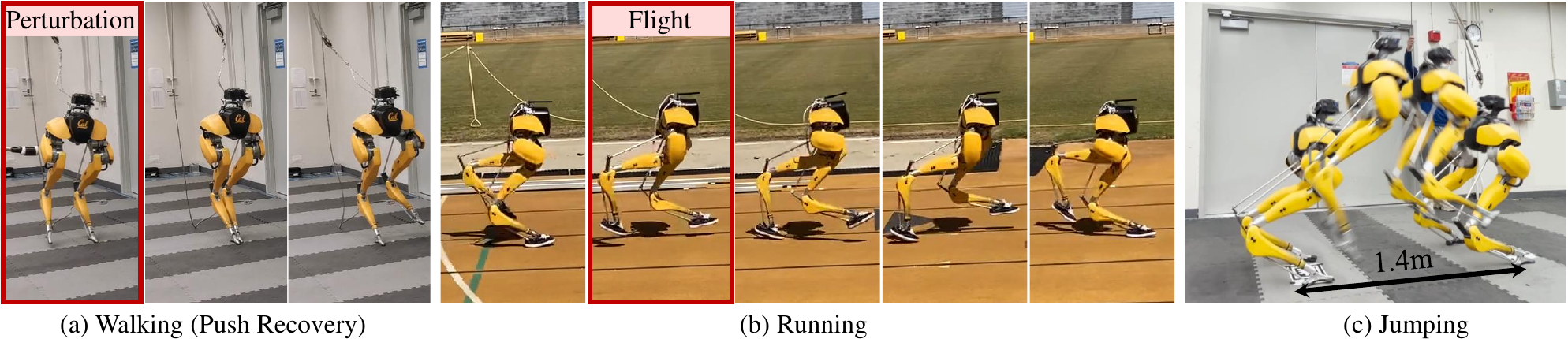}\label{fig:intro}} \\
  \small{Fig.~1: Cassie, a torque-controlled human-sized bipedal robot, performs various locomotion skills by using controllers developed through our framework. We present a unified RL framework that is able to train robust and agile controllers for a diverse range of highly dynamic skills, such as (a) walking, (b) running, and (c) jumping. The control policies developed by the proposed dual-history policy architecture and training system can adapt to changes in the robot dynamics, thereby enabling the direct transfer of learned policies to a real robot after training only in simulation without further real-world tuning. More experimental results can be best seen in the supplementary videos which are summarized in Table~I.
  }
}

\makeatletter
\apptocmd{\@maketitle}{~~~~~~~~~~~~~~~~~~~~~~~~~~~~~~~~~~~~~~~~~~~~~~~~~~~~~~~~~~~~~~~~~~~~~~~~~~\insertfig}{}{} 
\makeatother
\makeatletter
\newcommand\notsotiny{\@setfontsize\notsotiny\@vipt\@viiipt}
\makeatother

\begin{document}
\title{Reinforcement Learning for Versatile, Dynamic, and Robust Bipedal Locomotion Control}
\author{Zhongyu Li$^1$, Xue Bin Peng$^2$, Pieter Abbeel$^1$, Sergey Levine$^1$, Glen Berseth$^{3,4}$, Koushil Sreenath$^1$\\
\normalsize{$^1$University of California Berkeley~~~~$^2$Simon Fraser University~~~~$^3$Universit\'e de Montr\'eal~~~~$^4$Mila – Quebec AI Institute}
\thanks{Email: zhongyu\_li@berkeley.edu; koushils@berkeley.edu}
}

\maketitle
\begin{abstract}
This paper presents a comprehensive study on using deep reinforcement learning (RL) to create dynamic locomotion controllers for bipedal robots. 
Going beyond focusing on a single locomotion skill, we develop a general control solution that can be used for a range of dynamic bipedal skills, from periodic walking and running to aperiodic jumping and standing.
Our RL-based controller incorporates a novel dual-history architecture, utilizing both a long-term and short-term input/output (I/O) history of the robot. 
This control architecture, when trained through the proposed end-to-end RL approach, consistently outperforms other methods across a diverse range of skills in both simulation and the real world.
The study also delves into the adaptivity and robustness introduced by the proposed RL system in developing locomotion controllers. 
We demonstrate that the proposed architecture can adapt to both time-invariant dynamics shifts and time-variant changes, such as contact events, by effectively using the robot's I/O history. 
Additionally, we identify task randomization as another key source of robustness, fostering better task generalization and compliance to disturbances.
The resulting control policies can be successfully deployed on Cassie, a torque-controlled human-sized bipedal robot.
This work pushes the limits of agility for bipedal robots through extensive real-world experiments. We demonstrate a diverse range of locomotion skills, including: robust standing, versatile walking, fast running with a demonstration of a 400-meter dash, and a diverse set of jumping skills, such as standing long jumps and high jumps.
\end{abstract}
\section{Introduction}\label{sec:intro}
Human environments are diverse and predominantly tailored for bipedal locomotion, and therefore the overarching goal in the bipedal robot community has been to develop robots capable of reliably operating within these environments.
This paper aims to address one of the bottlenecks in achieving this objective: developing a solution for the control of diverse, agile, and robust legged locomotion skills, such as walking, running, and jumping for high-dimensional human-sized bipedal robots. 

Although research on bipedal robot locomotion has been ongoing for decades (e.g., \cite{Raibert-1984-15614}), developing a general framework capable of achieving robust control for diverse locomotion skills remains an open problem. 
The challenges arise from the complexity of the underactuated dynamics of bipedal robots and the distinct contact plan associated with each locomotion skill.  
First, given the floating base and resulting underactuated dynamics, bipedal robots rely on contacts with the environment in order to move. 
The consistent (and hard-to-model) contacts lead to discontinuities in the trajectories, requiring contact mode planning and stabilization during mode transitions.
However, due to the high dimensionality and nonlinearity of bipedal robots, leveraging its full-order dynamics model for motion planning and control is computationally expensive and intractable for online applications.
Second, the diverse nature of bipedal locomotion skills, whether periodic or aperiodic, presents significant challenges to the development of a simple and general framework.
For instance, running, unlike walking, introduces more complexity due to a repeated flight phase where the robot is underactuated. 
For periodic skills such as walking or running, we can achieve \textit{orbital stability} \citep{westervelt2003hybrid}, by allowing for small corrections over multiple gait cycles. 
However, aperiodic motions, such as jumping, lack this inherent stability, which again poses additional challenges due to the requirement of finite-time stability~\citep{goswami2009planar}, which is further compounded by a large impact force upon landing. 

In this work, we address the aforementioned challenges by leveraging reinforcement learning (RL) to create controllers for robots that feature high-dimensional, nonlinear dynamics in the real world. 
These controllers can leverage the robot's proprioceptive information to adapt to the robot's uncertain dynamics, which may be potentially time-varying due to wear and tear.
These controllers are able to generalize to new environments and settings, exhibiting robust behaviors to unexpected scenarios by utilizing the agility of bipedal robots.
Furthermore, our framework provides a general recipe for reproducing a variety of bipedal locomotion skills.

\subsection{Objective of this Work}
The high-dimensionality and nonlinearity of a torque-controlled human-sized bipedal robot may at first seem like a daunting obstacle for developing effective controllers.
However, these characteristics can also be advantageous by enabling complex agile maneuvers through the robot's high-dimensional dynamics.
Our objective is to develop a general control framework for such bipedal robots to unlock their full potential. 
The goal of this framework is to enable a range of dynamic bipedal locomotion skills in the real world with a limited prescription on the resulting maneuvers. The skills we consider are shown in Fig.~\ref{fig:intro}, which include robust standing, walking, running, and jumping. These skills can also be used to perform a diverse repertoire of tasks, including walking at various velocities and heights, running at different speeds and directions, and jumping to various targets, all while maintaining robustness during real-world deployment.
To achieve this, we leverage model-free RL, which allows the robot to learn through trial-and-error on the system's full-order dynamics.
In addition to real-world experiments, we also provide an in-depth analysis of the benefits of using RL for legged locomotion control and offer a detailed study into how to effectively structure the learning process to harness these advantages, such as adaptivity and robustness.

\subsection{Terminology}
First, we will establish the terminology we will use to describe various aspects of \emph{legged locomotion} in this paper. 
The term \textit{skill} is used to characterize a particular type of locomotion, which includes behaviors such as walking, running, and jumping. 
The robot can then leverage those skills to perform various \textit{tasks}, which are defined by a given goal. 
For example, these tasks can include following different target velocities while walking, turning at different angles while running, or leaping towards different target locations while jumping. We use the term \textit{versatile policy} to describe a control policy that can accomplish various tasks using one or more locomotion skills.

\subsection{Contributions}
This work advances the field of legged locomotion control for bipedal robots, with the following key contributions:

\textbf{Development of a new framework for general bipedal locomotion control:} 
We introduce a general RL framework for bipedal robots that is effective across a wide range of locomotion skills, spanning periodic skills such as walking and running, aperiodic skills such as jumping, and stationary skills such as standing. The resulting controllers can be directly deployed on a real robot without any additional tuning or training on the physical system.

\textbf{Novel design choices for RL-based control policy:} We present a new dual-history architecture for non-recurrent RL policies, which integrates both the long and short input/output (I/O) history with explicit length of the robot, for RL-based control. When combined with the proposed training strategy that trains the base policy with the short history jointly with the long history encoder, this architecture demonstrates state-of-the-art performance in learning dynamic bipedal locomotion control, offering consistent benefits across various locomotion skills, which are validated in both simulation and real world experiments, as detailed in Sec.~\ref{sec:policy_structure} and discussed in Sec.~\ref{subsec:dual_history_advantages}.

\textbf{Empirically investigating adaptivity in RL controllers:} In the control theory community, there have been notable efforts in bridging adaptive control and RL, like~\cite{annaswamy2023adaptive}. In this work, we conduct a detailed empirical study to investigate the adaptivity of the control policy developed through RL. We show that the adaptivity from RL includes not only time-invariant shifts in the dynamics but also time-variant changes like contact events. We validated this not only in simulation, as described in Sec.~\ref{subsec:adaptivity}, but also by several real-world (zero-shot-transferred) experiments such as in-place walking with minimal drift and targeted jumping using a bipedal robot as detailed in Sec.~\ref{sec:experiments}.

\textbf{Improving robustness in RL controllers:} Our study introduces a new dimension of robustness in RL-based control policies. Beyond the commonly-used dynamics randomization in robotics~\citep{peng2018sim}, we demonstrate that \emph{task randomization}, which trains the policy on a wide range of tasks, significantly enhances robustness by enabling task generalization. This approach, distinct from dynamics randomization, provides the robot with disturbance compliance, which is demonstrated in both simulation and real-world experiments in Sec.~\ref{sec:multi_skill}.

\textbf{Extensive real-world validation and demonstrations of novel bipedal locomotion capabilities:}
Our system is able to reproduce a wide variety of locomotion skills using Cassie, a human-sized bipedal robot, in the real world as detailed in Sec.~\ref{sec:experiments}. Cassie can track varying commands with negligible tracking errors and significant robustness to unexpected disturbances, including walking (Figs.~\ref{fig:variable_cmd},~\ref{subfig:push_recovery_rl}), running (Figs.~\ref{fig:400run},~\ref{fig:run_perturb}), and jumping (Figs.~\ref{fig:jump_snapshot_timeline},~\ref{fig:jump_perturb}). Additionally, we demonstrate novel capabilities for bipedal robots, such as robust standing recovery using different skills (Fig.~\ref{fig:robust_standing}), robust walking (Figs.~\ref{fig:walking_terrain},~\ref{fig:walking_compliance}) with control performance being consistent over a long time frame (Fig.~\ref{fig:variable_cmd}), completing a 400-meter dash using a running controller (Fig.~\ref{fig:400run}), and performing diverse bipedal jumps (Figs.~\ref{fig:jump_snapshot_timeline},~\ref{fig:various_jump}) including standing long and high jumps (Fig.~\ref{fig:jump_snapshot_timeline}). The
experiments are shown in the supplementary videos provided in Table~\ref{tab:video_list}.

This paper builds on our preliminary work presented at Robotics: Science and System~\citep{li2023robust}, which focused on bipedal jumping. We expanded our RL framework to encompass a broader range of skills, proving its applicability to both aperiodic skills like jumping and periodic skills such as walking and running. The effectiveness of this framework is consistently demonstrated across these skills through extensive real-world experiments. Furthermore, the added ablation study and benchmarks further explore the design decisions in our framework, shedding light on the crucial elements that enhance adaptivity and robustness in RL-based locomotion control.

We hope this work could serve as a milestone towards creating robust, versatile, and dynamic bipedal locomotion control in the real world, providing insights and guidance for future applications of RL for legged locomotion control and other complex systems.  

\begin{table}[!]
\scriptsize
\centering
\caption{A supplementary videos included in this paper.}
\label{tab:video_list}
\begin{tabular}{|c|l|l|}
\hline
\multicolumn{1}{|l|}{\textbf{Vid.}} & \textbf{Content} & \textbf{Link} \\ \hline
1 & Summary video & \url{https://youtu.be/sQEnDbET75g} \\ \hline
2 & Complete 400m dash & \url{https://youtu.be/wzQtRaXjvAk} \\ \hline
3 & Suppl. walking experiments & \url{https://youtu.be/vUewNUtSG3c} \\ \hline 
4 & Suppl. running experiments & \url{https://youtu.be/ad3ZrvUzbXM} \\ \hline
5 & Suppl. jumping experiments & \url{https://youtu.be/aAPSZ2QFB-E}\\ \hline
\end{tabular}
\end{table}

\section{Related Work}\label{sec:review}
Locomotion control for a bipedal robot requires solving a problem that tightly couples motion planning and control of the robot's whole body. 
Previous efforts can be broadly categorized into two main approaches: (1) model-based optimal control~(OC), and (2) model-free reinforcement learning~(RL). 
While this work is based on RL, the following review gives equal attention to relevant studies from both model-based OC and model-free RL domains. 
We hope it can provide a succinct survey of the recent trends in legged locomotion control, which could be informative for readers from both perspectives, with a primary focus on bipedal robots.
In the specific case of Cassie, which serves as the experimental platform in this work, we provide an overview of the most related work for its locomotion control in Table~\ref{tab:cassie_work}.

\begin{table*}[]
\centering
\tiny
\caption{Comparison of our work with prior studies on implementing real-world locomotion controllers for the bipedal robot Cassie. Our work provides an introduction to a general control framework to realize diverse periodic and aperiodic bipedal locomotion skills including robust walking and standing, fast running, and versatile jumping in the real world.}
\label{tab:cassie_work}
\begin{tabular}{ccccccc}
\hline
\multicolumn{7}{c}{\scriptsize{\textbf{Walking Skill}}} \\ \hline
\textbf{Previous Literature} & \textbf{Implementation} & \textbf{Variable Velocity} & \textbf{Variable Height} & \textbf{Consistency over Time} & \textbf{Consistent Perturbation} & \textbf{Change of Terrain} \\ \hline
\cite{gong2019feedback}& HZD, Model: Full-order & \textbf{Yes} & No & No & No & No \\
\cite{li2020animated,yang2022bayesian} & HZD, Model: Full-order & \textbf{Yes} & \textbf{Yes} & No & No & No \\
\cite{reher2021control} & HZD, Model: Full-order & \textbf{Yes} & No & Not demonstrated & Not demonstrated & No \\
\cite{xie2020learning} & RL, Model-free & \textbf{Yes} & No & Not demonstrated & Not demonstrated & No \\
\cite{siekmann2020learning} & RL, Model-free & Forward walking only & No & Not demonstrated & Not demonstrated & No \\
\cite{li2021reinforcement} & RL, Model-free & \textbf{Yes} & \textbf{Yes} & Not demonstrated & \textbf{Yes (untrained)} & No \\
\cite{siekmann2021sim} & RL, Model-free & \textbf{Yes} & No & Not demonstrated & Not demonstrated & \textbf{Yes (small, trained)} \\
\cite{siekmann2021blind} & RL, Model-free & \textbf{Yes} & No & Not demonstrated & Not demonstrated & \textbf{Yes (trained)} \\
\cite{dao2022sim} & RL, Model-free & Forward walking only & No & Not demonstrated & \textbf{Yes (trained)} & No \\
\cite{yu2022dynamic} & RL, Model-free & Sharp turn only & No & Not demonstrated & Not demonstrated & No \\
\cite{gong2022zero} & OC, Model: ALIP & \textbf{Yes} & No & Not demonstrated & Not demonstrated & \textbf{Yes (unmodeled)} \\
\cite{xiong20223} & OC, Model: H-LIP & \textbf{Yes} & \textbf{Yes} & Not demonstrated & Not demonstrated & \textbf{Yes (small, unmodeled)} \\
\cite{agrawal2022model} & OC, Model: Centrodial & \textbf{Yes} & No & Not demonstrated & Not demonstrated & \textbf{Yes (small, unmodeled)} \\
\textbf{Ours} & RL, Model-free & \textbf{Yes} & \textbf{Yes} & \textbf{Yes} & \textbf{Yes (untrained)} & \textbf{Yes (small, untrained)} \\ \hline
\multicolumn{7}{c}{\scriptsize{\textbf{Running Skill}}} \\ \hline
\textbf{Previous Literature} & \textbf{Implementation} & \textbf{Controlled Velocity} & \textbf{Transition from/to Standing} & \textbf{100m Dash Finish Time} & \textbf{400m Dash Finish Time} & \textbf{Uneven Terrain} \\ \hline
\cite{siekmann2021sim} & RL$\ddagger$ & No & Not demonstrated & Not demonstrated & Not demonstrated & Yes (small, trained) \\
\cite{yang2023impact} & OC$\ddagger$ & \textbf{Yes} & Not demonstrated & Not demonstrated & Not demonstrated & No \\
\cite{crowley2023optimizing} & RL with noticeable flight phase & No & Only transit from standing & \textbf{24.73s} & Not capable of turning & No \\
\textbf{Ours} & RL with noticeable flight phase & \textbf{Yes, w/ sharp turn (untrained)} & \textbf{Yes} & 27.06s & \textbf{2 min 34 sec} & \textbf{Yes (large, trained)} \\
\multicolumn{7}{l}{$\ddagger$ Although being termed as running, the demonstrated flight phase, foot clearance during flight, and speed in the real world are not comparable to the rest of the work listed here.}   \\ \hline
\multicolumn{7}{c}{\scriptsize{\textbf{Jumping Skill}}} \\ \hline
\multirow{2}{*}{\textbf{Previous Literature}} & \multirow{2}{*}{\textbf{Implementation}} & \multirow{2}{*}{\textbf{Targeted Landing}} & \multirow{2}{*}{\textbf{Apex Foot Clearance}} & \multirow{2}{*}{\textbf{Longest Flight Phase}} & \multicolumn{2}{c}{\textbf{Maximum Leap Distance}} \\ \cline{6-7} 
 &  &  &  &  & \multicolumn{2}{c}{\textbf{(Forward, Backward, Lateral, Turning, Elevation)}} \\ \hline
\cite{xiong2018bipedal} & Aperiodic Hop by OC$\ddagger$ & No & 0.18m & 0.42s & \multicolumn{2}{c}{In-place} \\ 
\cite{yang2021impact} & Aperiodic Hop by OC$\ddagger$ & No & 0.15m* & 0.33s* & \multicolumn{2}{c}{In-place} \\ 
\cite{siekmann2021sim} & Periodic Hop by RL$\ddagger$ & No & 0.16m* & 0.33s* & \multicolumn{2}{c}{Tracking a forward speed} \\ 
\cite{yang2023impact} & Aperiodic Jump by OC & No & 0.42m* & 0.33s* & \multicolumn{2}{c}{(0, 0, 0, 0, 0.41m)} \\
\textbf{Ours} & Aperiodic Jump by RL & \textbf{Yes} & \textbf{0.47m} & \textbf{0.58s} & \multicolumn{2}{c}{\textbf{(1.4m, -0.3m, $\pm$0.3m, $\pm$55$^\circ$, 0.44m)}} \\
\multicolumn{7}{l}{$\ddagger$ The demonstrated flight phase, apex foot clearance, and the resulting impacts upon landing are not comparable with the rest, so we use “hop" to distinguish from a “jump".} \\ 
\multicolumn{7}{l}{* Not provided in the paper and the listed value is roughly estimated from the accompanying video.} \\
\hline
\end{tabular}
\end{table*}

\subsection{Model-based Optimal Control for Bipedal Robots}
Locomotion control for bipedal robots can be formulated as an optimal control (OC) problem~\cite[Eq.~(1)]{wensing2022optimization}, with the robot's dynamics model appearing as a motion constraint. 
To manage the computational complexity of solving constrained optimization problems, this method often employs a cascaded optimization framework, starting with generating long horizon reference trajectories to low-level motion control and immediate reactive control, with progressively higher control rates and different modeling choices at each stage.

\subsubsection{Choice of Models}
The robot's full-order dynamics and contact models can be leveraged to optimize for a bipedal robot's trajectory and corresponding inputs for a specific behavior~\citep{hereid2019rapid,fevre2020rapid,marcucci2016two}.
However, given that bipedal robots typically have high-dimensional nonlinear dynamics, the use of detailed models is generally limited to offline optimization. 
Notably, the Hybrid Zero Dynamics (HZD) method \citep{westervelt2003hybrid} uses the bipedal robot's full-order model to design attractive periodic gaits offline with online feedback controllers to enforce the virtual constraints~\citep{hereid2018dynamic,reher2021control,gong2019feedback}. 
For online trajectory optimization, reduced-order models that simplify robot dynamics are necessary. 
Various reduced-order models, such as centroidal dynamics \citep{orin2013centroidal}, the linear inverted pendulum (LIP) \citep{kajita20013d}) and its variants like SLIP \citep{rummel2010stable}, ALIP \citep{gong2022zero}, H-LIP \citep{xiong20223}, are used for online optimization of reduced-order  dynamics, like the Center of Mass (CoM) and/or Center of Pressure (CoP) \citep{vukobratovic2004zero,pratt2012capturability,fernbach2020c}. These models also enable trajectory tracking control~\citep{kuindersma2014efficiently,daneshmand2021variable,gong2022zero}.
Reactive controllers, like whole-body control (WBC, \cite{sentis2006whole}), translate these reduced-order states to joint-level inputs, operating as a fast-solving QP that considers various constraints \emph{without} unrolling the full-order dynamics over a horizon like~\cite{moro2019whole,bouyarmane2011using,wensing2016improved}.
However, such a cascaded OC method does not fully utilize the robot's potential agility due to the limitations of reduced-order models and/or the inability of online controllers to re-plan whole-body maneuvers.
Our work overcomes these limitations by directly learning on the robot's full-order dynamics and approximately solving the OC problem by model-free RL. The resulting controllers are able to leverage the full agility potential of the underlying system.

\subsubsection{Contact Planning}
Legged robots require making and breaking contact with the environment. 
The resulting velocity jumps at contact make the robot's trajectory non-smooth.
This makes it challenging to solve an OC problem that decides each leg's contact mode throughout the movement \citep{posa2014direct}.
To simplify this, many studies, including most of the above-mentioned work, pre-define fixed contact sequences for specific locomotion skills, such as walking \citep{caron2019stair, hereid2019rapid, xiong20223}, running \citep{takenaka2009realrunning,sreenath2013embedding,ma2017bipedal}, and jumping \citep{goswami2009planar,chignoli2021humanoid,qi2023vertical}.
However, pre-defined handcrafted contact sequences may not be optimal. For instance, different jumping tasks might require varying contact schedules, like extended flight times for longer jumps. 
Consequently, there are efforts to integrate contact planning with trajectory optimization by computationally expensive mixed-integer programming~\citep{deits2014footstep,ibanez2014emergence}.
There are also attempts to avoid using explicit discrete variables, which results in contact-implicit methods, by enforcing complementarity constraints~\citep{dai2014whole,posa2014direct,drnach2021robust} or utilizing a bilevel optimization~\citep{landry2022bilevel,zhu2021contact,cleac2021fast}.  
However, given the high-dimensional nonlinear dynamics of bipedal robots, contact explicit or implicit planning is still limited to offline optimization.
As we will see, our method enables choosing contact plans in real-time for deployment in the real world.

\subsubsection{Scalability to Different Locomotion Skills}
The scalability of model-based OC across various bipedal locomotion skills and tasks is a significant challenge, largely due to the task-specific nature of robot models and control frameworks ~\cite{meduri2023biconmp}. 
For example, extending the HZD approach used for controlling 2D bipedal walking~\citep{sreenath2011compliant} to running~\citep{sreenath2013embedding} or 3D walking~\citep{da20162d} requires considerable efforts in finding appropriate periodic gaits and then designing specific controllers to stabilize them. Furthermore, HZD's reliance on the stability of periodic gaits limits its extension to aperiodic skills like jumping. 
This limitation is also observed in LIP-based methods, where separate frameworks and models are required for walking~\citep{takenaka2009realwalking} and running~\citep{takenaka2009realrunning} while jumping skills are excluded due to LIP's assumption of an approximately constant CoM height as in \cite{boroujeni2021unified}. 
Although in-place jumps have been achieved through other carefully engineered cascaded OC frameworks over different jumping phases, such as~\cite{xiong2018bipedal,kojima2019robot,qi2023vertical}, for each different jumping task, these require starting from scratch with offline trajectory optimization, often overlooking lateral or turning motions.
Furthermore, it is very challenging to develop a single model-based controller capable of managing diverse \emph{targeted} bipedal jumping tasks in the real world. This is because besides accomplishing a jump, the controller also needs to quickly adapt to the dynamics of the robot hardware and produce an accurate translational velocity at the take-off to land at the targets located at different places, as shown in this work.\footnote{While there are attempts from industry (like Boston Dynamics as patented in~\cite{deits2022robot}) using model-based OC that tackled similar problems, detailed reports are limited. As a research paper, our focus is on published findings.}
The model-based OC community, as seen in works like~\cite{yang2023impact}, is making promising strides toward developing general tracking controllers that are invariant to impact. 
These could potentially handle different contact sequences like walking, running, and jumping, but still rely on \emph{task-specific} optimized trajectories.

\subsection{Model-free Reinforcement Learning on Legged Locomotion Control}
Recent developments in deep RL have brought about exciting progress in creating locomotion controllers for quadrupedal robots in the real world, such as~\cite{margolisyang2022rapid,chen2022learning,feng2023genloco,fu2023deep}. 
However, due to the inherently less stable nature of bipedal robots, methods successful with quadrupeds might not directly apply to bipedal systems, as an example seen in~\cite{kumar2022adapting}. 
Therefore, within the bipedal robotics community, there are different strategies to employ RL tailored to the challenges of high dimensional nonlinearity.

\subsubsection{Control Policy Structure}
In RL-based locomotion control, the structure of the control policy is largely influenced by how observations are formulated, particularly through the use of robot states-only history (with no robot's input history) or I/O history (with both the robot's input and output).
For quadrupedal robots, there is no consensus on the history length to use: it ranges from a short I/O history of 1 to 15 timesteps~\citep{hwangbo2019learning,escontrela2022adversarial,huang2022creating} to longer sequences over 50 timesteps involving states-only history~\citep{lee2020learning,miki2022learning,shao2021learning} or I/O history~\citep{kumar2021rma}. 
The policy (neural network) architecture is chosen depending on the history length, with MLPs suited for shorter histories and recurrent units needed for longer sequences. 
Real-world deployments on quadrupedal robots have shown comparable performance across these varying history lengths.
For bipedal robots, a trend toward longer history lengths is observed, evolving from a single timestep state feedback with a need of residual action~\citep{xie2018feedback,xie2020learning,rodriguez2021deepwalk,castillo2022reinforcement}, to a short I/O history~\citep{li2021reinforcement}, to a longer sequence of states-only history~\citep{siekmann2020learning} or I/O history~\citep{kumar2022adapting,radosavovic2023learning}. 
While utilizing long history for robotic control as suggested by \cite{peng2018sim} is a common strategy, most prior ablation studies focus on contrasting long histories with immediate state or I/O feedback through MLPs, like \cite{peng2018sim,lee2020learning,kumar2021rma,siekmann2020learning}, with less exploration into shorter I/O histories. 
A recent study, \cite{singh2023learning}, reported that, short state histories yield better learning performance than a longer history in bipedal humanoid robot control, aligning with our finding in this work. 
This raises a question: while long histories are expected to enhance performance in RL-based control, how can their benefits be fully utilized? 
Our study presents an effective solution: a dual-history approach.

\subsubsection{Sim-to-real Transfer}
There are some attempts using RL to directly collect data and train on the robot hardware, such as \cite{haarnoja2019learning,wu2023daydreamer,smith2023demonstrating}, and there are also alternatives on leveraging a pre-training stage in simulation and finetuning on hardware like \cite{peng2020learning,smith2022legged,westenbroek2022lyapunov}, most of which are deployed on small-sized quadrupedal robots.
However, for human-sized bipedal robots, performing hardware rollouts is expensive, making it more appealing to directly transfer diverse dynamic bipedal skills from simulation to hardware.
Achieving zero-shot transfer needs extensive dynamics randomization in simulation as suggested by \cite{peng2018sim,peng2020learning}. 
There are two primary approaches to training policies under randomized dynamics: (1) end-to-end training with a history of robot measurements or I/O, which has been applied to bipedal robots like \cite{siekmann2020learning,siekmann2021sim,li2021reinforcement}, and (2) policy distillation where, with separated training stages, an expert policy with access to privileged environmental information supervises the training of a student policy with proprioceptive feedback. This method, initially developed for small servo-controlled bipedal robots in \cite{yu2019sim}, has been adapted to quadrupeds using strategies like Teacher-Student (TS, \cite{lee2020learning}) or RMA~\citep{kumar2021rma}, and is prevalent in quadrupedal robot community (\textit{e.g.}, \cite{fu2021minimizing,ji2022concurrent,margolis2023walk}). Though policy distillation offers advantages in quadrupedal locomotion controls~\citep{kumar2021rma}, its extension to torque-controlled bipedal robots benefits from an additional finetuning stage like \cite{kumar2022adapting}.
In this work, we demonstrate that end-to-end training is a more effective approach for developing controllers for bipedal robots encompassing a variety of dynamic locomotion skills.

\subsubsection{Scalability to Different Locomotion Skills}
Leveraging RL to learn diverse locomotion skills or tasks with a single policy poses a challenge that arises from the need to optimize multiple objectives for different tasks, as noted in \cite{kalashnikov2021mt}. 
Initially, efforts in this field focused on single skills with fixed tasks, like just walking forward in \cite{peng2020learning,siekmann2020learning,kumar2021rma}.
In developing a single-skill policy capable of multiple tasks, methods that provide varying commands for tracking different walking velocities without specifying reference motions have been considered, like \cite{hwangbo2019learning,fu2021minimizing,cheng2023legs}. 
While this method is viable for quadruped robots, it demands extensive reward tuning and may not suit bipedal robots due to their higher dimensionality.
For bipedal robots, approaches such as providing parameterized reference motions~\citep{li2021reinforcement} or using policy distillation from task-specific policies~\citep{xie2020learning} have been considered. 
Some methods also involve commanding periodic contact sequences, resulting in diverse but periodic bipedal gaits like \cite{siekmann2021sim}. 
However, the prescriptive contact sequence restricts the robot's potential to optimize its contact strategy for improved stability. 
Additionally, limitations arise when such a policy attempts to accomplish specific tasks, leading to follow-up works that require specialized training for distinct behaviors. 
This includes requiring hand-tuned strategies for transitioning between fast straight running and standing~\citep{crowley2023optimizing}, load carrying \citep{dao2022sim}, or sharp turns during walking \citep{yu2022dynamic}.
There are other attempts that sought to create a single policy that can perform diverse skills, regardless of periodicity, for bipedal humanoids using adversarial motion priors as developed in \cite{peng2021amp} in simulation. 
While this has seen success in transfer to real quadrupeds like \cite{escontrela2022adversarial,wu2023learning,li2023learning}, it is still an open question if more dynamic bipedal skills can be transferred to the real world due to larger sim-to-real gaps.
In light of these challenges, our work strikes a balance in versatility for bipedal robots.
We focus on developing skill-specific control policies that can perform a diverse set of tasks while maintaining a general framework suitable for developing different skills.
\section{Overview}\label{sec:overview}
\begin{figure}[t]
    \centering
    \includegraphics[width=\linewidth]{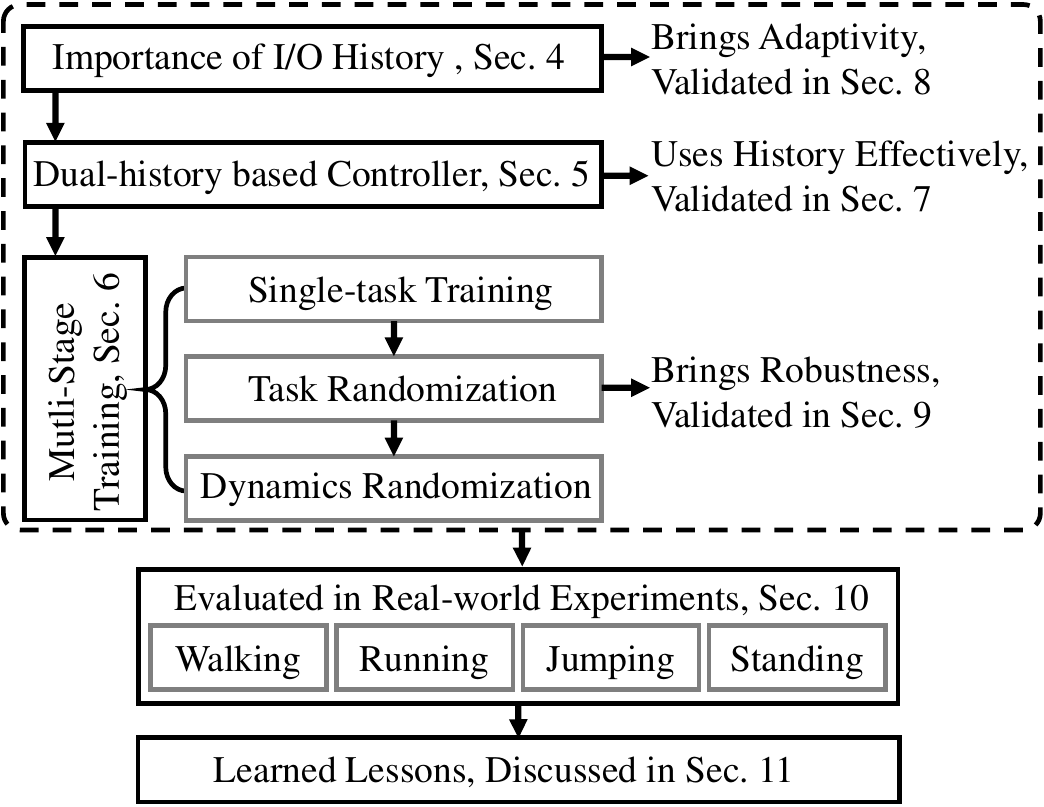}
    \refstepcounter{figure}
    \caption{Overview of this paper. First, Sec.~\ref{sec:background} introduces the formulation of the locomotion control problem and the importance of utilizing the robot's I/O history. The details of our dual-history-based control architecture for various bipedal locomotion skills are presented in Sec.~\ref{sec:controller}, followed by the training scheme discussed in Sec.~\ref{sec:training}. Then, detailed studies are conducted to validate the advantages of the proposed policy structure in Sec.~\ref{sec:policy_structure}, sources of adaptivity in Sec.~\ref{subsec:adaptivity}, and we investigate the sources of the robust behaviors observed in the proposed RL-based controller in Sec.~\ref{sec:multi_skill}. Extensive experiments using the proposed RL-based locomotion controllers to enable Cassie to perform robust standing, walking, running, and jumping skills are presented in Sec.~\ref{sec:experiments}. Insights and discussions for readers interested in applying RL to train bipedal robots are provided in Sec.~\ref{sec:discussion}.}
    \label{fig:overview}
\end{figure}

In this section, we provide an overview of the entire paper and proposed RL system for general bipedal locomotion control, as illustrated in Fig.~\ref{fig:overview}. 
We first provide an introduction of the importance of utilizing the robot's I/O history in the locomotion control in Sec.~\ref{sec:background}. In this section, we showcase that the robot's long I/O history can enable system identification and state estimations during real-time control, from both control and RL perspectives.
This results in the design of the backbone of this work: a new control architecture that utilizes a dual-history of both the bipedal robot's long-term and short-term I/O history, which is presented in Sec.~\ref{sec:controller}. 
Specifically, such a control architecture does not only utilize the long history but also exploits explicit short history of the robot.
The proposed dual-history structure enables effective use of robot's history, as the long history brings adaptivity (validated in Sec.~\ref{subsec:adaptivity}) and the short history complements the use of long history by enabling better real-time control (substantiated in Sec.~\ref{sec:policy_structure}). 
This control policy, which is represented by a deep neural network, is optimized by model-free RL in Sec.~\ref{sec:training}.
Since this paper aims to develop a controller that is capable of accomplishing diverse tasks using highly dynamic locomotion skills, the training in Sec.~\ref{sec:training} is characterized by multi-stage training in simulation. 
This training strategy provides a structured curriculum, starting with a single-task training where the robot focuses on a fixed task, followed by task randomization that diversifies the tasks the robot is trained on, and concluded by dynamics randomization which alters the dynamics parameters of the robot.
Such a training strategy is able to provide a versatile control policy that can perform a large variety of tasks and zero-shot transferred to the robot hardware.
Furthermore, task randomization can also enhance the robustness of the resulting policy through the generalization among different learned tasks. We show that such robustness results in compliant behaviors to disturbance, which is “orthogonal" to the one brought by dynamics randomization. This is validated in Sec.~\ref{sec:multi_skill}.

By using this framework, we obtain skill-specific versatile policies for walking, running, and jumping for a bipedal robot, Cassie. We evaluate the effectiveness of these control policies extensively in the real world in Sec.~\ref{sec:experiments}. 
As we will see, by the adaptivity of the proposed control architecture, the resulting policies can accomplish various tasks with minimal degradation from simulation to the real world and maintain consistency over a long timespan (more than one year). Because of the robustness brought by the proposed training strategy, especially the task randomization, the resulting policies showcase complicated recovery maneuvers when facing unexpected disturbances in the real-time and real world.   
We further summarize the insights and lessons learned during the development of this work in a discussion in Sec.~\ref{sec:discussion}, followed by a conclusion and future work in Sec.~\ref{sec:conclusion}.

\section{Background}\label{sec:background}
In this section, we introduce Cassie, our main experimental platform, along with its dynamics model. 
We then frame the bipedal locomotion control problem as a Partially Observable Markov Decision Process (POMDP), laying the groundwork for training policies with RL. We also emphasize the importance of incorporating the robot's input and output (I/O) history in feedback control from both model-based control and model-free RL perspectives.

\subsection{Cassie Robot Model}
\subsubsection{Floating-base Coordinates}\label{subsec:states}
As illustrated in Fig.~\ref{fig:controller}, Cassie is a human-sized bipedal robot, standing at a height of 1.1 m and weighing 31 kg.
On both the Left and Right (L/R) legs are 7 joints, consisting of the abduction $q_1$, rotation $q_2$, thigh $q_3$, knee $q_4$, shin $q_5$, tarsus $q_6$, and toe pitch $q_7$. This results in a total of 14 joints ($\mathbf{q}_j \in \mathbb{R}^{14}$).
Among those, $q^{L/R}_{1,2,3,4,7}$ are \emph{actuated} by motors (which is denoted as $\mathbf{q}_m \in \mathbb{R}^{10}$), while the shin and tarsus joints ($q^{L/R}_{5,6}$) are \emph{passive} and connected by leaf springs, as annotated in Fig.~\ref{fig:controller}. 
Cassie has a floating base $\mathbf{q}_b$ with 6 Degree-of-Freedoms~(DoFs) which are translational positions $q_{x,y,z}$ and rotational positions $q_{\phi,\theta,\psi}$. 
The generalized coordinates of the full system $\mathbf{q}$ can be represented as $\mathbf{q}=[\mathbf{q}_b, \mathbf{q}_j] \in \mathbb{R}^{20}$. 
\paragraph{The observable states}
Since the robot is a second-order mechanical system, we can denote the robot's states as its generalized coordinates $\mathbf{q}$ and their time derivatives $\dot{\mathbf{q}}$. 
However, there is only a part of states we can \emph{reliably} measure or estimate using the robot's onboard sensors. 
We denote observable states as $\mathbf{o}\in \mathbb{R}^{26}$ and it contains motors positions and their velocities ($\mathbf{q}_m$, $\dot{\mathbf{q}}_m$), which can be respectively measured and estimated by the joint encoders. The base orientation ${q}_{\phi,\theta,\psi}$ can be measured by an IMU, and base linear velocity $\dot{q}_{x,y,z}$ can be estimated by an EKF as used in~\citep{xie2020learning}.

\subsubsection{Full-order Dynamics Model}~\label{subsec:full_order_dynamics}
Cassie has a floating-base, a total of $n=20$ DoFs, and $n_a=10$ actuated joints, Its dynamics equation can be obtained by the Euler-Lagrange method:
\begin{equation}\label{eq:full_order_dynamics}
    \mathbf{M}(\mathbf{q})\ddot{\mathbf{q}} + \mathbf{C}(\mathbf{q}, \dot{\mathbf{q}})\dot{\mathbf{q}} + \mathbf{G}(\mathbf{q}) = \mathbf{B}\bm{\tau} + \bm{\kappa}_{\text{sp}}(\mathbf{q},\dot{\mathbf{q}}) + \bm{\zeta}_{\text{ext}},
\end{equation}\noindent 
where $\mathbf{M} \in \mathbb{R}^{n\times n}$, $\mathbf{C} \in \mathbb{R}^{n\times n}$, and $\mathbf{G}\in \mathbb{R}^{n}$ denote the generalized mass matrix, centrifugal and Coriolis matrix, and generalized gravity, respectively.  
The right-hand side of \eqref{eq:full_order_dynamics} contains the system inputs which includes generalized control input (motor torques) $\bm{\tau}\in\mathbb{R}^{n_a}$ (distributed by $\mathbf{B}\in \mathbb{R}^{n\times n_a}$), state-dependent spring torques $\bm{\kappa}_{\text{sp}}(\mathbf{q},\dot{\mathbf{q}})$, and generalized external force $\bm{\zeta}_{\text{ext}}$.
The $\bm{\zeta}_{\text{ext}}$ groups all the external forces applied from the environment, including foot contact wrenches denoted as $\mathbf{J}^T_{c} \mathbf{F}_c$ and any joint-level friction or perturbation wrenches added to the robot. 
Specifically, for the contact wrenches, $\mathbf{J}_c(\mathbf{q}) \in \mathbb{R}^{n_c\times n}$ is the contact Jacobian and the dimension of contact wrenches ($n_c$) will vary when the robot has a different number of support legs with the ground (ranging from no stance leg to two stance legs).

\subsubsection{System Identification and Adaptive Control}
For locomotion control of bipedal robots like Cassie, we can borrow ideas from system identification and adaptive control, which can adapt to changes in the dynamics of the robot. One approach is to leverage a sequence of the past system's input and output (I/O history) to identify the system parameters and change the control law over time, see \cite[Ch. 9]{landau2011adaptive}. 
For example, given the dynamical system (bipedal robot) governed by \eqref{eq:full_order_dynamics}, by utilizing a sequence of robot's input ($\bm{\tau}$) and robot's output ($\mathbf{o}$), with an assumption that the system states ($\mathbf{q}$ and their time derivatives) can be approximated by a sequence of $\mathbf{o}$ as derived in \cite{lim2023proprioceptive}, the modeling parameters in $\mathbf{M}$, $\mathbf{C}$, $\mathbf{G}$, $\bm{\kappa}$ and external forces $\bm{\zeta}_{\text{ext}}$ can be identified and therefore the control strategy in \eqref{eq:full_order_dynamics} can be adjusted accordingly. 
There are two main design choices in adaptive control: \emph{indirect} and \emph{direct} methods. 
In the \emph{indirect} approach, system parameters are first explicitly estimated (see~\cite{ljung1998system}) and the control law is indirectly adjusted based on the identified parameters. 
In contrast, the \emph{direct} approach alters the control law directly based on the system's I/O history. 
While both methods have their advantages and disadvantages as discussed in~\cite{annaswamy2023adaptive}, the \emph{indirect} approach may struggle with unknown system parameters or the ones that are challenging to identify. 
Therefore, in this paper, we choose to develop a control policy through model-free RL by learning from the bipedal robot's I/O history to directly adapt to changes in the robot's full-order model \eqref{eq:full_order_dynamics}. 
Our method is designed to align with the \emph{direct} adaptive control category in contrast to the strategies like Teacher-Student~\citep{lee2020learning} or RMA~\citep{kumar2021rma} that need to estimate pre-selected system parameters, which can be viewed as \emph{indirect} methods.

\subsection{RL Preliminaries}
\subsubsection{Legged Locomotion Control as a POMDP}
The locomotion control on bipedal robots can be formulated as a Partially Observable Markov Decision Process (POMDP). 
At each timestep $t$, the environment is at state $\mathbf{s}_t$, the agent (\textit{i.e.}, the robot) makes an observation $\mathbf{o}_t$ from the environment, takes an action $\mathbf{a}_{t}$ and interacts with the environment, the environment transits to a new state $\mathbf{s}_{t+1}$, and the agent receives a reward $r_t$.  
Such a process will repeat until the end of the episode of length $T$. 
In a POMDP, the robot only has access to partial information of the environment, \textit{e.g.,} the bipedal robot can only access the observable states $\mathbf{o}$ instead of the full environment states, which include the full coordinates $\mathbf{q}$ and their time derivatives as explained in Sec.~\ref{subsec:states}.
In a POMDP, the observation $\mathbf{o}_t$ is obtained by the observation function $O(\mathbf{o}_{t} | \mathbf{s}_{t}, \mathbf{a}_{t-1})$ conditioned on both the current environment state and action the agent has taken.
The RL objective is to maximize the expected return, $\mathbb{E}[\sum^T_{t=0} \gamma^t r_t]$, by finding an optimal policy $\pi^*$ that selects an action $\mathbf{a}_t$ at each time step.
The expected return is the sum of discounted rewards collected by the agent throughout an episode, with $\gamma$ representing a discount factor.

\subsubsection{Solving POMDP with I/O History}
Solving a POMDP can be formulated as finding the optimal policy $\pi^*$ that maps the \emph{process history} to the optimal action. 
The \emph{process history} at timestep $t$ contains the entire history of the agent's observations and actions, \textit{i.e.}, $\{<\mathbf{o}_t,\mathbf{a}_{t-1}>, <\mathbf{o}_{t-1},\mathbf{a}_{t-2}>, \dots, <\mathbf{o}_1,\mathbf{a}_{0}>\}$~\citep{spaan2012partially}. 
In the control context, it is the robot's I/O history. 
One method is to update the belief state from the process history with Bayesian filters at each timestep, and then transform the POMDP into a belief state MDP. 
Alternatives can be to formulate a Finite State Controller (FSC) that contains the memory of agent's past observations and actions through \emph{internal states} to constitute and optimize a policy graph \cite[Ch. 4]{spaan2012partially}.
While the full history is needed to find an optimal policy, it is computationally intractable. 
Instead, a finite memory can be kept in the FSC and the POMDP can be solved \emph{approximately} (see \cite{meuleau2013learning}).
In all of these methods, the robot's process history (\textit{i.e.}, I/O history) is utilized for solving a POMDP. 
Therefore, we choose to train a policy that has a finite memory (with a fixed length $h$) of robot's I/O pair, \textit{i.e.},  $\pi(\mathbf{a}_t | \mathbf{o}_{t:t-h},\mathbf{a}_{t-1:t-h-1})$. 
Furthermore, in this work, to enable the robot to accomplish a variety of goals, we parameterize tasks using commands $\mathbf{c}$, and the policy $\pi(\mathbf{a}_t | \mathbf{o}_{t:t-h},\mathbf{a}_{t-1:t-h-1}, \mathbf{c}_t)$ is now conditioned on both the robot's I/O history and the given command $\mathbf{c}_t$.

\subsubsection{Task Parameterization}~\label{subsec:command}
Different locomotion tasks are parameterized by a command and this parameterization may vary for different locomotion skills. 
For the walking skill, the command $\mathbf{c}_{\text{walk}} \in \mathbb{R}^4$ is defined as $[\dot{q}^d_{x,y}, q^d_{z,\psi}]$, which specifies the desired walking velocity in sagittal $\dot{q}^d_x$ and lateral $\dot{q}^d_y$ directions, desired walking height $q^d_z$, and desired turning yaw angle $q^d_{\psi}$, respectively. 
For the running skill, the command is represented by $\mathbf{c}_{\text{run}}=[\dot{q}^d_{x,y}, q^d_{\psi}] \in \mathbb{R}^3$. 
For jumping, the command $\mathbf{c}_{\text{jump}} \in \mathbb{R}^4$ is defined as $[q^d_{x,y,\psi}, e^d_z]$, which specifies the target planar position $q^d_{x,y}$, turned angle $q^d_{\psi}$, and change of elevation $e^d_z$ in the vertical jump direction after the robot lands. 
\section{Bipedal Locomotion Controller with I/O History}\label{sec:controller}
In this section, we describe the proposed general control framework for bipedal robots, leveraging the robot's dual I/O history as a fundamental element to enable transfer to environments with uncertain dynamics. 
This control architecture is the backbone of the RL system developed in this work, which will show advantages in terms of control performance in both the simulation and the real world. 

\begin{figure}[t]
    \centering
    \includegraphics[width=\linewidth]{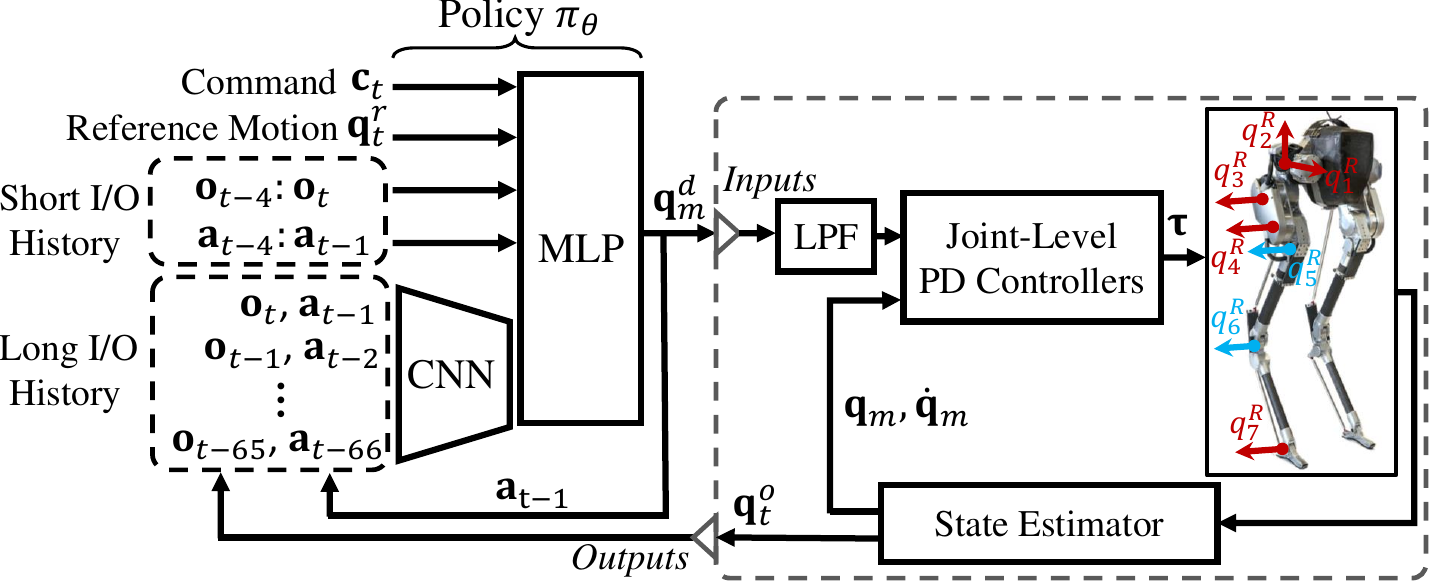}
    \caption{The proposed RL-based controller architecture that leverages a dual-history of input ($\mathbf{a}$) and output ($\mathbf{o}$) (I/O) from the robot. The control policy $\pi_\theta$, operating at 33 Hz, processes a 2-second long I/O history. This data is initially encoded via a 1D CNN along its time axis before being merged with a base MLP. In addition, a short history spanning 4 timesteps is directly input into the base MLP, combined with skill-specific reference motion $\mathbf{q}^r_t$ and variable commands $\mathbf{c}_t$ that parameterize the tasks. The policy outputs desired motor positions $\mathbf{q}^d_m$ as the robot's actions, which are then smoothed using a low-pass filter (LPF). These filtered outputs are employed by joint-level PD controllers operating at 2 kHz to specify motor torques $\bm{\tau}$. This architecture is general for various locomotion skills like standing, walking, running, and jumping. This figure also annotates the generalized coordinates for Cassie, which include actuated joints ($q^{L/R}_{1,2,3,4,7}$, marked as red) and passive joints ($q^{L/R}_{5,6}$, marked as blue).
    }
    \label{fig:controller}
\end{figure}

\paragraph{Control Framework}
Our locomotion control policy $\pi_\theta$ is represented by a deep neural network with parameters $\theta$. 
The same policy architecture will be used for a variety of different locomotion skills, and as such, it is crafted to rely on minimal skill-specific design choices. 
As shown in Fig.~\ref{fig:controller}, the policy outputs the desired robot motor positions $\mathbf{q}^d_{m}\in \mathbb{R}^{10}$ for the robot, which is the agent's action $\mathbf{a}_t$. 
The action is first smoothed by a Low Pass Filter~(LPF)~\citep{peng2020learning}, details of which are discussed in Appendix~\ref{appendix:lpf}. The filtered actions are then used by joint-level PD controllers to calculate the motor torques $\bm{\tau}$ that will be applied to the robot's actuated joints. The policy is queried at a rate of 33 Hz, whereas the PD controllers operate at a higher frequency of 2~kHz.

The policy's input at each timestep $t$ consists of four components: given command $\mathbf{c}_t$, reference motion $\mathbf{q}^r_t$, the robot's short I/O history $<\mathbf{o}_{t:t-4}, \mathbf{a}_{t-1:t-4}>$, and the robot's long I/O history $<\mathbf{o}_{t:t-65}, \mathbf{a}_{t-1:t-66}>$.
The time-varying command $\mathbf{c}_t$, as defined in Sec.~\ref{subsec:command} represents the task the robot is to accomplish using the desired locomotion skill. 
The locomotion skill is specified by a \emph{preview} of a skill-specific reference motion $\mathbf{q}^r_t$ at the current timestep $t$ for the robot.
It includes the incoming desired motor positions for the robot $\mathbf{q}^r_t=[\mathbf{q}^d_m(t+1), \mathbf{q}^d_m(t+4), \mathbf{q}^d_m(t+7)]$ which is sampled at timesteps that are 1, 4, and 7 ahead. 
The preview of the reference motion conveys the upcoming desired trajectory to the robot, aiding it in avoiding being shortsighted.
If the desired base height is not included in command $\mathbf{c}_t$, such as for running and jumping, the current base height from the reference motion $q^r_z(t)$ is then also included in the $\mathbf{q}^r_t$.

\paragraph{The use of Robot's I/O History} 
In addition to the observations described above, the observations also include a history of the robot's inputs and outputs (I/O) to the control policy. 
As depicted in Fig.~\ref{fig:controller}, the \emph{inputs} are represented by the desired motor positions (the robot's actions $\mathbf{a}$), while the \emph{outputs} are represented by the observable states of the robot, denoted as $\mathbf{o}$ and described in Sec.~\ref{subsec:states}.
This history is processed through two streams, which we refer to as a dual-history approach. 
The first stream offers a brief, four-timestep history of the robot's I/O, which is directly provided as input into the base network. 
This short history, lasting approximately 0.1 seconds, provides the robot with recent feedback for real-time control.
In addition to the short history, a longer I/O history, spanning two seconds, is also provided as input to the policy. This long history comprises of 66 pairs of robot I/O data ($<\mathbf{o}_{t-k}, \mathbf{a}_{t-k-1}>$, $k\in[0,65] \cap \mathbb{Z}$). 
This long I/O history contributes more to the identification of the system's dynamics, encompassing elements like the LPF, PD controllers, the robot itself governed by \eqref{eq:full_order_dynamics}, and its state estimator.
Such a structure allows the controller to effectively leverage information from both short-term and long-term I/O history. 
It is important to leverage both of these information. 
The long-term history is useful for system identification and inferring state estimates, especially for ballistic movements involved during the flight phase. 
The short-term history is also important from two perspectives: it can provide \emph{explicit} feedback of recent measurements on the robot, which is critical for real-time control, and it can help the robot to determine the weight of the past information which sometimes may not be important.
As we will show in Sec.~\ref{sec:policy_structure}, this dual-history structure leads to significant performance improvements for bipedal locomotion control.

\paragraph{Details of Policy Representation}
As depicted in Fig.~\ref{fig:controller}, the policy architecture $\pi_\theta$ consists of two main components: a base network, modeled by a multilayer perceptron (MLP), and a long-term history encoder modeled by a 1D convolutional neural network (CNN), which computes an embedding of the long history that is provided as input to the base network.  
The base MLP has two hidden layers with 512 \textit{tanh} units each. 
The 1D CNN encoder consists of two hidden layers. 
Their configurations, defined as [kernel size, filter size, stride size], are [6, 32, 3] and [4, 16, 2] with \textit{relu} activation and no padding, respectively. 
The 66-timestep long I/O history is encoded via this CNN encoder through temporal convoluations along the time axis, and then compressed into a latent representation before being provided as input to the base MLP.
The output layer of the base MLP consists of \textit{tanh} units that specify the mean of the Gaussian distribution of the normalized action (w.r.t. the motor range). 
The standard deviation of the action distribution is specified by a fixed value $0.1I$.

\paragraph{General Policy Structure for Different Skills}
The control policy structure introduced in Fig.~\ref{fig:controller} is a general design and can be widely applied to a large variety of locomotion skills, such as standing, walking, running, and jumping. 
To train policies for different locomotion skills, a user needs only to simply provide different reference motions and commands to the policy, while the underlying architecture of the policy remains unchanged.
Throughout this paper, the same control policy architecture will be used for all experiments.

\section{Multi-Stage Training for Versatile Locomotion Controllers}\label{sec:training}
Following the construction of the bipedal locomotion controller, our next step in this section involves developing a general framework for training the control policy through reinforcement learning. 
Much like the control structure, this training framework extends beyond a single, specific locomotion skill and is general to various skills.
Such a framework is designed to train the robot in simulation and to transfer to the robot hardware without further fine-tuning. 

\begin{figure}[t]
    \centering
    \includegraphics[width=\linewidth]{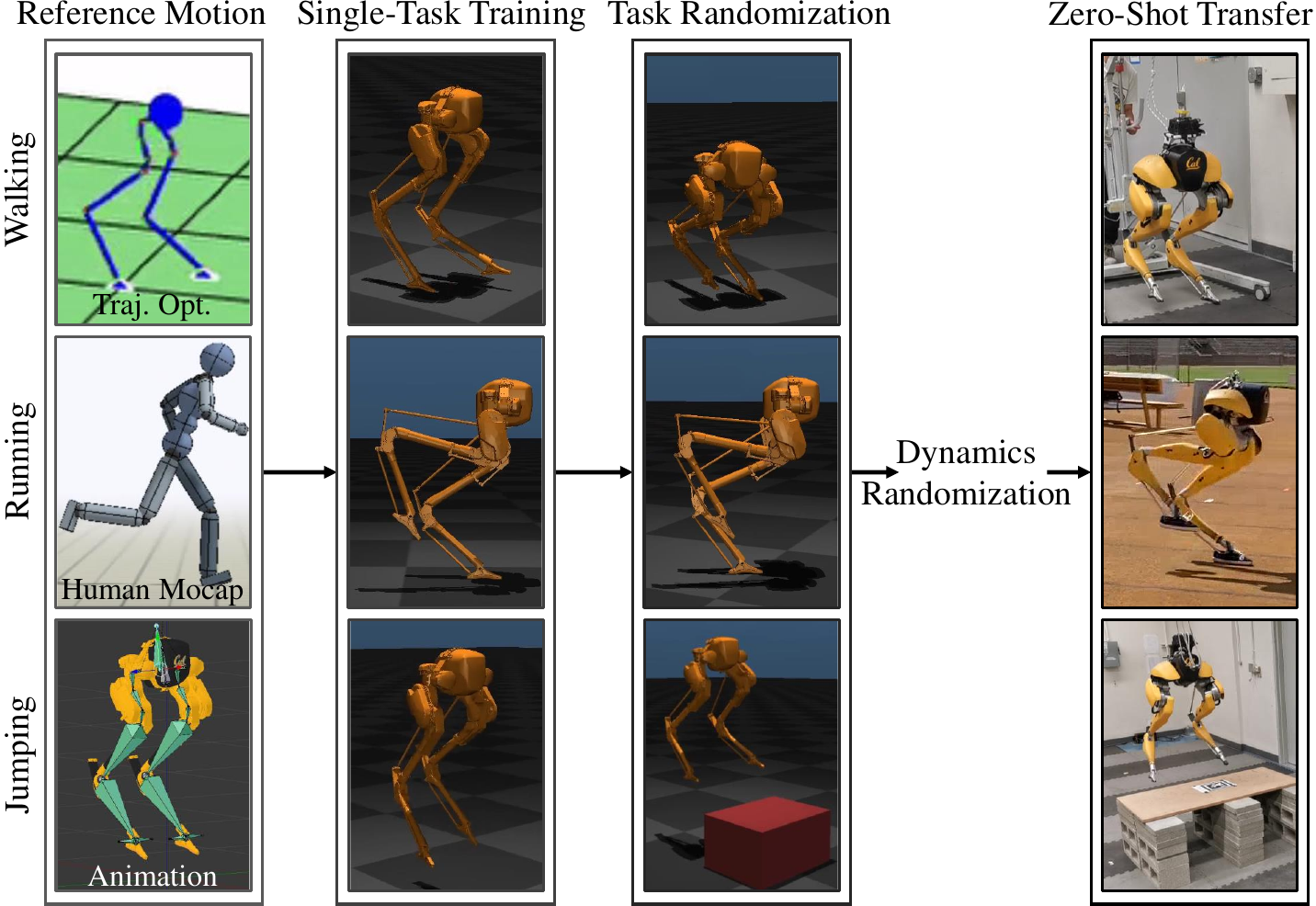}
    \caption{The multi-stage training framework to obtain a versatile control policy that can be zero-shot transferred to the real world. It starts with single-task training stage, where the robot is encouraged to mimic a single reference motion with a fixed goal. This is followed by task randomization stage, which expands the range of tasks the robot learns and fosters task generalization resulting in a versatile policy. Once the robot is adept at various locomotion tasks and their transitions, extensive dynamics randomization is incorporated to enhance policy robustness for sim-to-real transfer. This framework is suitable for diverse bipedal locomotion skills, including walking, running, and jumping, and for learning from different sources of skill-specific reference motions such as trajectory optimization, human mocap, and animation.}
    \label{fig:multi_stage_training}
\end{figure}

\subsection{Overview}
It is challenging to train a robot to accomplish diverse tasks through a single control policy via RL. 
This challenge is further compounded when dealing with highly dynamic locomotion maneuvers where the robot has limited support, such as varying running speeds or jumping to different locations. 
In such scenarios, the robot often struggles to learn effectively and may adopt overly conservative strategies, such as merely standing, to circumvent the challenges.
Therefore, we develop a multi-stage training strategy that incorporates a structured curriculum to facilitate the training of versatile locomotion control policies. 
This strategy can be summarized into three stages as shown in Fig.~\ref{fig:multi_stage_training}: (1) single-task training, (2) task randomization, and (3) dynamics randomization.
In the first stage, we concentrate on training the robot to acquire a locomotion skill from scratch, employing a fixed command. 
The primary aim is to equip the robot with the ability to master the skill itself, such as just walking forward, running forward, or jumping in place, while avoiding undesired maneuvering strategies.
In the second stage, we introduce diverse commands to encourage the robot to perform a large variety of tasks using the skill it has acquired to develop a versatile policy.
Following the robot's proficiency in a simple simulation environment, the third stage implements extensive randomization of dynamics parameters in the simulation. 
This process is designed to robustify the policy to ensure a successful zero-shot transfer from simulation to robot hardware.

The design of POMDP, including aspects like reward and episode design, may indeed differ across different stages to serve specific objectives. 
However, the overall multi-stage training scheme remains a general approach to developing different locomotion skills, and as we will see, this method requires only the change of skill-specific reference motion and hyperparameters for learning different skills.

\paragraph{Combining a Standing Skill} 
Learning a standing skill for the bipedal robot is useful for deployment in the real world, where the robot may need to come to a stop after walking, running, or jumping. 
In the context of the aperiodic jumping skill, the robot learns to maintain a stance pose during the post-landing phase alongside the jumping skill. 
However, this standing skill is not introduced for periodic skills like walking and running. 
To address this gap, in Stage 2, we introduce an additional sub-stage to enable the robot to learn the transition to standing (and back) once it has acquired a versatile policy, with details introduced in Appendix~\ref{appendix:transition_to_stand}.
While previous approaches have explored using separate policies for such a transition, as demonstrated in \cite{crowley2023optimizing}, our work showcases the advantages of having a single policy for transitioning between standing and other locomotion skills. 
This approach can not only realize a rapid transition but also enable the robot to generalize the learned locomotion skill to significantly improve robustness during standing.

\subsection{Reference Motion}
For each locomotion skill, we provide one or a set of reference motions, which provide examples of the type of locomotion maneuvers that the robot is to perform.
As shown in Fig.~\ref{fig:multi_stage_training}, our framework can accommodate diverse sources of reference motion, including reference motions from trajectory optimization, motion capture, and keyframe animations.

\paragraph{Trajectory Optimization}
For the \emph{walking} skill, we leverage the trajectory optimization method to generate a library of reference motions that depict diverse periodic walking gaits based on the robot's full-order dynamics.
The resulting \textit{gait library} is parameterized by the walking commands $[\dot{q}^d_x, \dot{q}^d_y, q^d_z]$ ranging from $[-1.0, -0.3, 0.65]$ to $[1.0, 0.3, 1.0]$, and consists of 1331 different reference motions.
One reference motion is represented by a set of B\'ezier trajectories of each actuated motor with a fixed timespan of the walking period (0.8 seconds). 
A more detailed description of the process for generating the gait library is provided in~\cite[Sec. III-B]{li2020animated} and \citep{hereid2019rapid}. 
Note that we do not consider turning yaw command $q^d_\psi$ when building the reference gait library.  

\begin{table*}[!htp]
\centering
\scriptsize
\caption{The components $\mathbf{r}$ and their respective weights $\mathbf{w}$ in the reward function used for training the bipedal robot across various skills and training stages. A nominal weight vector is provided for different components, with adjustments necessary based on the skill and stage of training, primarily influenced by task diversity (training for single goal versus diverse goals) and the existence of a flight phase. The reward tuning process is relatively streamlined, as only 24 out of 72 reward terms vary across the three different bipedal locomotion skills over multi-stage training. Blank cells indicates no change on the corresponding nominal value.}
\label{tab:reward}
\begin{tabular}{|cccccccc|}
\hline
\multicolumn{1}{|c|}{\multirow{3}{*}{\textbf{Reward Component} $\mathbf{r}$}} &
  \multicolumn{7}{c|}{\textbf{Weight} $\mathbf{w}$} \\ \cline{2-8} 
\multicolumn{1}{|c|}{} &
  \multicolumn{1}{c|}{\multirow{2}{*}{\textbf{Nominal Value}}} &
  \multicolumn{2}{c|}{\textbf{Walking Skill}} &
  \multicolumn{2}{c|}{\textbf{Running Skill}} &
  \multicolumn{2}{c|}{\textbf{Jumping Skill}} \\ \cline{3-8} 
\multicolumn{1}{|c|}{} &
  \multicolumn{1}{c|}{} &
  \multicolumn{1}{c|}{\textbf{Stage 1}} &
  \multicolumn{1}{c|}{\textbf{Stage 2, 3}} &
  \multicolumn{1}{c|}{\textbf{Stage 1}} &
  \multicolumn{1}{c|}{\textbf{Stage 2, 3}} &
  \multicolumn{1}{c|}{\textbf{Stage 1}} &
  \textbf{Stage 2, 3} \\ \hline
\multicolumn{8}{|c|}{\textbf{Motion Tracking}} \\ \hline
\multicolumn{1}{|c|}{Motion position: $r(\mathbf{q}_m, \mathbf{q}^r_m(t))$} &
  \multicolumn{1}{c|}{15} &
  \multicolumn{1}{c|}{} &
  \multicolumn{1}{c|}{} &
  \multicolumn{1}{c|}{} &
  \multicolumn{1}{c|}{} &
  \multicolumn{1}{c|}{} &
  -7.5 \\ \hline
\multicolumn{1}{|c|}{Pelvis height: $r(q_z, q^r_z(t)+\delta_z)$} &
  \multicolumn{1}{c|}{5} &
  \multicolumn{1}{c|}{} &
  \multicolumn{1}{c|}{} &
  \multicolumn{1}{c|}{} &
  \multicolumn{1}{c|}{} &
  \multicolumn{1}{c|}{} &
  -2 \\ \hline
\multicolumn{1}{|c|}{Foot height: $r(\mathbf{e}_z, \mathbf{e}^r_z(t)+\delta_z)$} &
  \multicolumn{1}{c|}{10} &
  \multicolumn{1}{c|}{-7} &
  \multicolumn{1}{c|}{-7} &
  \multicolumn{1}{c|}{} &
  \multicolumn{1}{c|}{} &
  \multicolumn{1}{c|}{} &
  {} \\ \hline
\multicolumn{8}{|c|}{\textbf{Task Completion}} \\ \hline
\multicolumn{1}{|c|}{Pelvis position: $r(q_{x,y}, q^d_{x,y})$} &
  \multicolumn{1}{c|}{7.5} &
  \multicolumn{1}{c|}{-1.5} &
  \multicolumn{1}{c|}{-1.5} &
  \multicolumn{1}{c|}{} &
  \multicolumn{1}{c|}{} &
  \multicolumn{1}{c|}{+5.5} &
  +7.5 \\ \hline
\multicolumn{1}{|c|}{Pelvis velocity: $r(\dot{q}_{x,y}, \dot{q}^d_{x,y})$} &
  \multicolumn{1}{c|}{15} &
  \multicolumn{1}{c|}{} &
  \multicolumn{1}{c|}{} &
  \multicolumn{1}{c|}{} &
  \multicolumn{1}{c|}{} &
  \multicolumn{1}{c|}{-15} &
  -2.5 \\ \hline
\multicolumn{1}{|c|}{Pelvis orientation: $r(\cos(q_{\phi,\theta,\psi}, [0,0,q^d_\psi]), 1)$} &
  \multicolumn{1}{c|}{10} &
  \multicolumn{1}{c|}{-2.5} &
  \multicolumn{1}{c|}{+2.5} &
  \multicolumn{1}{c|}{-5} &
  \multicolumn{1}{c|}{} &
  \multicolumn{1}{c|}{+2.5} &
  {} \\ \hline
\multicolumn{1}{|c|}{Pelvis angular rate: $r(\dot{q}_{\phi,\theta,\psi}, [0,0,\dot{q}^d_\psi])$} &
  \multicolumn{1}{c|}{3} &
  \multicolumn{1}{c|}{} &
  \multicolumn{1}{c|}{} &
  \multicolumn{1}{c|}{} &
  \multicolumn{1}{c|}{+4.5} &
  \multicolumn{1}{c|}{} &
  +7 \\ \hline
\multicolumn{8}{|c|}{\textbf{Smoothing}} \\ \hline
\multicolumn{1}{|c|}{Foot Impact: $r(F_z, 0)$} &
  \multicolumn{1}{c|}{10} &
  \multicolumn{1}{c|}{-7} &
  \multicolumn{1}{c|}{} &
  \multicolumn{1}{c|}{} &
  \multicolumn{1}{c|}{} &
  \multicolumn{1}{c|}{-5} &
  {} \\ \hline
\multicolumn{1}{|c|}{Torque: $r(\bm{\tau}, 0)$} &
  \multicolumn{1}{c|}{3} &
  \multicolumn{1}{c|}{} &
  \multicolumn{1}{c|}{} &
  \multicolumn{1}{c|}{} &
  \multicolumn{1}{c|}{} &
  \multicolumn{1}{c|}{} &
  {} \\ \hline
\multicolumn{1}{|c|}{Motor velocity: $r(\dot{\mathbf{q}}_m, 0)$} &
  \multicolumn{1}{c|}{0} &
  \multicolumn{1}{c|}{} &
  \multicolumn{1}{c|}{} &
  \multicolumn{1}{c|}{} &
  \multicolumn{1}{c|}{+3} &
  \multicolumn{1}{c|}{} &
  {} \\ \hline
\multicolumn{1}{|c|}{Joint acceleration: $r(\ddot{\mathbf{q}}, 0)$} &
  \multicolumn{1}{c|}{3} &
  \multicolumn{1}{c|}{} &
  \multicolumn{1}{c|}{} &
  \multicolumn{1}{c|}{} &
  \multicolumn{1}{c|}{} &
  \multicolumn{1}{c|}{} &
  -3 \\ \hline
\multicolumn{1}{|c|}{Change of action: $r(\mathbf{a}_t, \mathbf{a}_{t+1})$} &
  \multicolumn{1}{c|}{3} &
  \multicolumn{1}{c|}{} &
  \multicolumn{1}{c|}{} &
  \multicolumn{1}{c|}{+2} &
  \multicolumn{1}{c|}{+2} &
  \multicolumn{1}{c|}{-3} &
  +7 \\ \hline
\end{tabular}
\end{table*}

\paragraph{Motion Capture}
The reference motion for the \emph{running} skill is derived from motion capture data collected from a human actor~\citep{sfudatabase}. 
We retargeted the original human motion to the Cassie's morphology using inverse kinematics. 
This process involves matching the motion of key points, such as the pelvis and toes, between the human and the robot. 
More details on the retargeting process is available in ~\cite[Sec. II-B]{li2020animated}. 
Note that we only have \textit{one single} reference motion for periodic running, with an average speed of 3 m/s and foot clearance of around 0.1 m during flight, without lateral or turning movements.
We also retarget \textit{one single} human motion of the transition from running to standing to serve as the reference motion for Cassie to transition between the two behaviors.

\paragraph{Animation}
We provide the reference motion for the \textit{jumping} skill using the animation technique. 
The robot jumping motion is directly hand-crafted in a 3D animation creation suite. 
Similar to running, we only provide a \textit{single} jumping-in-place animation for reference motion, with an apex foot height of 0.5 m, a jumping timespan $T_J=1.66$ seconds, and ending with a stance pose. 
For readers interested in creating animation for bipedal robots, please refer to~\cite{li2020animated}.

Please note that we do not perform trajectory optimization to translate the kinematically feasible reference motion from motion capture or animation to be dynamically feasible for the robot.

\subsection{Reward}\label{subsec:reward}
We now formulate the reward function $r_t$ that the agent receives at each timestep $t$ in order to encourage the robot to perform the desired locomotion skills while completing the desired tasks. 
In this work, the reward $r_t$ the agent receives is the weighted summation of several reward components $\mathbf{r}$, \textit{i.e.}, $r_t = (\mathbf{w}/||\mathbf{w}||_1)^T \mathbf{r}$ with the component vector $\mathbf{r}$ and weight vector $\mathbf{w}$ listed in Table~\ref{tab:reward}.
Each element in $\mathbf{r}$ shares the same format as:
\begin{equation}\label{eq:reward}
    r(\mathbf{u},\mathbf{v}) = \exp({-\alpha||\mathbf{u}-\mathbf{v}||_2}).
\end{equation}\noindent 
By maximizing \eqref{eq:reward}, the robot is incentivized to minimize the distance between two vectors, $\mathbf{u}$ and $\mathbf{v}$. 
In \eqref{eq:reward}, a different scaling factor $\alpha>0$ is introduced in each term to normalize units, resulting in an output range of $(0,1]$.

\subsubsection{Reward Components}
The reward $r_t$ comprises three key terms: (1) motion tracking, (2) task completion, and (3) smoothing. 
Each of these terms consists of several individual reward components, each serving specific objectives, as detailed in Table~\ref{tab:reward}. 

\paragraph{Motion tracking}
The \textit{motion tracking} term is crafted to incentivize the agent to follow the provided reference motion for a specific skill.
This objective is achieved through several components, including the motor position reward $r(\mathbf{q}_m, \mathbf{q}_m^r(t))$, the global pelvis height $r(q_z, q^r_z(t)+\delta_z)$, and the global foot height $r(\mathbf{e}_z, \mathbf{e}^r(t)+\delta_z)$ at each time step.
The addition of $\delta_z$ in the reference vertical displacement takes into account variations in terrain height, such as when the robot encounters changing terrain while running ($\delta_z\coloneqq$ the time-varying terrain height) or when it jumps to different elevations ($\delta_z\coloneqq$ the time-invariant target elevated height). 
There is some privileged environment information used in the reward, such as the robot's global height $q_z$, foot height $\mathbf{e}_z$, or terrain height $\delta_z$. 
These terms provide the robot with better knowledge of the environment and current states during training, but are excluded from the actor's observation. We will frequently see such an exploitation of privileged information in the reward. 

\paragraph{Task completion}
We incorporate a \textit{task completion} term into the reward to ensure that the robot accomplishes the assigned tasks using the acquired locomotion skill. 
Within this term, we motivate the robot to align its movement with the desired velocity and turning rate by including $r(\dot{q}_{x,y}, \dot{q}^d_{x,y})$ and $r(\dot{q}_{\phi,\theta,\psi}, [0,0,\dot{q}^d_{\psi}])$, respectively. 
Additionally, we introduce global pose tracking components $r(q_{x,y}, q^d_{x,y})$ and $r(\cos(q_{\phi,\theta,\psi}-[0,0,q^d_{\psi}]),1)$ in the reward.
Note that for the orientation tracking term, we opt to match $1$ with the $\cos$ of the orientation error within the range of $[-\pi, \pi]$ to prevent singularities when the robot undergoes a transition between $-\pi$ and $\pi$. 
Furthermore, as this work does not consider changes in robot roll and pitch ($q_{\phi, \theta}$), we set $q^d_{\phi, \theta}$ and its rate of change $\dot{q}^d_{\phi, \theta}$ to zero, contributing to the stabilization of the robot's pelvis.
For periodic walking and running skills, the desired velocities $\dot{q}^d_{x,y,\psi}$ are initially provided, and the position terms $q_{x,y,\psi}$ are calculated by integrating the corresponding commands over time. 
Conversely, for aperiodic jumping skills, the desired (time-invariant) landing targets $q_{x,y,\psi}$ are specified first, and average velocity terms are introduced to shape the sparse position reward as $\dot{q}^d_{x,y,\psi}=q^d_{x,y,\psi}/T_J$ where $T_J$ is the jumping timespan.

\paragraph{Smoothing}
Finally, we also add a \textit{smoothing} term to discourage the robot from learning jerky behaviors.
We incentivize the robot to reduce impact forces through $r(F_z,0)$, reduce energy consumption via $r(\bm{\tau},0)$, produce smooth motions by minimizing motor velocities $r(\dot{\mathbf{q}}_m,0)$, damp out joint acceleration $r(\ddot{\mathbf{q}},0)$, and regulate changes in the actions $r(\mathbf{a}_t, \mathbf{a}_{t+1})$.

\subsubsection{Reward Weights}
By employing different choices of weights $\mathbf{w}$, we can emphasize certain reward terms as more critical for the robot's performance while diminishing the significance of others, thereby influencing the robot's acquired maneuvers. 
In this work, despite covering different dynamic locomotion skills and including several training stages for each skill, we demonstrate that the reward components and their associated weight values align with a unified choice (\textit{i.e.}, the nominal value in Table~\ref{tab:reward}) applicable to different skills and stages. 
This homogenization can be achieved with some adjustments in accordance with general principles governing the tuning of specific weight, as detailed in Table~\ref{tab:reward} and introduced below. 

\paragraph{Weights across different stages}
In Stage 1, which involves the initial training of a specific locomotion skill from scratch, our emphasis is placed on training the robot to master the desired skill. As a result, the weight of the \textit{motion tracking} term takes precedence in the reward, encouraging the robot to closely mimic the reference motion.
As the robot becomes proficient in the desired skill, upon entering Stage 2, where the commands are randomized, we can adjust the weight of the \textit{task completion} term to outweigh other terms. 
This adjustment serves to motivate the robot to accomplish different tasks, especially the ones beyond the reference motion, such as executing turning motions in addition to walking, various running and turning speeds that extend beyond the single running reference motion, and various landing targets beyond the single jumping-in-place animation.
To prevent the robot from adopting overly conservative behavior, especially during initial training, the weight of the \textit{smoothing} term remains relatively low. 
As the robot solidifies its locomotion skill, this term can be gradually increased to refine the robot's movements.
But this term is always the least across multiple stages.

\paragraph{Weights across different skills}
As detailed in Table~\ref{tab:reward}, the variations in the weights among different locomotion skills are generally not substantial, with the primary distinctions arising from the presence of the flight phase in the skill. 
Notably, in skills such as running and jumping, where a significant flight phase is involved, the \textit{motion tracking} term for foot height $r(\mathbf{e}_z, \mathbf{e}^r_z(t) + \delta_z)$ carries a larger weight compared to the walking skill.
Furthermore, to encourage the robot to tackle the challenges caused by the flight phase while deviating from the reference motion to explore diverse tasks, we assign higher weights to the \textit{task completion} term. 
The \textit{smoothing} term largely remains consistent across skills, with some adjustments like the change of action term $r(\mathbf{a}_t, \mathbf{a}_{t+1}$) to strengthen the smoothing of aggressive movements for running and jumping.

\subsection{Episode Design}~\label{subsec:episode_design}

\subsubsection{Unified Approach}
Across all of the diverse locomotion skills and training stages developed in this study, the episode design is consistent and unified. 
The episode duration is set to 2500 timesteps, corresponding to a total timespan of 76 seconds. 
In Stage 2, where variable tasks are introduced, we randomize the command after random time intervals, ranging from 1 second to 15 seconds. 
The specific ranges for uniformly randomized commands for each task are detailed in Table~\ref{tab:command} in Appendix~\ref{appendix:command}. 
Note that we employ a long horizon in an episode to enable the robot to explore more scenarios of transitions among different tasks.

One exception is made for the episode length in the first stage of aperiodic tasks like jumping. 
For such cases, the episode length is adjusted to cover a complete trajectory (\textit{e.g.}, 1.66 seconds for jumping motion), with a significant extension to enable the robot to learn to maintain the last standing pose after landing. 
In this work, the Stage 1 training of jumping employs 750 timesteps (22 seconds).

\subsubsection{Early Termination Conditions}
When training dynamic locomotion skills on bipedal robots, relying solely on reward design could be insufficient. 
The robot might still exhibit undesirable behavior despite achieving suboptimal return, such as maintaining a stance pose without engaging in locomotion because the robot can easily attain part of the reward in this way. 
To address this, in addition to the standard termination conditions, such as the robot falling over ($q_z<0.55$m) or the tarsus joints $q_6^{L/R}$ hitting the ground, we incorporate two additional termination conditions among all skills and stages, as elaborated below. 

\paragraph{Foot height tracking tolerance}
The episode will terminate early if the deviation between the robot's foot height and its reference motion exceeds a threshold $E_e$, \textit{i.e.}, if $|\mathbf{e}_z - \mathbf{e}^r_z(t) - \delta_z| > E_e$. 
This condition is empirically found to be particularly effective in promoting the development of locomotion skills involving flight phases, like running and jumping. 
Without this condition for \textit{motion tracking}, especially during the initial stages of training, the robot tends to remain mostly on the ground.
For skills lacking a flight phase, such as walking, this condition can still be incorporated, although it may not be necessary.

\paragraph{Task completion tolerance}
We underscore the importance of \textit{task completion} to the robot by introducing a condition based on the robot's deviation from the commanded base position and orientation, $|q_{x,y,\psi} - q^d_{x,y,\psi}| > E_t$, with $E_t$ being the acceptable tracking error threshold.
For walking and running, we consistently check this condition to encourage the robot to diminish accumulated tracking errors. 
In the case of jumping, this condition is evaluated after the robot's landing, stimulating the robot to jump to the specified target.

\paragraph{Tolerance across different stages}
The error thresholds $E_e$ and $E_t$ may require adjustments as the training progresses, depending on the robot's proficiency in the desired locomotion skill. 
For instance, the foot height tracking error threshold $E_e$ may start with a tight constraint in the initial stage (Stage 1) and progressively relax in subsequent stages.
Ideally, this condition can be completely removed in the end.
This strategy provides the robot more flexibility in exploring diverse maneuvers in the later stages, surpassing the capabilities of the provided reference motion. 
In contrast, the tracking error threshold $E_t$ can be gradually reduced during training as the robot becomes more skilled in locomotion and can place a stronger emphasis on task completion.
Despite the diversity of skills, we observe a consistent trend in such tolerance adjustments across the multiple training stages.

\subsection{Dynamics Randomization}\label{subsec:domain_rand}
\begin{table}[]
\scriptsize
\caption{The range of dynamics randomization. Introducing simulated external perturbations to the robot's base or incorporating variable terrain is optional but only recommended after the robot has learned to handle general dynamics randomization. However, for highly dynamic skills like aperiodic jumping, external perturbations may not enhance robustness and could stop the robot from learning meaningful maneuvers.}
\centering
\label{tab:randomization}
\begin{tabular}{cc}
\hline
\textbf{Parameters}              & \textbf{Range}                    \\ \hline
\multicolumn{2}{c}{\textbf{Dynamics Randomization (General)}}        \\ \hline
Ground Friction Coefficient             & {[}0.3, 3.0{]}                    \\
Joint Damping Ratio              & {[}0.3, 4.0{]} Nms/rad            \\
Spring Stiffness                 & {[}0.8, 1.2{]} $\times$ default   \\
Link Mass                        & {[}0.5, 1.5{]} $\times$ default   \\
Link Inertia                     & {[}0.7, 1.3{]} $\times$ default   \\
Pelvis (Root) CoM Position   & {[}-0.1, 0.1{]} m in $q_{x,y,z}$  \\
Other Link CoM Position      & {[}-0.05, 0.05{]} m + default     \\
Motor PD Gains                   & {[}0.7, 1.3{]} $\times$ default   \\
Motor Position Noise Mean        & {[}-0.002, 0.002{]} rad           \\
Motor Velocity Noise Mean        & {[}-0.01, 0.01{]} rad/s           \\
Gyro Rotation Noise              & {[}-0.002, 0.002{]} rad           \\
Linear Velocity Estimation Error & {[}-0.04, 0.04{]} m/s             \\
Communication Delay              & {[}0, 0.025{]} s                \\ \hline
\multicolumn{2}{c}{\textbf{External Perturbation (Optional)}}        \\ \hline
Force \& Torque                  & {[}-20, 20{]} N \& {[}-5, 5{]} Nm \\
Elapsed Time Interval (Walking)            & {[}0.1, 3.0{]} s                \\ 
Elapsed Time Interval (Running)            & {[}0.1, 1.0{]} s                \\ \hline
\multicolumn{2}{c}{\textbf{Randomized Terrain (Optional)}}           \\ \hline
Terrain Type                     & Waved, Slopes, Stairs, Steps      \\ \hline
\end{tabular}
\end{table}
As illustrated in Fig.~\ref{fig:multi_stage_training}, in training Stage 3, we introduce randomized dynamics parameters in the simulation environment to train a policy that can stay robust and generalize the acquired locomotion skills to deal with the uncertainty in both \textit{dynamics modeling} and \textit{measurements}. 
The objective of this training is to enable successful transfer from simulation to real-world scenarios where the dynamics parameters are uncertain.
At each episode, dynamics parameters, as detailed in Table~\ref{tab:randomization}, are sampled from their respective uniform distributions. 

Specifically, to address \textit{modeling uncertainty}, we introduce extensive randomization of modeling parameters, including ground friction coefficient, robot joint damping ratio, and mass, inertia, and Center-of-Mass (CoM) position of each link. 
In the case of Cassie, which features passive joints connected by leaf springs, the introduction of $\pm20\%$ randomized stiffness for the leaf springs has been important for successful sim-to-real transfer, particularly for skills like running and jumping that involve substantial compression of the springs.
We also incorporate randomization into the PD gains of the joint-level PD controllers, adding a range of $\pm30\%$ deviation from their default values. 
This randomization is applied independently to each joint-level PD controller. 
This stimulates the diversity in motor responses that can mimic the effects of uncertain motor dynamics seen in real-world scenarios, including motor aging and degradation, which is found effective in facilitating the sim-to-real transfer.

To tackle \textit{measurement uncertainty}, we add a simulated noise to the observable states $\mathbf{o}_t$. 
The noise is modeled as a normal distribution with the mean uniformly sampled from the range specified in Table~\ref{tab:randomization}. 
Please note that, given the reliable sensors like joint encoders and IMU on Cassie's hardware, the range of measurement noise we applied is small. 
However, for robots equipped with lower-quality sensors, a wider simulated noise range may be necessary. 
Additionally, we simulate a communication delay, caused by a zero-order hold, between the policy and the robot's real-time computer, which influences the timing of sending actions and receiving observations. 
For highly dynamic motions, such as jumping and running, the delay plays a significant role in the sim-to-real gap, and the ability to tackle delays during real-time control is critical for stability. 

The randomization strategy mentioned above is applied consistently across various locomotion skills. 
Additionally, we have also investigated the use of other sources of randomization during training. They are randomized perturbation and randomized terrains as detailed below.

\paragraph{Use of Randomized Perturbation}
In our study, we explored the use of randomized external perturbation wrenches applied to the robot's pelvis, hypothesizing that it could create more diverse training scenarios and enhance robustness by deviating the robot from its nominal trajectories. This perturbation (typically being impulse), incorporating forces and torques applied for random durations as specified in Table~\ref{tab:randomization}, can be simulated during training.
However, we find that while this approach can increase the robustness of policies in real-world applications, it introduces complex hyperparameters that are challenging to choose and complicates the training process. For walking, frequent perturbations led to an overemphasis on perturbed scenarios, impairing the robot's performance in normal walking gaits. 
In running, longer perturbations hindered the robot's ability to learn effective gaits. 
As a result, we employed different elapsed time intervals for walking and running, as detailed in Table~\ref{tab:randomization}.
Moreover, in the case of jumping skills or transitions from locomotion to standing, these perturbations proved even more problematic, preventing the robot from acquiring meaningful jumping or standing abilities. Therefore, we chose to exclude external perturbation from the training for both jumping and transition-to-standing skills.

\paragraph{Use of Randomized Terrain}
If the control policy aims to enable the robot to traverse uneven terrain, it is essential to simulate terrain changes during training. 
We developed an algorithm to randomize various types of terrains using parameterized height maps including wave terrain (characterized by sine functions), sloped terrain, monotonic stairs, and random steps, as listed in Table~\ref{tab:randomization}. 
It's important to note that our control policy doesn't incorporate vision, so the robot has to adapt to terrain changes based on its I/O history, making it a more challenging problem.
Terrain randomization is recommended only after the robot has proficiently trained in other dynamics randomization to avoid excessive learning complexity. 
In this study, we introduced terrain randomization for running skills for example.

\subsection{Training Details}
We develop the simulation of Cassie within the aforementioned environments using MuJoCo based on~\citep{todorov2012mujoco,cassiemj}. 
The training of all control policies is carried out using Proximal Policy Optimization (PPO) developed by~\citep{schulman2017proximal} in simulation.
The control policy (actor) is as described in Sec.~\ref{sec:controller}, with a value function represented by a 2-layered MLP that has access to ground truth observations.
Given the varying complexities of different training stages and skills, the number of training iterations differs across stages and skills. 
The numbers of iterations and other hyperparameters are provided in Table~\ref{tab:num_iter} and Table~\ref{tab:hyperparam_ppo}, respectively, in Appendix~\ref{appendix:hyperparameters}. 

Up to this point, design details of the proposed RL framework for learning versatile, robust, and dynamic bipedal locomotion skills have been introduced. 
We will next proceed to validate several key design components. 

\section{Advantages of Policy Architecture}\label{sec:policy_structure}
In this section, we first evaluate the advantages of the proposed bipedal locomotion control framework in Sec.~\ref{sec:controller} through an extensive ablation study of its design choices. 
This evaluation unfolds from two key perspectives: (1) the learning performance in challenging simulation environments with extensively randomized dynamic parameters, and (2) the control performance during the sim-to-real transfer on a real robot, without any tuning. 
The simulation offers a controlled setting to test the capacity of the control architectures in learning control strategies for large changes in the system dynamics, while real-world experiments allow us to evaluate the control efficacy on the fixed but uncertain dynamics of the robot's hardware. 
As we will demonstrate, our proposed dual-history control architecture and training system result exhibits the best learning performance and sim-to-real transfer.

\subsection{Baselines}

\begin{figure}[t]
    \centering
    \includegraphics[width=\linewidth]{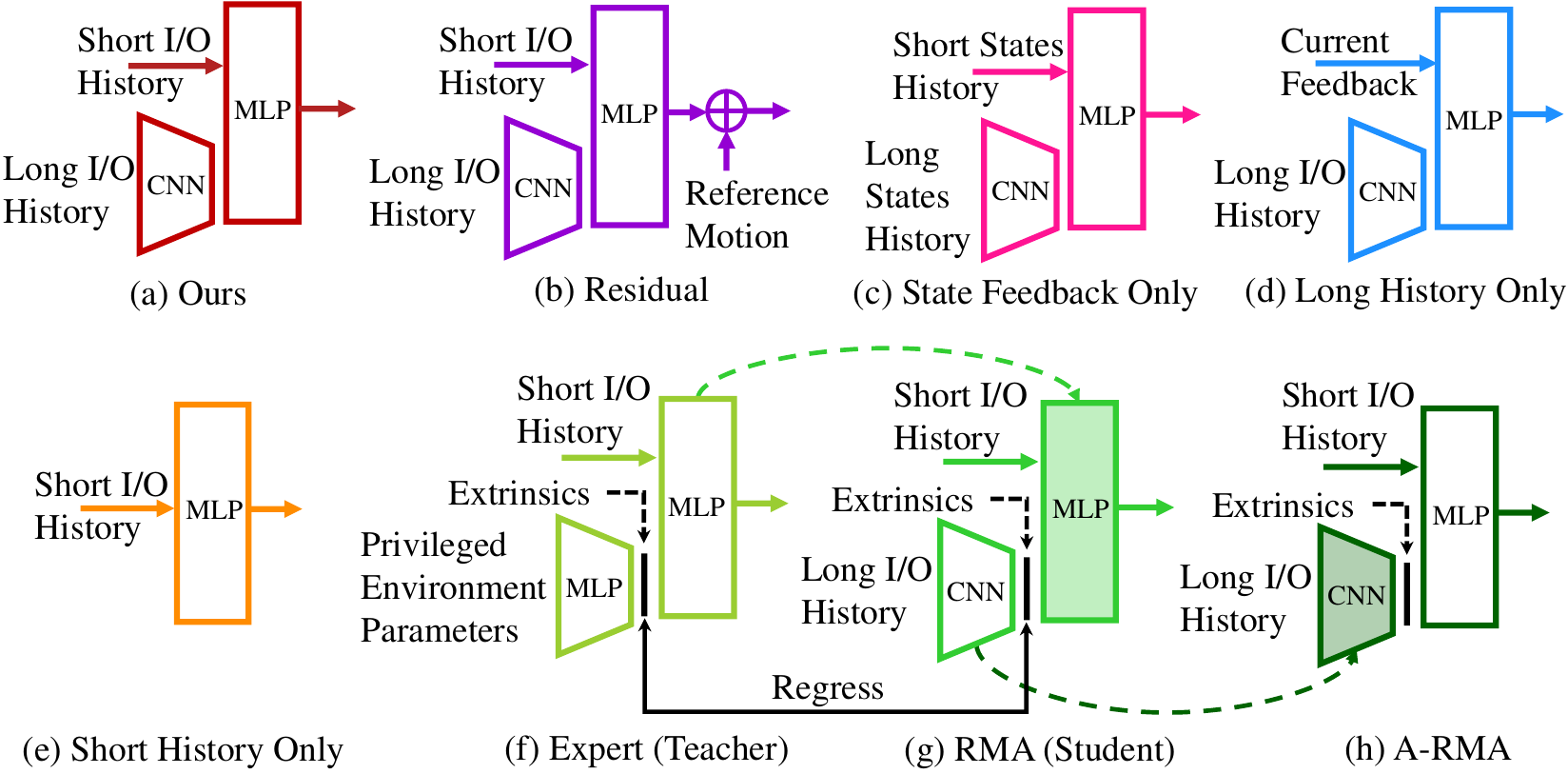}
    \caption{Illustration of our proposed and various baselines for RL-based control policy architectures for bipedal robot locomotion. Fig.~\ref{fig:benchmark}a, \textbf{Ours} integrates both short and long-term I/O histories, with the base MLP and long history encoder jointly trained to specify motor positions. Fig.~\ref{fig:benchmark}b, the \textbf{Residual} approach aligns with our architecture but adds a residual term to the reference motor position. Fig.~\ref{fig:benchmark}c, the \textbf{State Feedback Only} baseline uses our model structure but relies solely on robot's states history, excluding input history. Fig.~\ref{fig:benchmark}d, the \textbf{Long History Only} approach depends on long I/O history without using short I/O history, while the \textbf{Short History Only} approach (Fig.~\ref{fig:benchmark}e) focuses only on short-term I/O history, excluding the CNN encoder. The \textbf{RMA/Teacher-Student} method utilizes a two-phase policy distillation, with an expert (teacher) policy (Fig.~\ref{fig:benchmark}f) guiding the training of an RMA (student) policy (Fig. \ref{fig:benchmark}g), which can be improved by \textbf{A-RMA} (Fig.~\ref{fig:benchmark}h) which introduces an additional phase where the base MLP is finetuned while keeping the long I/O history encoder's parameters fixed. Notably, all expert, RMA, and A-RMA policies in this study incorporate short I/O histories into the base MLP, which is a new modification in this work to enable equitable comparison with ours. All of these architectures have the command and reference motion as input to the base MLP, as detailed in Fig.~\ref{fig:controller} and omitted for brevity.}
    \label{fig:benchmark}
\end{figure}

\begin{figure*}[!htp]
    \centering
    \includegraphics[width=\linewidth]{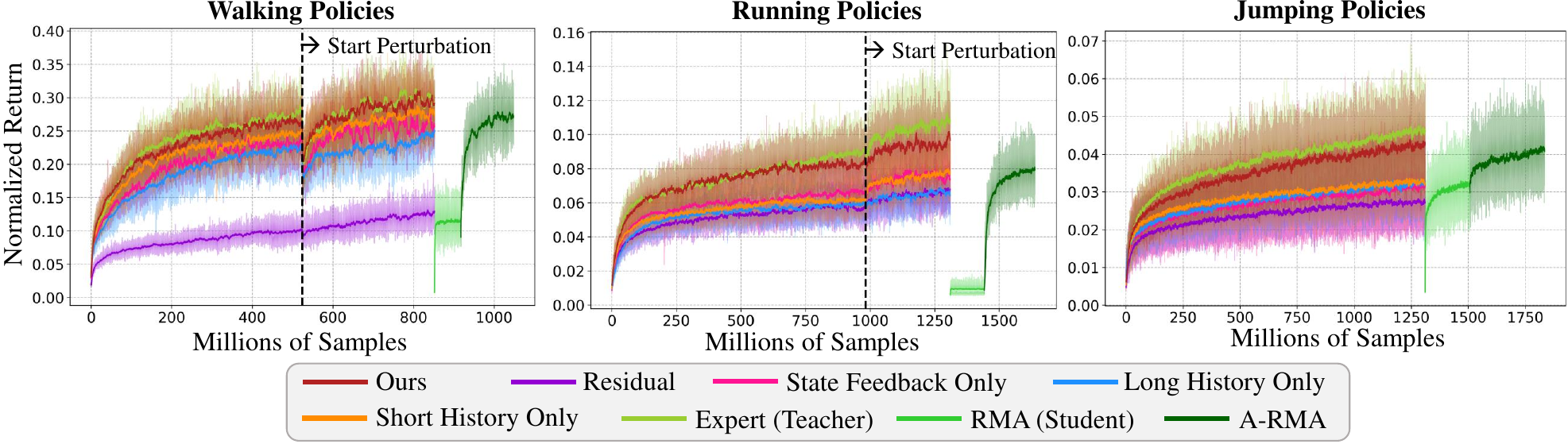}
    \caption{The learning performance using different policy structure designs illustrated in Fig.~\ref{fig:benchmark}. It is assessed during Stage 3 training, which incorporates both task and dynamics randomization. These curves represent the average normalized episodic return across 3 distinct policy trainings from different random seeds, with shaded regions indicating the range between minimum and maximum returns. Note that there is no perturbation training for the jumping policy as it prevents the robot from learning the dynamic jumping skill. Our proposed method consistently outperformed other baselines across different skills. Notably, our policy's performance is comparable to that of the expert policy, which has access to privileged information but is not deployable in the real world. In contrast, the residual method shows the worst return. Using only a long history does not offer a clear benefit compared to a policy relying solely on a short history. In fact, the short history only policy even outperforms the long history only approach. Additionally, even with a dual-history approach like ours, omitting the robot's input history (only using state feedback) results in no improvement over short I/O history only. The student or RMA methods exhibit significant regression loss in bipedal locomotion control; particularly in dynamic skills like running, RMA fails to learn. This suggests the necessity of the A-RMA stage for further training, though it requires considerably more training samples and yields slightly lower returns compared to our method. }
    \label{fig:dynrand_three}
\end{figure*}

We compare the proposed controller structure and other baselines listed below and illustrated in Fig.~\ref{fig:benchmark}. Those choices of baselines include ablations on several design choices proposed in this work: (1) choices of action space, (2) history of the robot's I/O or state feedback only, (3) history length (long or short history), (4) dual-history or long history only, (5) end-to-end training or policy distillation. 
The models used in the experiment include:
\begin{itemize}[leftmargin=9pt]
    \small
    \item \textbf{Ours} (Fig.~\ref{fig:benchmark}a): 
    as detailed in Fig.~\ref{fig:controller}, our architecture leverages both short-term and long-term I/O history, with the long-term history encoded by a CNN, while the short-term history is directly fed into the base MLP. The CNN and MLP are trained jointly and the policy outputs specify the desired motor positions.
    \item \textbf{Residual} (Fig.~\ref{fig:benchmark}b): 
    this architecture aligns with our proposed one, but it produces an output that represents a residual term that is then added to the reference motor position at the current timestep, \textit{i.e.}, $q^d_m = \mathbf{a}_t + q^r_m(t)$. Such an approach is employed in prior works like~\cite{lee2020learning,xie2020learning,siekmann2020learning}. It is worth noting that the robot “knows" the current reference motor position to add by taking the reference motion as input.
    \item \textbf{State Feedback Only} (Fig.~\ref{fig:benchmark}c):
    this variant retains the same model structure and action space as ours. However, it differs in its observation by relying solely on the historical states (robot's output history), omitting the robot's input history. Such a choice, \emph{without} the use of short history, is more commonly seen in previous work, such as~\cite{lee2020learning,siekmann2020learning,siekmann2021sim,crowley2023optimizing}.
    \item \textbf{Long History Only} (Fig.~\ref{fig:benchmark}d): this policy architecture relies only on a long-term I/O history encoded by the CNN. This configuration serves as a baseline in~\cite{kumar2021rma}. 
    As suggested by~\cite{peng2018sim}, the base MLP has direct access to the robot's immediate (last-timestep) state feedback. For brevity, in this work, \emph{Long History Only} refers to the method that provides the lastest observation (state feedback) along with the long history encoder, which can be modeled using different neural network architectures.
    \item \textbf{Short History Only} (Fig.~\ref{fig:benchmark}e): this policy relies solely on short-term I/O history, excluding the long-term I/O history CNN encoder. This is used for bipedal locomotion control in~\cite{li2021reinforcement} and is more commonly seen in quadruped control like~\cite{huang2022creating,escontrela2022adversarial,feng2023genloco}.
    \item \textbf{RMA/Teacher-Student}: this architecture utilizes the policy distillation method that involves two training phases. First, an expert (teacher) policy (Fig.~\ref{fig:benchmark}f) is trained by RL that has access to privileged environment information (listed in Table~\ref{tab:randomization}). The privileged information is encoded into an $8$D extrinsics vector by an MLP encoder. \emph{This expert (or teacher) policy can only be used in simulation}. Second, the expert policy is employed to \textit{supervise} the training of an RMA (student) policy (Fig.~\ref{fig:benchmark}g). 
    The RMA policy copies the base MLP from the expert policy and only learns to leverage the long I/O history encoder to estimate the teacher's extrinsic vector. 
    Such a policy distillation method is used in~\cite{lee2020learning,kumar2021rma} and widely adopted in quadrupedal locomotion control.
    \item \textbf{A-RMA} (Fig.~\ref{fig:benchmark}h): 
    after the RMA training, an additional training phase is introduced. In this phase, the long I/O history encoder's parameters remain fixed, while the base MLP is updated again through RL, as introduced by~\cite{kumar2022adapting}. 
    \emph{Notably, during the implementation, all expert, RMA, and A-RMA policies incorporate short I/O histories to the base MLP in this study, which is a new modification in this work to enable an equitable comparison with ours.}
\end{itemize}

For each locomotion skill studied in this work (walking, running, and jumping), using our proposed method and baseline methods, we obtained 3 policies trained by the identical multi-stage training framework introduced in Sec.~\ref{sec:training}. These policies were obtained from the same choice of hyperparameters, but from different random seeds. This was done for each of these policy architectures. 
In other words, $3\times3\times8=72$ different control policies are obtained, representing 3 locomotion skills, 3 random seeds, 8 policy architectures, and evaluated below.

\subsection{Learning Performance}

We begin our evaluation by benchmarking our method against various baselines in simulation by examining their respective learning curves. 
Our focus lies on the training performance during Stage 3, as shown in Fig.~\ref{fig:dynrand_three}, the most challenging stage that involves both randomized tasks and dynamics parameters. 
This particular emphasis on Stage 3 also stems from its critical role in sim-to-real transfer, a pivotal concern within the scope of controlling real bipedal robots in our study. 

\subsubsection{Benchmark Analysis}
As depicted in Fig.~\ref{fig:dynrand_three}, the learning performance remains uniform across various locomotion skills, as outlined and analyzed below. 

\paragraph{Choices of Action}
We observe that when the policy produces a residual term (purple curves) instead of directly specifying the desired motor position, the learning performance consistently deteriorates across various locomotion skills. 
Although the added reference motion to the action might accelerate the learning of the desired skill initially, this added reference motion could also introduce additional movements to the robot. 
Consequently, the policy expends more effort correcting the added movements rather than effectively controlling the robot, which becomes more problematic when the robot explores maneuvers beyond the reference motion. 
Hence, we recommend readers reconsider the use of residual learning in the context of locomotion control, regardless of whether the added action is optimized for dynamic feasibility (as in walking) or limited to kinematic feasibility (as in running or jumping).

We also consider another action space option: joint-level torque. RL has been used to learn torque control for quadrupedal and bipedal robots as demonstrated in~\cite{chen2022learning} and \cite{kim2023torque} respectively. However, torque control's high update frequency requirements, such as 250 Hz in~\cite{kim2023torque}, restrict its ability to utilize extensive robot I/O history. Prior methods using torque control have thus been limited to a single timestep of robot feedback. Later, we will demonstrate that a longer I/O history, like 2 seconds, could improve the adaptivity of RL-based controllers, but recording and learning with such a long history is computationally expensive for high-frequency torque controllers, as the action (robot's input) history needs to update at the same frequency as the controller. Therefore, our benchmark focuses on policies that can operate effectively at lower frequencies.

\begin{figure*}[!htp]
    \centering
    \includegraphics[width=\linewidth]{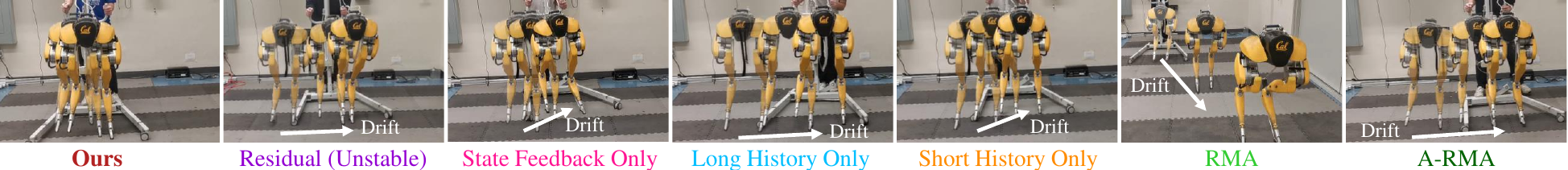}
    \caption{Snapshots from experiments of the bipedal robot Cassie controlled by different policies trained by different methods illustrated in Fig.~\ref{fig:benchmark}. The test is to control the robot to walk in place, \textit{i.e.}, $(\dot{q}^d_x, \dot{q}^d_y, q^d_{\psi}) = \mathbf{0}$. Each snapshot compiles three frames taken at the \emph{same} 1$^{\text{st}}$, 4$^{\text{th}}$, 7$^{\text{th}}$ second after initialization. Our method exhibits minor drift (the robot does not leave its initial place as demonstrated in the figure), outperforming other approaches, which all result in notable sagittal and/or lateral drifts (marked by the white arrows). Furthermore, the residual policy failed to control the robot to maintain a stable gait. This benchmark underscores the advantages of our proposed architecture in adapting to the real robot's dynamics for better tracking performance after zero-shot transfer. These results are consistent over different policies trained from different random seeds, which are reported in Fig.~\ref{fig:bm_errobar}.}
    \label{fig:bm_realworld}
\end{figure*}

\begin{figure*}[!htp]
    \centering
    \begin{subfigure}{0.49\linewidth}
      \centering
      \includegraphics[width=\linewidth]{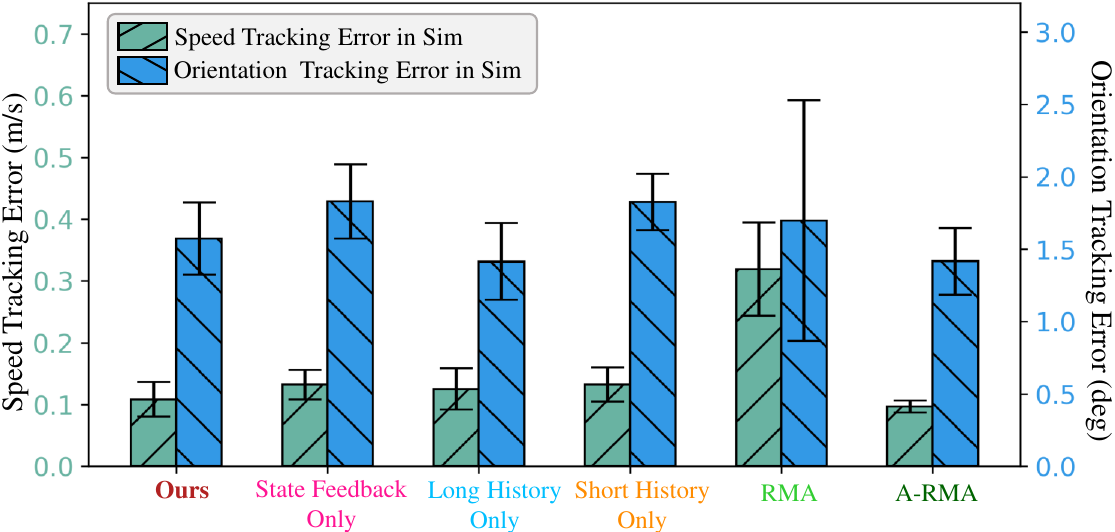}
      \caption{The bar chart of tracking performance (zero-speed) using the \textit{same} policies tested in the simulation.}
      \label{subfig:bm_errorbar_sim}
    \end{subfigure}    
    \begin{subfigure}{0.49\linewidth}
      \centering
      \includegraphics[width=\linewidth]{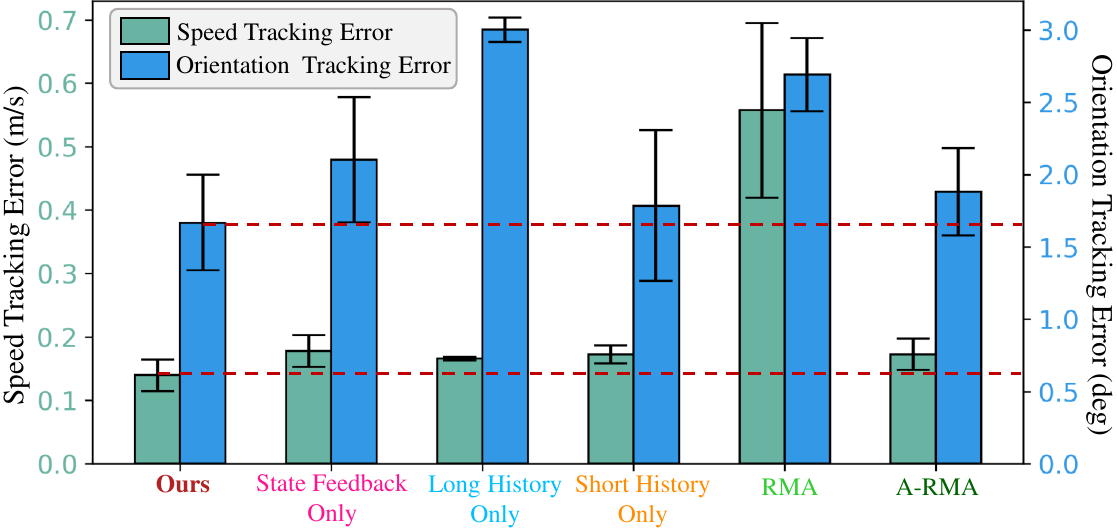}
      \caption{The bar chart of tracking performance (zero-speed) using the obtained walking policies tested on the robot hardware. }
      \label{subfig:bm_errorbar_real}
    \end{subfigure}    
    \caption{The bar chart of the tracking performance using different policies obtained by different methods (Fig.~\ref{fig:benchmark}) in the in-place walking experiments in the simulation (Fig.~\ref{subfig:bm_errorbar_sim}) and the real world (Fig.~\ref{subfig:bm_errorbar_real}). The speed tracking error (Mean Absolute Error, MAE, of $||\dot{q}_{x,y}||_2$) and orientation tracking error (MAE of $q_{\phi,\theta,\psi}$) using the policies trained from \emph{different random seeds} but same method are recorded. They are shown in green and blue bars, respectively. The bar height is the average tracking error over three random seeds and the error bar represents the standard deviation among these three tests. We exclude the residual approach as it can not maintain a stable gait. Training from different random seeds, while all methods perform similarly well in the simulation, when transferring the same policies to the hardware, our method consistently demonstrates the minimal speed tracking error and orientation tracking error over other methods. This underscores the capacity of the proposed method in terms of adapting to the robot dynamics in the real world.}
    \label{fig:bm_errobar}
\end{figure*}

\paragraph{Choices of Observation}
A comparison between our proposed method (red curves), which considers the robot's I/O history, and a baseline that relies solely on the robot's output history (pink curves) reveals an important finding: omitting the robot's input (action) history leads to a decline in learning performance, even with an identical policy architecture and training approach. 
In other words, providing only the history of the robot's state feedback in the observation is insufficient.
This comparison illustrates the critical role of utilizing both the robot's input and output history in developing a control policy through model-free RL.
This combined I/O history is crucial for the control policy to perform system identification and state estimation to infer the robot's system parameters and states, which enhances its adaptivity to uncertain dynamics and external perturbations.
This importance becomes particularly pronounced in our work, where we tackle the control of a highly nonlinear high-dimensional system (bipedal robot) using limited and noisy, delayed measurements (partial observability).

\paragraph{Long History versus Short History}
When leveraging long I/O history alone (blue curves), the learning performance fails to surpass that of the policy utilizing only short I/O history (orange curves) or others.
However, when we provide the base MLP with direct access to the short I/O history while having a long history encoder, as proposed in our approach, the resulting learning performance exhibits significant improvement. 
This enhancement is rooted in the real-time control context, where the short history becomes crucial for information about the robot's very recent I/O trajectory. 
While the long history does contain this short-term information, it can become obscured after passing through an encoder. Hence, providing the policy with direct access to \emph{explicit} short history within the base MLP complements the utilization of long history. Further discussion is presented in Sec.~\ref{subsec:dual_history_advantages}.
Additionally, in Appendix~\ref{appendix:history_length}, we delve into the effects of varying history lengths for temporal encoding. We found that increasing the history length can enhance training performance, but improvements tend to plateau with longer histories and may even hinder training by introducing redundant information for locomotion control. The 2-second history tends to perform consistently well in all the skills developed in this work.

\begin{remark}
   In Appendix~\ref{appendix:lstm_tcn}, we explore the effects of training performance using different temporal encoders such as TCN~\citep{lee2020learning} and LSTM~\citep{siekmann2020learning}. We found that a dual-history approach consistently enhances learning performance for non-recurrent policies like TCN that encode an explicit history length. However, for recurrent policies like LSTM, the dual-history approach does not significantly aid learning. Additionally, recurrent policies tend to converge to more suboptimal policies than non-recurrent ones and are sensitive to hyperparameter tuning across different MDPs (locomotion skills).
\end{remark}

It's important to clarify that our aim is not to dispute the use of a specific temporal encoding structure or the use of a long history. 
Instead, our emphasis lies in underlining the importance of incorporating short history when working with long history data, which can be encoded by many choices of structures.

\paragraph{Comparison with Policy Distillation Methods}
A comparison between our method that jointly trained the long history encoder with the base MLP (red curves) and policy distillation methods that separate the training of the base MLP and history encoder (green curves) highlights the significance of using a right training strategy. 
We observed that the RMA (student) policy exhibits significant degradation compared to the expert policy, primarily due to the unavoidable error when utilizing the robot's long history to estimate pre-selected environment parameters (encoded to the extrinsics vector). 
This decline in RMA's performance becomes particularly prominent when dealing with challenging locomotion skills, such as running, where it fails to learn.
A-RMA, which continues fine-tuning the base MLP with the frozen estimation encoder, can enhance RMA performance but it still falls slightly short when compared to our proposed method, despite having significantly more training samples. 
In the case of running, where the encoder struggles to estimate environment parameters, A-RMA essentially reproduces the result of the short history-only policy (the orange one), avoiding the use of long history.
\emph{It's worth noting that although there are other potential ways to enhance RMA learning like~\cite{fu2023deep}, our proposed method consistently demonstrates performance similar to the expert policy which is the theoretical upper bound for the student policy. Importantly, our approach is deployable in the real world, whereas the expert policy is not.}

\subsection{Case Study: In-place Walking Experiments}
We proceed to conduct a case study on the policies trained using various methods, as depicted in Fig.~\ref{fig:benchmark}, in the real world. 
We choose to assess the walking policies for controlling Cassie to sustain an in-place walking gait in the real world without any tuning or global position feedback.
This relatively straightforward scenario can provide valuable insights into the adaptivity of the trained policies. 
If the policy is unable to adapt to the dynamics of the real robot, it may result in obvious drift, even if it effectively maintains the robot's walking gait.

As shown in Fig.~\ref{fig:bm_realworld}, the results highlight a significant contrast.
Our proposed method demonstrates notably lower tracking errors and successfully maintains the robot's in-place walking, with minimal drift observed in both sagittal and lateral directions. 
Conversely, policies utilizing long history only, short history only, and dual-history with only state feedback, when trained under the same number of samples, result in substantial drift to the robot's left.
Furthermore, RMA demonstrates the most obvious sagittal shift and it walks forward using a fast speed even with a zero velocity command. 
While A-RMA reduces this sagittal drift, it still experiences considerable lateral movement. 
Moreover, the residual policy fails to maintain a stable gait on the real robot.
The corresponding video is recorded in Vid. 3 in Table~\ref{tab:video_list}.

Since we obtained three distinct policies trained from different random seeds, we test each of these policies for the same in-place walking in both the simulation using the robot nominal model and the real world, and compare the statistical results in Fig.~\ref{fig:bm_errobar}.  
The robot's speed ($||\dot{q}_{x,y}||_2$) and base orientation ($q_{\phi,\theta,\psi}$) tracking errors (Mean Absolute Error, MAE) over a 10-second test are evaluated. 
Note that, besides the turning yaw ($q_{\psi}$) that is included in the command, we also evaluate the robot's tracking error for the pelvis (base) roll $q_{\phi}$ and pitch $q_{\theta}$ angles. This is done to see which control policy can better stabilize the robot to a small base roll and pitch movement as they are supposed to be zeros. 
As recorded in Fig.~\ref{subfig:bm_errorbar_sim}, all policies trained by different methods exhibit similar good tracking performance in terms of slow speed (except RMA) and floating-base orientation when commanded to walk in place in simulation. 
However, upon transferring to real hardware (Fig.\ref{subfig:bm_errorbar_real}), other methods exhibit significant speed drift and floating base rotational tracking errors.
For example, the long history only policy shows the worst performance in stabilizing the robot's pelvis with a large floating base orientation, yet it has minimal base oscillation in simulation. A-RMA, while best at maintaining the robot's position in simulation, causes significant drift to the robot's left on real hardware. In contrast, our method achieves better sim-to-real transfer performance with minimal degradation and consistently results in better control performance in terms of command tracking and stabilizing the floating base. This underscores the advantages of our approach in adapting to the dynamics of the robot hardware. 

Consequently, we can conclude that the proposed method excels in bridging the sim-to-real gap, demonstrating better performance in effectively controlling the robot in the real-world environment.

\subsection{Summary of Results}
For achieving dynamic bipedal locomotion skills through RL, the results in this section indicate the following design choices for the control policy structure:\\
(1) Have the policy to directly specify motor-level commands rather than using a residual term;\\
(2) Utilize a history of both the robot's input and output, rather than relying solely on the robot's state feedback;\\
(3) When using the robot's I/O history, favor the inclusion of long-term history, but complement it with short-term history in the base policy for enhanced performance, which leads to a dual-history approach; and\\
(4) Train the base policy and the history encoder in an end-to-end manner, as it yields better performance, reducing training complexity and samples compared to policy distillation methods.

These design choices lead to our proposed policy structures, showcasing consistently best learning performance in simulation across various dynamic locomotion skills and delivering improved control results in the real world.
\section{Source of Adaptivity}\label{subsec:adaptivity}
\begin{figure*}[!htp]
\centering
\begin{subfigure}{0.66\linewidth}
  \centering
  \includegraphics[width=\linewidth]{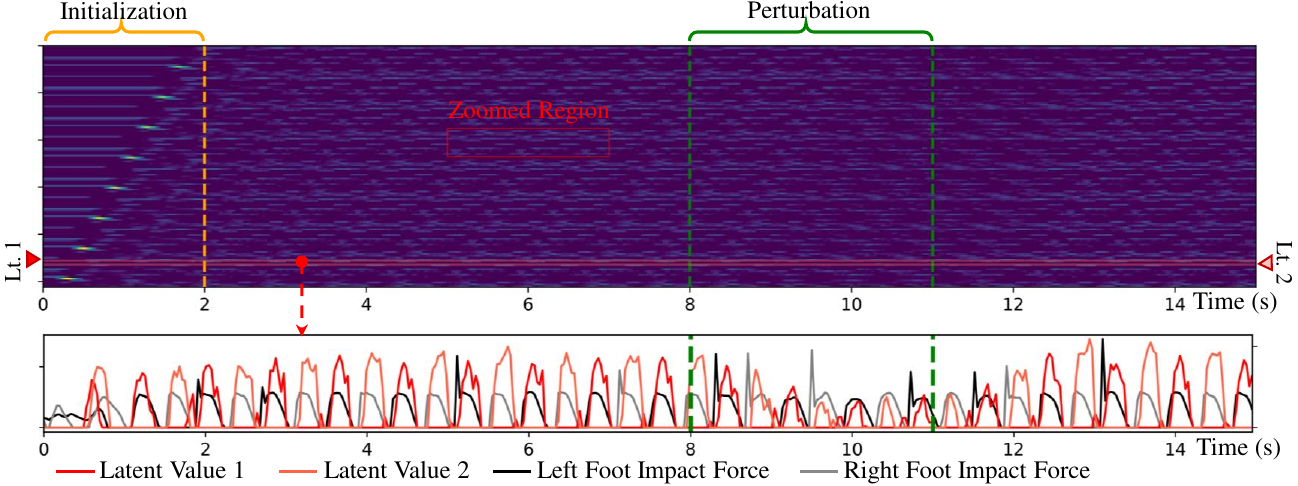}
  \caption{(Top) Recorded latent representation after long-term I/O history encoder during \underline{running}. (Bottom) Comparison of two selected dimensions (marked as red lines in the top plot) with recorded impact forces on each of the robot's feet.}
  \label{subfig:latent_run_force}
\end{subfigure}
\begin{subfigure}{0.33\linewidth}
  \centering
  \includegraphics[width=\linewidth]{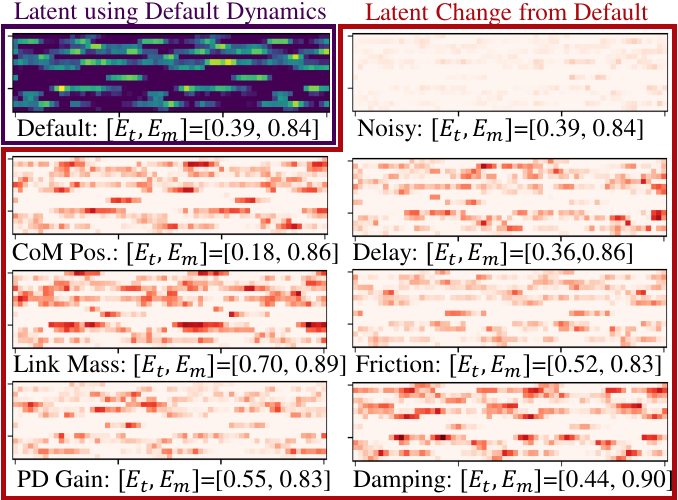}
  \caption{The blue plot shows the robot's latent representation with default dynamics parameters during running. The red plots indicate changes in the same region under different dynamics.}
  \label{subfig:latent_run_zoomed}
\end{subfigure}

\begin{subfigure}{0.66\linewidth}
  \centering
  \includegraphics[width=\linewidth]{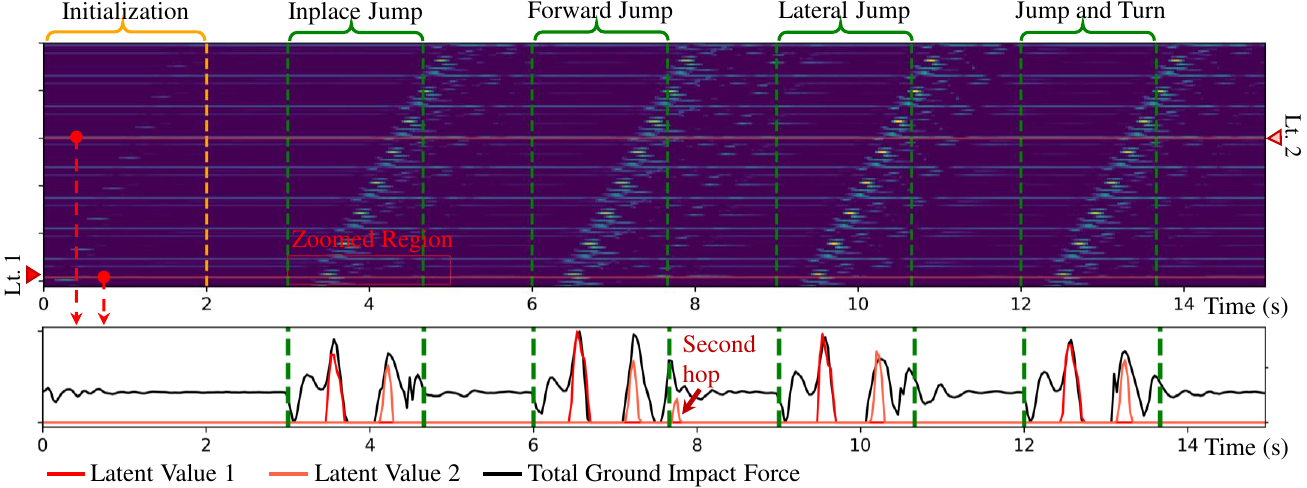}
  \caption{(Top) Recorded latent representation after long-term I/O history encoder during \underline{jumping}. (Bottom) Comparison of two selected dimensions (marked as red lines in the top plot) with recorded total impact forces on both of the robot's two feet.}
  \label{subfig:latent_jump_force}
\end{subfigure}
\begin{subfigure}{0.33\linewidth}
  \centering
  \includegraphics[width=\linewidth]{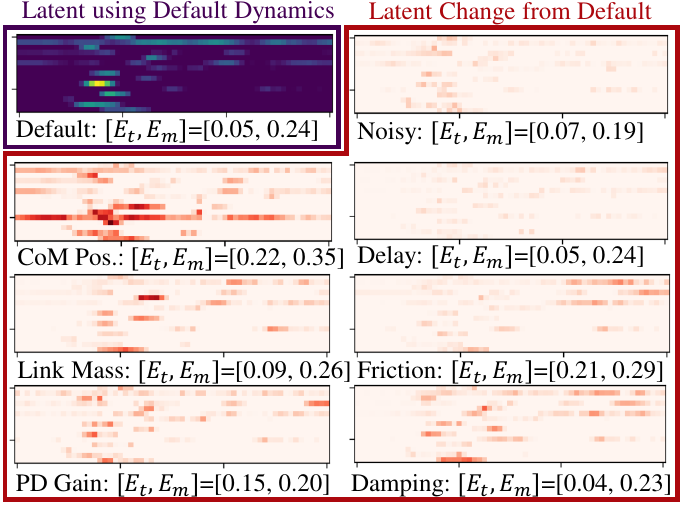}
  \caption{The blue plot shows the robot's latent representation with default dynamics parameters during jumping. The red plots indicate changes in the same region under different dynamics.}
  \label{subfig:latent_jump_zoomed}
\end{subfigure}
\caption{Adaptivity test in simulation (MuJoCo). We evaluate the latent representation profiles from the output of the long-term I/O history encoder during bipedal running (Fig.~\ref{subfig:latent_run_force}) and jumping (Fig.~\ref{subfig:latent_jump_force}) where lighter color represents a larger relative value. 
After the initialization of the history encoder (first 2 seconds), variations in the latent embedding are noticeable for different activities like running versus jumping or standing, for specific jumping targets, and for the existence of external perturbations. Two latent values (indicated by red horizontal lines in the top plots) are plotted in the bottom plots to show thta the latent space also captures the changes in the impact forces.  
To conduct an ablation study on time-invariant dynamics parameters, we examine the same regions inside red boxes (titled zoomed region) in Fig.~\ref{subfig:latent_run_force},~\ref{subfig:latent_jump_force} and show these regions in Fig.~\ref{subfig:latent_run_zoomed},~\ref{subfig:latent_jump_zoomed} when the robot is performing constant-speed running and in-place jumping respectively. 
The \emph{Latent using Default Dynamics} shows these regions when then robot's nominal model is simulated without any changes to the dynamics or introducing noise or delays. The \emph{Latent Change from Default} shows the delta change in the same regions when we adjust the CoM position of each link (marked as \textit{CoM Pos}), link mass (\textit{Link Mass}), PD gains used on each motor (\textit{PD Gain}), ground friction (\textit{Friction}), joint damping ratio (\textit{Damping}), communication delay (\textit{Delay}), and noise levels (\textit{Noisy}) one at the time.  
In these plots, a larger change is represented by a darker color. This shows that the latent embedding can reflect different changes in dynamics, while effectively filtering out noise which causes little changes. 
Despite significant environment changes, control performance metrics like task completion error ($E_t$) and motion tracking error ($E_m$) show minor degradation, showcasing adaptivity to dynamic changes using the proposed controller. We further visualize the corresponding saliency map in Appendix.~\ref{appendix:saliency}.}
\label{fig:latent_vis}
\end{figure*}

To understand why the proposed method shows advantages in achieving better learning performance in the challenging scenario of performing dynamic locomotion skills with varying environment parameters, we delve into an examination of the latent representation provided by the encoder for long I/O history, as shown in Fig.~\ref{fig:latent_vis} and Appnedix~\ref{appendix:walking_latent}.
These results are obtained in simulation. 
As we will see, the encoder that utilizes long I/O history can capture both time-varying events and time-invariant changes in dynamics parameters in all the locomotion skills we developed.

\subsection{Time Varying Embedding}
In this subsection, we observe that the latent embedding from the long I/O history encoder can effectively adapt to time-varying external disturbances or locomotion tasks, implicitly estimating contact events and/or external forces across all three distinct locomotion skills.
\paragraph{Periodic Running}
Using the running policy obtained by the proposed method, we recorded the latent embedding over a 15-second duration, initiating from a standing position and progressing to follow the constant running speed command of 3 m/s. 
A persistent backward perturbation force of 40 N was applied to the robot base from 8 to 11 seconds. 
The evolution of latent values over time is depicted in Fig.~\ref{subfig:latent_run_force}.

Given that running is a periodic skill, the latent embedding also exhibits a periodic pattern once the gait has stabilized, as seen in Fig.~\ref{subfig:latent_run_force} after 2 seconds. 
Additionally, the presence of the perturbation introduces variations in the latent embedding (within the green dashed lines), showcasing the capability to capture time-varying disturbances. 

Additionally, we found two intriguing latent dimensions highlighted by the red line and plots in Fig.~\ref{subfig:latent_run_force}. 
These two latent signals exhibit a strong correlation with the impact force on the robot's left and right foot. 
Specifically, the values of these two latent dimensions vary in a tendency consistent with the ground truth impact force recorded in simulation, reaching zero when the corresponding force is zero (the respective foot is in a swing phase).
This result highlights a crucial and demanding capability for controlling legged robots: contact estimation, as the trajectory breaks upon contact. 

The advantage of implicit contact estimation becomes more evident with changes in these two latent dimensions in the presence of external perturbation (shown within the green dashed lines in Fig.~\ref{subfig:latent_run_force}). 
Despite the unchanged magnitude of the ground impact force, these two latent values shift to a lower envelope during the perturbation, followed by a recovery to the previous envelope after the perturbation is taken out. 
While an exact explanation for this change remains elusive as it is learned purely through data, it \emph{may} be attributed to the notion that both external perturbation and ground reaction force can be treated as a generalized external force $\bm{\zeta}_{\text{ext}}$ applied to the robot, as implied by the robot's full-order dynamics \eqref{eq:full_order_dynamics}. 
The robot may have unconsciously learned to embed such external forces all together into these signals and leverage them in control, without explicit human engineering, through the provided long I/O history. 
A similar capacity for such a long history encoder for periodic walking skill is also observed, and we analyze it briefly in Appendix~\ref{appendix:walking_latent}.  

\paragraph{Aperiodic Jumping}
Despite the periodic running and walking skills, we also discovered a similar capacity of the I/O history encoder to capture time-varying information in aperiodic skills like jumping. 
As depicted in Fig.~\ref{subfig:latent_jump_force}, Cassie was commanded to execute different jumping tasks every 3 seconds, starting from an in-place jump to a 1.4-meter forward jump, followed by a 0.5-meter sideways jump and a jump and turn of $-60^\circ$ before standing. 

The recorded time-evolving latent representation reveals a distinct contrast between jumping phases (that have more varying and non-zero signals) and standing phases (with less varying signals). 
Furthermore, for different jumping tasks, the latent values during the jumping phases differ, as illustrated in Fig.~\ref{subfig:latent_jump_force}.

Moreover, we discovered two latent dimensions that strongly correlate with contact events during jumping, as indicated by the red lines and plotted in Fig.~\ref{subfig:latent_jump_force}. 
In contrast to a common human expectation for a single binary variable of contact during jumping (either in contact or not), the robot learned to utilize two distinct signals: take-off event and landing event, estimated by Latent Value 1 and 2, respectively, as depicted in Fig.~\ref{subfig:latent_jump_force}. 
For example, the Latent Value 1 begins to increase and drops to zero just before the total contact force reaches zero (indicating the robot is starting to take off), while Latent Value 2 only becomes active at the moment the robot lands. 
Furthermore, during the forward jump the robot executes a small hop after it lands, Latent Value 2 exhibits an additional spike during that additional hop.
Such separated signals for take-off and landing could provide more informative cues for controlling a bipedal jumping skill, given the difference in control complexity during the flight and landing phases.

\subsection{Adaptive Embedding for Changes in Dynamics}
We now assess how the latent embedding changes with (time-invariant) variations in the robot dynamics model, specifically, alterations in the $\mathbf{M}$, $\mathbf{C}$, $\mathbf{G}$, $\bm{\kappa}$, and Jacobians in~\eqref{eq:full_order_dynamics}, or with corrupted measurements (including noise and delays).
Using the aforementioned running policy, we focus on a segment of the periodic pattern illustrated in Fig.~\ref{subfig:latent_run_force}, highlighted by the red block. 
The zoomed-in pattern, observed when controlling the robot with the default dynamics model, is visualized in Fig.~\ref{subfig:latent_run_zoomed}. 

We conduct an ablation study using the same running policy in the same task (tracking a 3 m/s command) with changes in the dynamics parameters.
In particular, we adjust one specific modeling parameter one at time, such as the link CoM position (increased by $8$ cm for all links), link mass ($30\%$ greater for all links), joint damping ratio (8 times the default value), and PD gains used on each motor ($40\%$ higher than the default value), and lower ground friction (static friction ratio set to 0.5 for slippery ground). 
As shown in Fig.~\ref{subfig:latent_run_zoomed}, each change of these dynamics parameters results in a significant shift in the resulting latent embedding compared to the pattern obtained with the default model.
However, evaluating control performance metrics, such as task completion (tracking error $E_t$ between the robot's actual speed and desired speed) and motion tracking (motion deviation $E_m$ from the robot's actual motor position and reference motion), reveals minimal change. 

A similiar ablation study is undertaken in jumping and walking skills, recorded in Fig.~\ref{subfig:latent_jump_zoomed} and Appendix~\ref{appendix:walking_latent}, respectively.
Specifically, within the in-place jumping task utilizing the obtained jumping policy, we modify the same dynamics parameters as in running, resulting in varying patterns of the latent embedding for different dynamics models.
Despite these alterations, the control performance metrics, such as task completion $E_t$ (the error in the landing position compared to the desired one) and motion tracking error $E_m$, also show minimal changes.

When assessing the influence of corrupted measurements on the latent embedding, the latency in control and observation (0.025 s) induced noticeable changes in running and walking, as showcased in Fig.~\ref{subfig:latent_run_zoomed} and Fig.~\ref{subfig:latent_walk_zoomed}. 
Surprisingly, the introduction of significantly larger noise (2 times larger than the upper bound used during training as listed in Table~\ref{tab:randomization}) does not exert a considerable effect on the resulting latent embedding for all running, jumping, and walking skills. 
This result underscores the efficacy of the long history encoder in effectively filtering out the added zero-mean noise, which may be also brought by the CNN structure. 

\subsection{Summary of Results}
This ablation study highlights the adaptivity of the proposed controller. 
Empowered by the history encoder's ability to capture meaningful information from the I/O history, this controller can adapt to time-varying events such as external perturbations or contacts, time-invariant changes in dynamics parameters, and filter out measurement noise, and therefore, effectively execute the control task with minimal performance degradation. 
This capability explains why the proposed structure excels in the challenging training setting with a large range of randomization of dynamics parameters.
Notably, in order to fully harness such advantages brought by the robot's long I/O history, a proper structure (dual-history approach) and training strategy (end-to-end training) are necessary, as discussed in Sec.~\ref{sec:policy_structure}.

\section{Advantages of Versatile Policies and Source of Robustness}\label{sec:multi_skill}

\begin{figure*}[t]
\centering
\begin{subfigure}{0.487\linewidth}
  \centering
  \includegraphics[width=\linewidth]{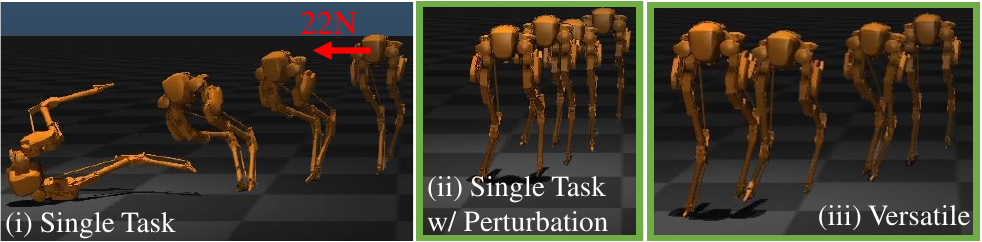}
  \caption{Walking with Consistent Unknown Lateral Perturbation Force}
  \label{subfig:bm_single_sim_perturb_walk}
\end{subfigure}
\begin{subfigure}{0.5\linewidth}
  \centering
  \includegraphics[width=\linewidth]{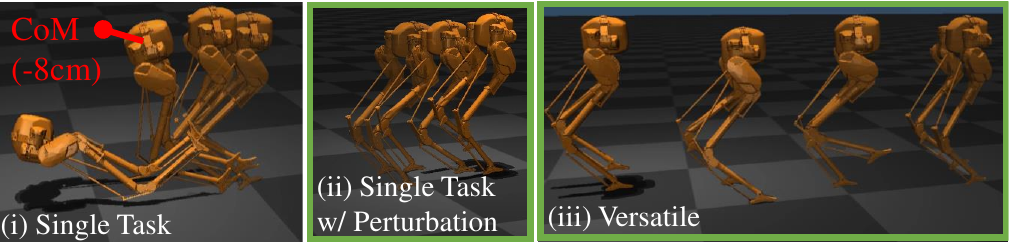}
  \caption{Walking with Errors in Center of Mass Positions of All Links}
  \label{subfig:bm_single_sim_ipos_walk}
\end{subfigure}
\begin{subfigure}{\linewidth}
  \centering
  \includegraphics[width=\linewidth]{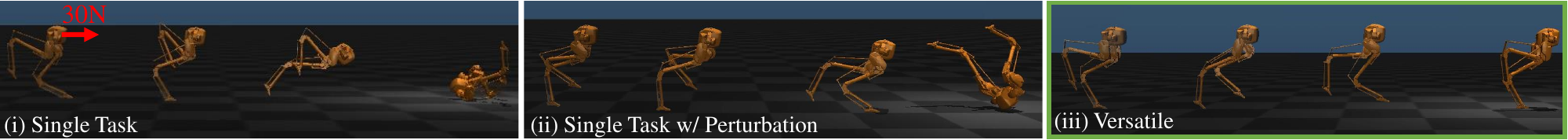}
  \caption{Running with Consistent Unknown Lateral Perturbation Force}
  \label{subfig:bm_single_sim_perturb_run}
\end{subfigure}
\begin{subfigure}{\linewidth}
  \centering
  \includegraphics[width=\linewidth]{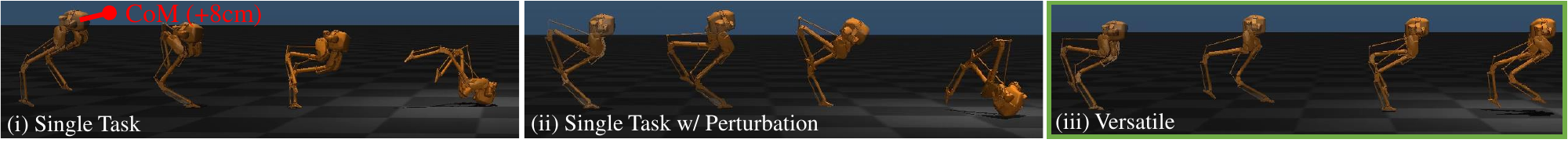}
  \caption{Running with Errors in Center of Mass Positions of All Links}
  \label{subfig:bm_single_sim_ipos_run}
\end{subfigure}
\begin{subfigure}{0.4425\linewidth}
  \centering
  \includegraphics[width=\linewidth]{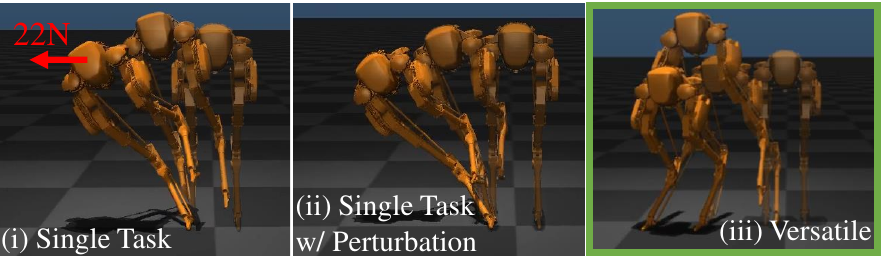}
  \caption{Jumping with Consistent Unknown Lateral Perturbation Force}
  \label{subfig:bm_single_sim_perturb_jump}
\end{subfigure}
\begin{subfigure}{0.5425\linewidth}
  \centering
  \includegraphics[width=\linewidth]{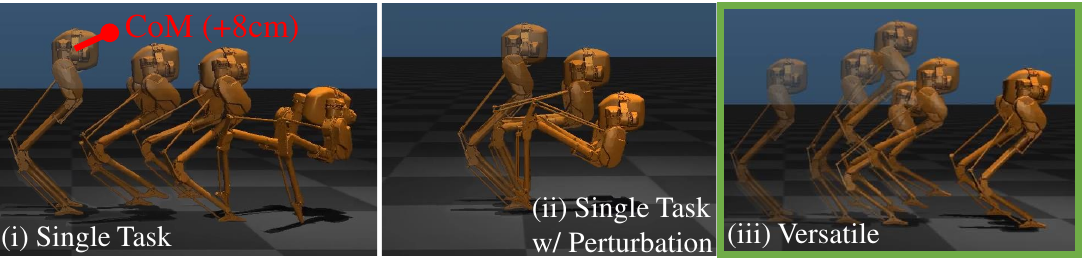}
  \caption{Jumping with Errors in Center of Mass Positions of All Links}
  \label{subfig:bm_single_sim_ipos_jump}
\end{subfigure}

\caption{Robustness test in simulation (MuJoCo). We conduct an ablation study to evaluate the relationship between the robustness and task randomization using the baselines developed in Sec.~\ref{subsec:multi_skill_baselines}. The green bounding box indicates the scenario where the policy successfully controlled the robot without falling over. The robot is subjected to consistent perturbation forces (with its direction marked by a red arrow) or CoM position offsets of all links (with the robot's entire CoM conceptually drawn as a red point) while performing fixed locomotion tasks like walking at 0.6 m/s, running at 3 m/s, or jumping in place. Policies focused on a single task, even with extensive dynamics randomization, failed in these out-of-distribution scenarios. With additional perturbation training, the robot can maintain stability during walking, as seen in Fig.~\ref{subfig:bm_single_sim_perturb_walk}ii and Fig.~\ref{subfig:bm_single_sim_ipos_walk}ii. However, such policies are less effective for more dynamic skills like running or jumping, as illustrated in Fig.~\ref{subfig:bm_single_sim_perturb_run}ii and Fig.~\ref{subfig:bm_single_sim_ipos_jump}ii.
In contrast, a versatile policy, trained across a variety of tasks, exhibits better robustness even without specific perturbation training. The robot changes its gait under external perturbations or CoM shifts, as demonstrated by using lateral (Fig.~\ref{subfig:bm_single_sim_perturb_walk}iii) and backward walking gaits (Fig.~\ref{subfig:bm_single_sim_ipos_walk}iii), highlighting the compliance of versatile controllers. This compliance is also observed during running and jumping (Fig.~\ref{subfig:bm_single_sim_perturb_run}iii and Fig.~\ref{subfig:bm_single_sim_ipos_jump}iii), which is the key to success in these challenging robustness tests.}
\label{fig:bm_single_sim}
\end{figure*}

We now validate another key finding in this study, which is the source of robustness when using RL to obtain a locomotion controller, in both simulation and the real world. 
Our findings indicate that, in the context of bipedal locomotion control, a single versatile policy capable of executing various tasks can significantly improve the robustness compared to policies specialized in individual tasks. This is realized by task randomization and generalization.

\subsection{Baselines}\label{subsec:multi_skill_baselines}
We conduct a benchmark among the proposed versatile policies and task-specific baselines across all four distinct locomotion skills, including standing, walking, running, and jumping. For each of these four skills, we develop three policies, and they are:
\begin{itemize}[leftmargin=9pt]
    \small
    \item \textbf{Single Task}: The policy that is only trained with a single fixed task and dynamics randomization (excluding simulated perturbations). For walking, the task is to walk forward at 0.6 m/s; for running, the task is running at 3 m/s; for jumping, the task is to jump in place; and for standing, the task is to stand still.
    \item \textbf{Single-Task w/ Perturbation}: The policy that is trained with the same fixed task, same dynamics randomization as the \textbf{Single-Task} policy, but is also trained with additional simulated perturbations.  
    \item \textbf{Versatile (Ours)}: The policy that is trained to handle the complete spectrum of commands (tasks) detailed in Table~\ref{tab:command}, with the dynamics randomization but \emph{without} any perturbations. For the walking and running versatile policies, they also learned the transition between locomotion and standing. 
\end{itemize}

All of them used the proposed policy structure as shown in Fig.~\ref{fig:controller} and have trained with the same range of dynamics randomization introduced in Sec.~\ref{subsec:domain_rand} until convergence. For the policies trained with simulated perturbations, the range of perturbation is also the same as specified in Table.~\ref{tab:randomization}.

The above-mentioned policies are examples trained with three different methods: (1) dynamics randomization (single-task policies trained \emph{without} simulated perturbations), (2) perturbation training (single-task policies trained \emph{with} perturbations during dynamics randomization), and (3) task randomization (versatile policies). \emph{Usually, perturbation training is considered as an additional dynamics randomization.}

\subsection{Source of Robustness}
We first start our validation in a simulation, which provides a controlled and safe environment to assess the robustness of the control policies. 
In this validation, we are going to show that the robustness of a versatile locomotion policy comes from two key factors: (1) the generalization of trained randomized dynamics parameters introduced by \emph{dynamics randomization}, and (2) the generalization of various trained locomotion tasks, thereby endowing the robot with notable \emph{compliance}, brought by \emph{task randomization}.

For each locomotion control policy we've developed (including walking, running, and jumping), we test them under two distinct forms of uncertainties: (1) applying a consistent force on the robot's pelvis, and (2) introducing a significant deviation in the CoM position for all links of the robot, as depicted in Fig.~\ref{fig:bm_single_sim}. 
It's important to note that these uncertainties exceed the bound used during training, as outlined in Table~\ref{tab:randomization}. 
For example, unlike the impulse perturbation used during training, here we apply a consistent perturbation, and the CoM position offset is greater than what was covered in the training scenarios. 
The results of these out-of-distribution tests are illustrated in Fig.~\ref{fig:bm_single_sim}.

Consider the example of a walking task: when a robot is commanded to walk forward and faces a consistent lateral pulling force of 22 N, a single-task policy tailored for forward walking fails due to the lateral perturbation pushing the robot beyond its training distribution, as seen in Fig.~\ref{subfig:bm_single_sim_perturb_walk}(i). 
However, if the robot is trained with perturbation using a single-task policy, it can progress forward with minor lateral deviation under such force, shown in Fig.~\ref{subfig:bm_single_sim_perturb_walk}(ii).
In contrast, using a versatile policy that is not trained with perturbations but trained with various walking tasks, the robot still stabilizes and performs the forward walking task, albeit differently. 
This is evident from Fig.~\ref{subfig:bm_single_sim_perturb_walk}(iii), where the robot significantly deviates to the right, using learned side walking skill to compensate for the external force, resulting in a \emph{compliant} gait.

Similar results are observed with a CoM position offset of $-8$ cm in all links. Without training for perturbations or other tasks, the robot, even with randomized CoM training, cannot handle this deviation, as shown in Fig.~\ref{subfig:bm_single_sim_ipos_walk}(i). 
When trained with perturbation, since the robot has learned to stabilize under a backward force, it uses such a learned control strategy to counter the backward CoM position offset and keep walking forward with a reduced speed, as depicted in Fig.\ref{subfig:bm_single_sim_ipos_walk}(ii). 
Yet, using a versatile policy without perturbation training, the robot instead leverages backward walking gaits to offset the rearward CoM shift, as seen in Fig.~\ref{subfig:bm_single_sim_ipos_walk}(iii).

\begin{figure*}[t]
\centering
\begin{subfigure}{0.154\linewidth}
  \centering
  \includegraphics[width=\linewidth]{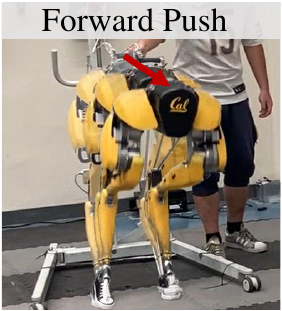}
  \caption{Standing skill only}
  \label{subfig:bm_multi_stand_only}
\end{subfigure}
\begin{subfigure}{0.154\linewidth}
  \centering
  \includegraphics[width=\linewidth]{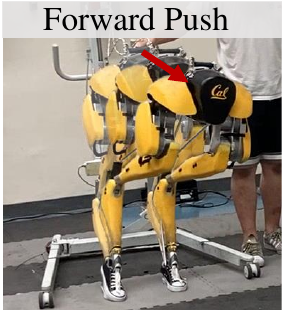}
  \caption{With perturbation}
  \label{subfig:bm_multi_stand_perturb}
\end{subfigure}
\begin{subfigure}{0.68\linewidth}
  \centering
  \includegraphics[width=\linewidth]{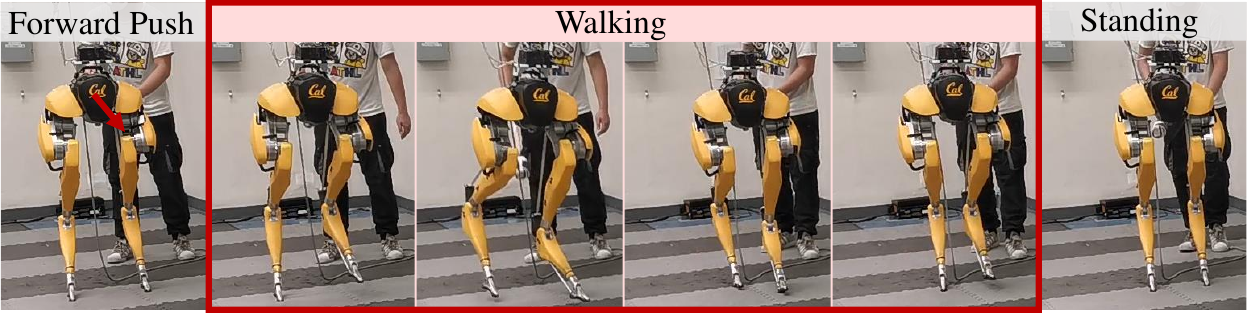}
  \caption{Trained with both walking skill and standing skill (trained without perturbation)}
  \label{subfig:bm_multi_walking}
\end{subfigure}
\caption{Robustness test for bipedal standing skill in the real world. If the robot is trained exclusively with the standing skill, regardless of whether it underwent extensive dynamics randomization with (Fig.~\ref{subfig:bm_multi_stand_perturb}) or without simulated perturbation (Fig.~\ref{subfig:bm_multi_stand_only}), the robot tends to fall when pushed beyond its support region. In contrast, the robot using our versatile policy, trained in both walking and standing skills as shown in Fig.~\ref{subfig:bm_multi_walking}, demonstrates better robustness. Despite not being trained with perturbation during standing, the robot can spontaneously break contact when pushed forward and employ its generalized walking skills to regain its standing pose. The video of such a comparison is recorded in Vid. 3 in Table~\ref{tab:video_list}.}
\label{fig:robust_standing}
\end{figure*}

\begin{figure*}[!htp]
\centering
\begin{subfigure}{0.9\linewidth}
  \centering
  \includegraphics[width=\linewidth]{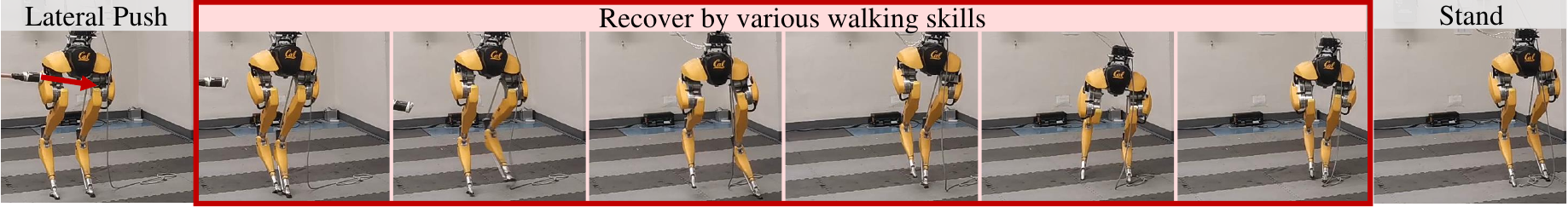}
  \caption{A complex recovery maneuver using various walking skills (walking sideway, backwards, and changing heights) after being perturbed during standing}
  \label{subfig:stand_recover_by_walk}
\end{subfigure}
\begin{subfigure}{0.575\linewidth}
  \centering
  \includegraphics[width=\linewidth]{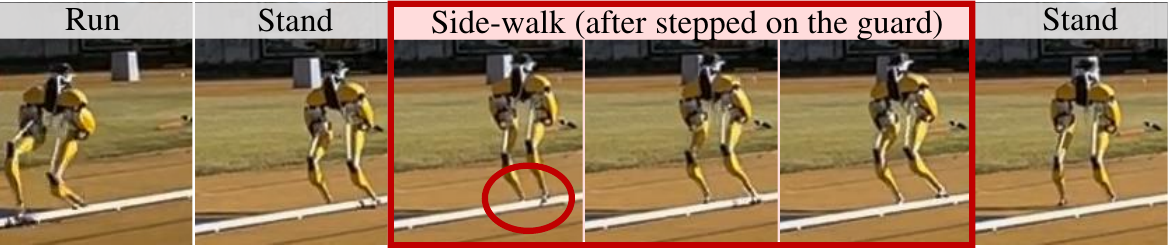}
  \caption{Robust standing maneuver aided by running skills}
  \label{subfig:stand_recover_by_run}
\end{subfigure}
\begin{subfigure}{0.4\linewidth}
  \centering
  \includegraphics[width=\linewidth]{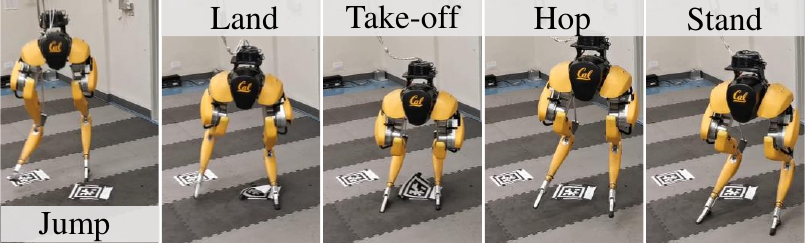}
  \caption{Robust standing maneuver aided by jumping skills}
  \label{subfig:stand_recover_by_jump}
\end{subfigure}
\caption{Robust bipedal standing skill realized by versatile policies in the real world. The robot, while standing, can intelligently break contact and engage learned locomotion skills as needed. For instance, as shown in Fig.~\ref{subfig:stand_recover_by_walk}, when perturbed, the robot deviates from its standing command and autonomously employs a sequence of contacts and varying walking skills over a long horizon to return to a standing position. This robustness, a result of task randomization, enhances real-world deployment capabilities. For example, in transitions to standing, like in Fig.~\ref{subfig:stand_recover_by_run}, the robot uses side-stepping skills acquired during running to aid recovery from stepping on a track guard. Similarly, with the versatile jumping skill in Fig.~\ref{subfig:stand_recover_by_jump}, the robot breaks contact after landing from an unstable state, adjusting its pose in the air for better landing, all without relying on any human-defined contact sequences.}
\label{fig:robust_standing_run_jump}
\end{figure*}

Please note that we are not trying to argue that perturbation training is inferior to task randomization. 
For example, in the walking task, robots trained with a single task and perturbations showed more favor ability to complete the assigned task (less tracking error shown in Fig.~\ref{subfig:bm_single_sim_perturb_walk}(ii), \ref{subfig:bm_single_sim_ipos_walk}(ii)), compared to those controlled by versatile policies (Fig.~\ref{subfig:bm_single_sim_perturb_walk}(iii),\ref{subfig:bm_single_sim_ipos_walk}(iii)). 
Therefore, incorporating external perturbations during dynamics randomization is still recommended following task randomization. 
However, the enhancement in robustness from simulated dynamics is not as significant as that from task randomization, especially in more dynamic tasks like running and jumping.

As depicted in Fig.~\ref{subfig:bm_single_sim_perturb_run}(i)(ii), under a constant forward perturbation of 30 N while running, policies trained only with single task fail to maintain stable gaits, regardless of extensive dynamics randomization and additional perturbation training. 
In contrast, robots using the versatile policy that has trained for faster running can adapt to such perturbations, as demonstrated in Fig.~\ref{subfig:bm_single_sim_perturb_run}(iii). 
Similar results are observed in scenarios involving a $+8$ cm CoM position offset during running (Fig.\ref{subfig:bm_single_sim_ipos_run}), and in lateral perturbation (by using a lateral jump) and forward CoM position offset (by using a forward jump) cases in jumping (Fig.~\ref{subfig:bm_single_sim_perturb_jump}, \ref{subfig:bm_single_sim_ipos_jump}). 
Note that all of these versatile policies are not trained with simulated perturbations. 

These studies highlight the two distinct sources of robustness of RL policies: (1) Training with specific simulated dynamics, including external perturbations, allows functioning within an expanded scenario range but limits the robot to the trained task; 
(2) Training with a diverse task set enables the robot to generalize learned tasks for greater robustness and compliance, even without extensive dynamics randomization.

\begin{figure}[t]
    \centering
    \includegraphics[width=0.7\linewidth]{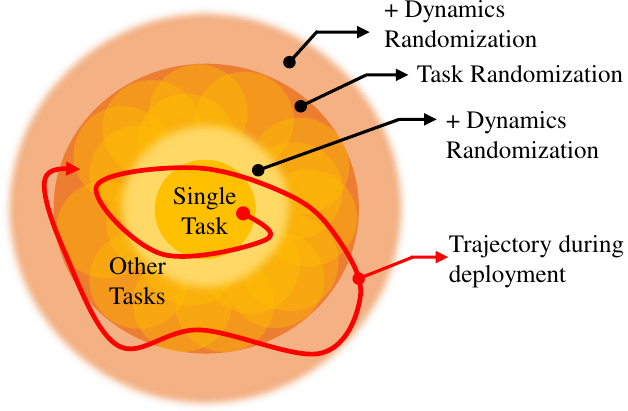}
    \caption{An illustration of the concept of training distributions using different methods to enhance robustness. During deployment, as conceptually illustrated by the red curve, we want the robot controlled by its RL-based policy to operate inside the training distribution of the robot's trajectories. When the training is focused on a single task, the training distribution is confined to nominal trajectories specific to that task, drawn as the yellow region. Incorporating extensive dynamics randomization, including simulated perturbations or varying terrains, can expand this distribution. However, this expansion is still centered around the fixed task. Task randomization significantly broadens the training distribution (to the orange region) by enabling the robot to learn and generalize various control strategies across different tasks (marked as different faded yellow regions). It is important to note that task randomization can be combined with dynamics randomization, further widening the training distribution and enhancing the policy's robustness.}
    \label{fig:generalization_illustration}
\end{figure}

\subsection{Case Study: Robust Standing Experiments}

We conduct a case study in the real world to gain insights into the underlying source of robustness in control policies developed through RL. 
This case study centers on evaluating the performance of the standing skill.

In this scenario, as shown in Fig.~\ref{fig:robust_standing}, we introduce an external forward perturbation to the robot's pelvis while it maintains a standing pose. 
When policies are exclusively trained for the standing skill, regardless of whether they were trained with external perturbations, the robot keeps losing its balance if it is forced to lean beyond its support region, as recorded in Figs.~\ref{subfig:bm_multi_stand_only},~\ref{subfig:bm_multi_stand_perturb}. 
Conversely, when we employ the versatile walking policy, which is also trained with the standing skill, and subject the robot to a similar forward perturbation while it is standing, as demonstrated in Fig.~\ref{subfig:bm_multi_walking}, the robot initially leans forward. However, should the robot start leaning outside its support region, it demonstrates intelligent recovery maneuvers. 
It transits to a walking gait by breaking contact, executes several steps, including forward and backward walking, and then smoothly reverts to a standing pose again. 
Remarkably, this transition occurs without any human-provided commands, as the only reference motion provided is a standing motion. 
Also, we do not simulate external perturbation during the training of the standing skill using this walking policy.

\begin{figure*}[!htp]
    \centering
    \begin{subfigure}{\linewidth}
    \includegraphics[width=\linewidth]{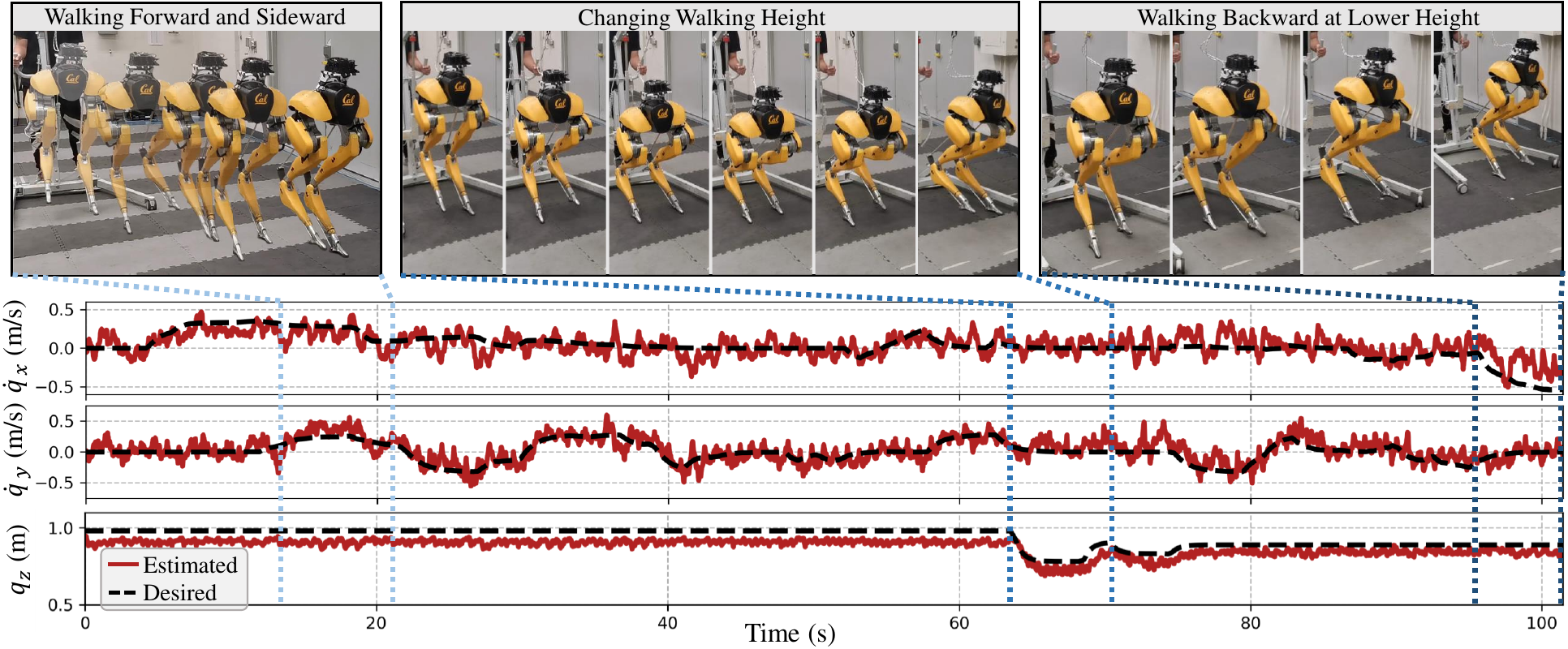}
    \caption{Experiments on variable command tracking conducted on August 9, 2022. The tracking errors (Mean Absolute Error, MAE) in the (sagittal velocity $\dot{q}^d_x$, lateral velocity $\dot{q}^d_y$, walking height $q^d_z$) are (0.10 m/s, 0.10 m/s, 0.06 m), respectively.}\label{subfig:variable_cmd}
    \end{subfigure}
    \begin{subfigure}{\linewidth}  
    \includegraphics[width=\linewidth]{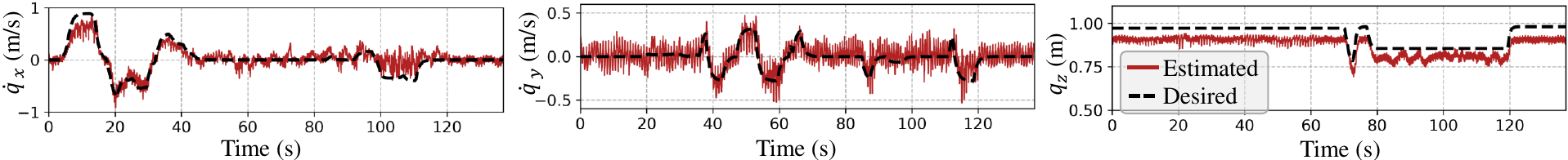}
    \caption{Experiments on variable command tracking conducted on June 30, 2023, 325 days after the one conducted Fig.~\ref{subfig:variable_cmd} using the same controller without any tuning. The tracking errors (MAE) in the $(\dot{q}^d_x, \dot{q}^d_y, q^d_z)$ directions are (0.10 m/s, 0.08 m/s, 0.06 m), repsectively}\label{subfig:variable_cmd_1214}
    \end{subfigure}
    \begin{subfigure}{\linewidth}
    \includegraphics[width=\linewidth]{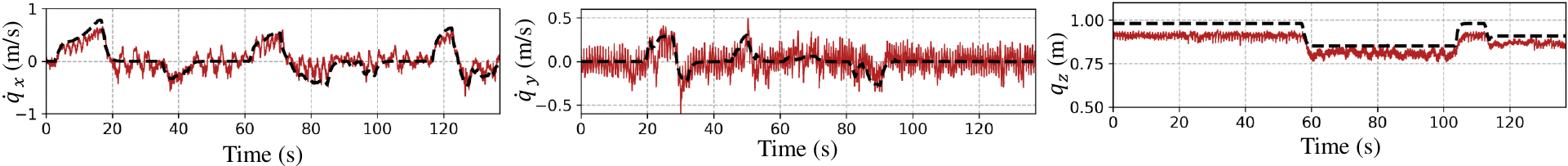}
    \caption{Experiments on variable command tracking conducted on December 14, 2023, 492 days after the one conducted in Fig.~\ref{subfig:variable_cmd} using the same controller without any tuning. The tracking errors (MAE) in the $(\dot{q}^d_x, \dot{q}^d_y, q^d_z)$ directions are (0.11 m/s, 0.09 m/s, 0.05 m), respectively}\label{subfig:variable_cmd_0630}
    \end{subfigure}  
    \caption{Fig.~\ref{subfig:variable_cmd} records a snapshot of the robot reliably tracking varying commands using the obtained versatile walking controller in the real world. The desired command and actual estimated values are recorded onboard during the experiments and drawn in the lower part. The dashed blue lines match the corresponding interval of frames in the real world. The robot can track different desired commands including changing sagittal velocity $\dot{q}^d_x$, lateral velocity $\dot{q}^d_y$, and walking height $q^d_z$, with a considerable accuracy that has a tracking error over a long test. Similar variable command tracking experiments are conducted after a long timespan, such as the one conducted after 325 days in Fig.~\ref{subfig:variable_cmd_0630} and the one after 492 days in Fig.~\ref{subfig:variable_cmd_1214}. Although the robot hardware has been changing due to wear and tear after such a long time, the same RL-based controller can still effectively control the robot to track variable commands with similar tracking errors, without any tuning in the real world.}\label{fig:variable_cmd}
\end{figure*}

The presence of a robust standing skill is highly advantageous in real-world deployments.
For example, when the robot is laterally perturbed while standing, as shown in Fig.~\ref{subfig:stand_recover_by_walk}, it utilizes its varied walking skills to recover and return to a stand. This complex sequence, transitioning through various walking maneuvers to lower its center of mass and then resuming a standing pose, highlights the policy's capability for long-horizon recovery maneuvers. 
Notably, this sophisticated recovery was not specifically trained but is a natural outcome of the versatile policy.
This benefit extends to other skills as well. 
When using the versatile running policy, as seen in Fig.~\ref{subfig:stand_recover_by_run}, the robot halts from running and steps on the track guard, as recorded in Fig.~\ref{subfig:stand_recover_by_run}.
Without loss of balance, the robot promptly disengages from the guard and uses its acquired side-stepping skills when learning the running skill, and maintains a stable stance pose afterward.
Similarly, with the versatile jumping policy (Fig.~\ref{subfig:stand_recover_by_jump}), after an unstable landing from a complex multi-axis jump, the robot executes a corrective hop, which is learned from diverse jumping tasks, to better correct itself in the air and to have a more stable landing configuration. These results can be better seen in the videos listed in Table~\ref{tab:video_list}.

\subsection{Understand Robustness from Training Distributions}
The above-mentioned findings can be conceptually illustrated in Fig.~\ref{fig:generalization_illustration}, similar to the notion of an \emph{invariant set} in control theory. 
To effectively control the robot, an RL policy needs to address a range of disturbances encountered during deployment. 
A model-free RL policy can be particularly effective when the robot's trajectory lies within its training distribution, as the policy has been specifically trained and optimized for such scenarios.
Single-task policies have a limited training distribution, focused only on that task, and training with random dynamics w/wo perturbations can extend this distribution, as shown in Fig.~\ref{fig:generalization_illustration}.
However, as evidenced in Fig.~\ref{fig:bm_single_sim}, dynamics randomization only modestly expands this distribution, primarily enhancing robustness within the single-task scheme.
In contrast, task randomization broadens the training distribution significantly. 
By enabling the robot to handle diverse tasks, this approach can effectively expand the range of the trained robot's I/O trajectory. 
When faced with disturbances during deployment, the robot can still remain within this enhanced training distribution, as illustrated in Fig.~\ref{fig:generalization_illustration}.
When combined with dynamics randomization after the task randomization, as proposed in our work in Fig.~\ref{fig:multi_stage_training}, the training distribution can further expand substantially and therefore enhance the robustness of the RL policy.

\begin{remark}
    The range of dynamics randomization can not be arbitrarily large as it will introduce significant challenges during training (the robot may fail to learn meaningful skills). The dynamics randomization range used in this work (Table.~\ref{tab:randomization}) is challenging enough for Cassie to learn, which can be evident by the low converged training return in Fig.~\ref{fig:dynrand_three}. Further adding dynamics randomization has limitations as suggested in~\cite{xie2021dynamics}. Therefore, task randomization can be viewed as an “orthogonal" way to improve the robustness rather than further pushing the range of the dynamics randomization. 
\end{remark}

\subsection{Summary of Results}
In conclusion, compared to policies focused solely on specific tasks, versatile policies exhibit significant improvements in robustness, which is validated in both simulation and the real world. 
This enhanced robustness stems from their ability to generalize learned tasks and to find better maneuvers to tackle unforeseen situations without being limited to adhering to given commands, ultimately leading to better stability. 
In RL-based locomotion control, one key source of robustness is the capacity to perform versatile tasks. 
Hence, task randomization, which diversifies the tasks during training, is recommended for future development on legged locomotion control.
\section{Dynamic Bipedal Locomotion in the Real World}\label{sec:experiments}

We now extensively evaluate the obtained versatile policies for walking, running, and jumping in the real world. 
All of these policies are able to effectively control the robot in the real world without further tuning after being trained in simulation. 
Moreover, the main objective of the experiments is to evaluate two fundamental aspects of the proposed method: (1) the obtained policies' \textit{adaptivitiy} to the actual robot dynamics in the real world by effectively controlling the real robot to achieve the assigned tasks, as in simulation; and (2) the policies' \textit{robustness} in addressing different uncertainties in the real world by leveraging various learned and tasks.

\subsection{Walking Experiments}
We now examine the versatile walking policy on Cassie. 
The following experiments are conducted by a \textit{single} controller. As we will see, the obtained walking policy shows consistent control performance on the robot hardware over a long timespan with notable robustness and compliance.

\subsubsection{Tracking Performance}\label{subsec:tracking_performance_walking}
The policy demonstrates the capacity to reliably track varying, fast-changing commands, with consistent control performance over a long period that is more than a year. Such a result showcases the adaptivity of the proposed policy that adapts to the dynamics of the real robot, as discussed below.

\paragraph{Variable Commands}

\begin{figure*}[!htp]
    \centering
    \includegraphics[width=\linewidth]{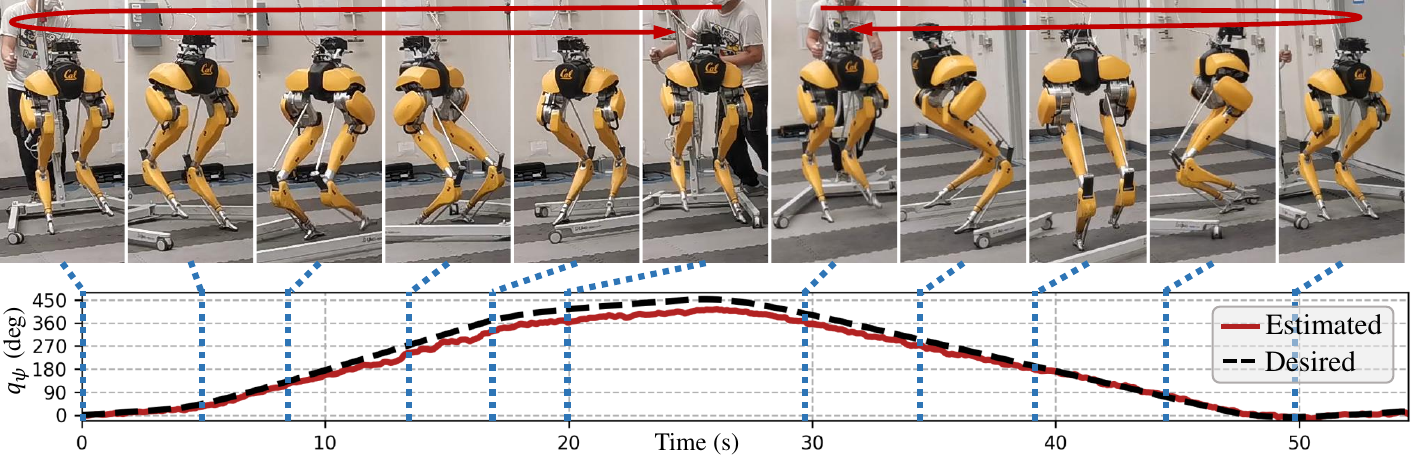}
    \caption{A snapshot from the real world demonstrating the robot reliably tracking various turning yaw commands $q^d_{\psi}$ using the same controller used in Fig.~\ref{fig:variable_cmd}. Both the desired and actual turning angles are recorded in the lower part, with dashed lines matching the corresponding frames in the real world. The robot can execute full turns in both counterclockwise and clockwise directions.}
    \label{fig:turning}
\end{figure*}

\begin{figure*}[!htp]
\centering
\begin{subfigure}{0.345\linewidth}
  \centering
  \includegraphics[width=\linewidth]{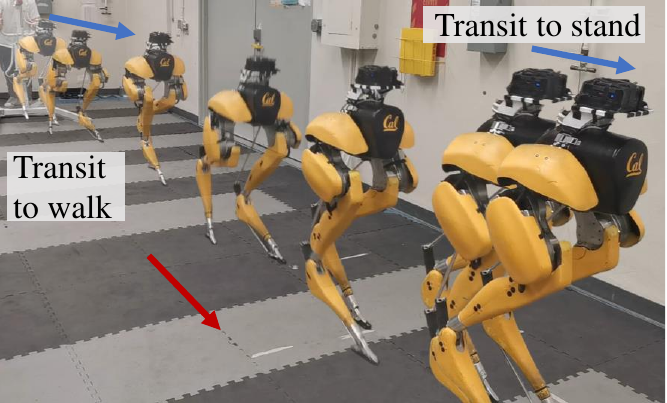}
  \caption{Fast Forward Walking}
  \label{subfig:fast_forward}
\end{subfigure}
\begin{subfigure}{0.345\linewidth}
  \centering
  \includegraphics[width=\linewidth]{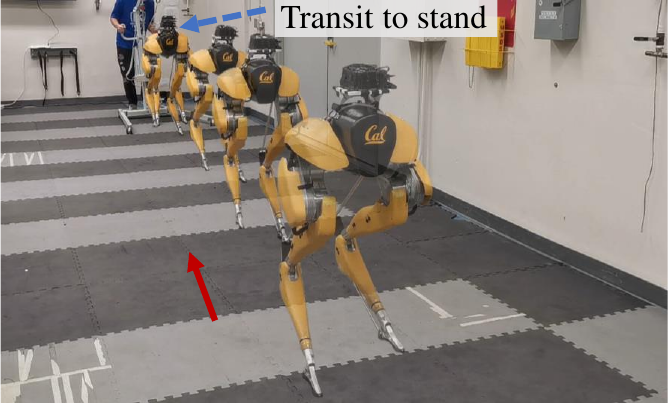}
  \caption{Fast Backward Walking}
  \label{subfig:fast_backward}
\end{subfigure}
\begin{subfigure}{0.2725\linewidth}
  \centering
  \includegraphics[width=\linewidth]{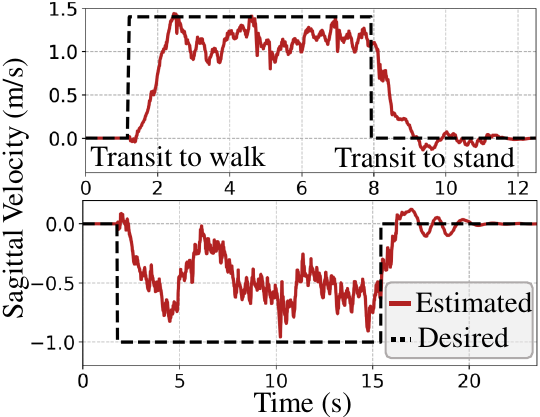}
  \caption{Corresponding Recorded Data}
  \label{subfig:fast_walking_log}
\end{subfigure}
\caption{Snapshots of the robot transiting from standing to fast forward walking (Fig.~\ref{subfig:fast_forward}) or backward walking (Fig.~\ref{subfig:fast_backward}) and back to standing, using the same walking policy. Earlier frames appear more faded. The robot's commanded and actual sagittal velocities are recorded in Fig.~\ref{subfig:fast_walking_log}. This demonstrates the robot's capability to quickly switch from a stationary stance to dynamic fast walking and smoothly transition back to standing with a single command, even during dynamic maneuvers like fast walking.}
\label{fig:fast_walking}
\end{figure*}

As recorded in Fig.~\ref{subfig:variable_cmd}, the walking policy demonstrates efficient control over the robot's following of diverse commands, including variations in sagittal and lateral velocities ($\dot{q}_{x,y}$) and walking height ($q_z$). 
Throughout the test, the tracking errors, evaluated by Mean Absolute Error (MAE), remain reasonably low. 
The MAE in the $(\dot{q}_x, \dot{q}_y, q_z)$ are (0.10 m/s, 0.10 m/s, 0.06 m) respectively.
Additionally, as indicated in Fig.~\ref{fig:turning}, the policy exhibits reliable control, enabling the robot to accurately track varying turning commands, either clockwise or anti-clockwise.

\paragraph{Consistency over a Long Timespan}
The robot hardware, especially for bipedal robots, keeps changing due to wear and tear over time. 
For instance, joint friction might vary due to the impacts during operation, and these changes can accumulate due to the bipedal robot's large number of DoFs and over extended periods. 
Consequently, controllers dependent on the control gains tuned on hardware, like most model-based controllers, often require manual updates to deal with these hardware changes.
In contrast, our proposed RL-based walking policy can adapt to changing robot dynamics and control the robot without the need for tuning on the hardware. 
This adaptivity is evident as the policy consistently performs well over extended periods, maintaining its ability to track variable commands even after 325 and 492 days, as showcased in Fig.~\ref{subfig:variable_cmd_0630} and Fig.~\ref{subfig:variable_cmd_1214}, respectively. 
Compared with the test conducted earlier (Fig.~\ref{subfig:variable_cmd}), the changes of the tracking errors (MAE) in $(\dot{q}_x, \dot{q}_y, q_z)$ during these two tests are ($+0$ m/s, $-0.02$ m/s, $+0$ m) and ($+0.01$ m/s, $-0.01$ m/s, $-0.01$ m), respectively, over a relatively long testing timespan (over 2 minutes) with different combinations of commands.
Despite significant accumulated changes in the robot's dynamics over this period, the same controller from Fig.~\ref{subfig:variable_cmd} continues to effectively manage varying walking tasks with only minimal tracking error degradation.

\paragraph{Fast Walking}
In addition to previously demonstrated moderate walking speeds, the policy exhibits the ability to control the robot to perform fast walking maneuvers, both forwards and backwards, as shown in Fig.~\ref{fig:fast_walking}. 
The robot can transition from a standstill to rapidly achieve a forward walking speed with an average velocity of $1.14$ m/s to track the command of $1.4$ m/s, quickly returning to a standing position as commanded, as shown in Fig.~\ref{subfig:fast_forward} with data recorded in Fig.~\ref{subfig:fast_walking_log}.
Notably, during the transition from a stance pose with zero speed, the robot can quickly swing its legs forward with a considerable amount of acceleration while maintaining gait stability to follow the step-input command. 
Similar capacity is observed in the context of fast backward walking (Fig.~\ref{subfig:fast_backward}), where the robot seamlessly shifts from a standing pose to perform a backward walking gait with an average velocity of $-0.5$ m/s, achieving the specified $-1$ m/s walking task before promptly returning to a standing position upon command.

\begin{figure*}[t]
\centering
\begin{subfigure}{0.495\linewidth}
  \centering
  \includegraphics[width=\linewidth]{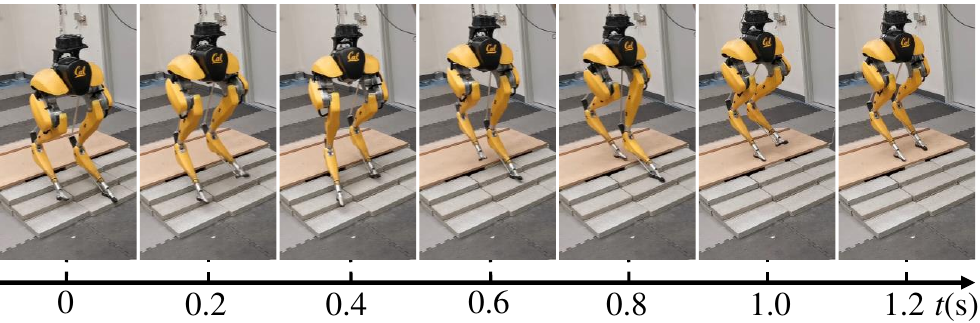}
  \caption{Walking Backwards on Stairs}
  \label{subfig:walking_stairs}
\end{subfigure}
\begin{subfigure}{0.495\linewidth}
  \centering
  \includegraphics[width=\linewidth]{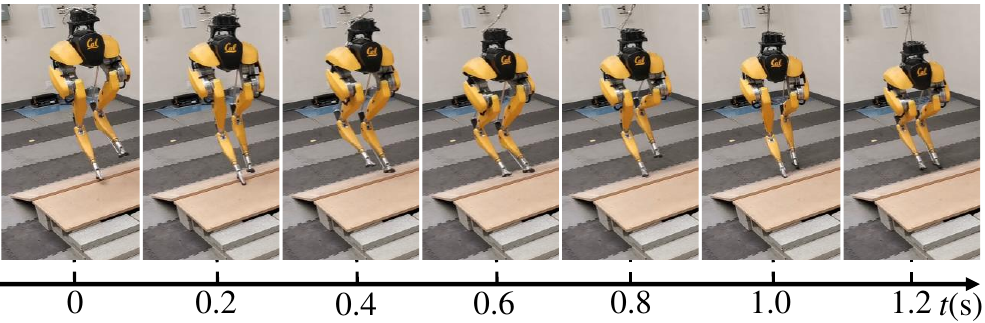}
  \caption{Walking Backwards on Slope}
  \label{subfig:walking_slope}
\end{subfigure} 
\caption{Snapshots of the robot Cassie performing a robustness test by walking \emph{backwards} over small varied terrains such as stairs (Fig.~\ref{subfig:walking_stairs}) or slope (Fig~\ref{subfig:walking_slope}). Snapshots are aligned with time stamps. Although the policy wasn't specifically trained for terrain deviations, it successfully maintains stable walking gaits and robustness in the face of ground elevation changes in real-world scenarios as recorded in these set of snapshots.}
\label{fig:walking_terrain}
\end{figure*}

\begin{figure*}[t]
\centering
\begin{subfigure}{0.65\linewidth}
  \centering
  \includegraphics[width=\linewidth]{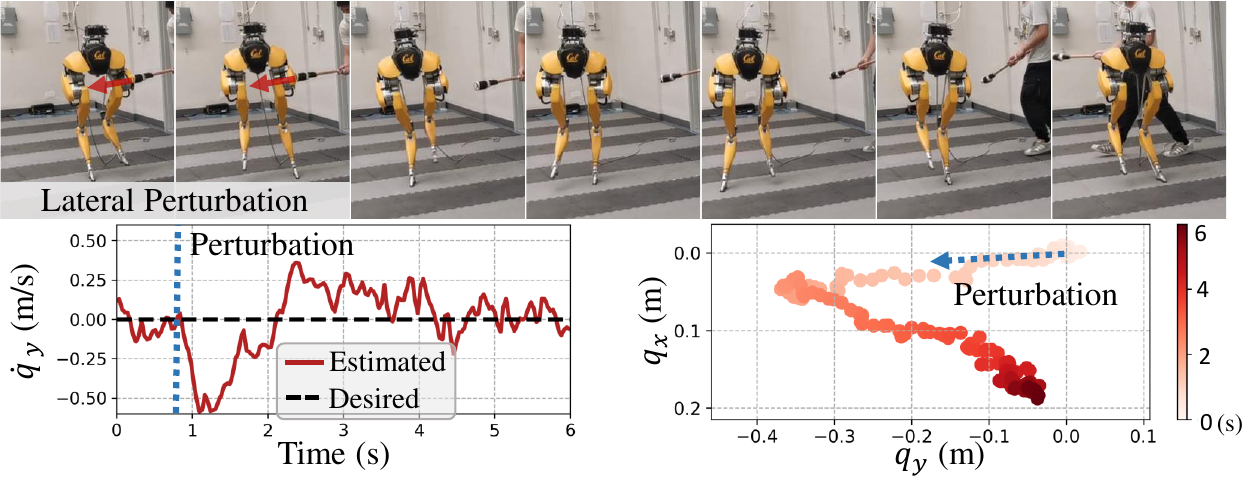}
  \caption{Recovery maneuvers using the proposed RL-based controller from lateral perturbation}
  \label{subfig:push_recovery_rl}
\end{subfigure}
\begin{subfigure}{0.31\linewidth}
  \centering
  \includegraphics[width=\linewidth]{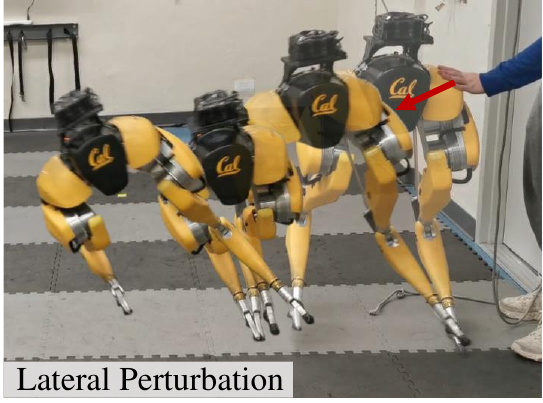}
  \caption{A model-based controller that failed to recover the robot from lateral perturbation}
  \label{subfig:push_recovery_mpc}
\end{subfigure}  
\caption{Robustness test of the robot walking against unknown perturbations. Using the proposed RL-based versatile policy, as seen in Fig.\ref{subfig:push_recovery_rl}, the robot, despite being pushed laterally and accelerated to $-0.5$ m/s, still maintains a stable walking gait and compensates such a lateral impulse by walking in the opposite direction. The corresponding lateral velocity $\dot{q}_y$ is recorded in the lower part of the figure. Additionally, the planar position $(q_y,q_x)$ is estimated, with points appearing in progressively darker colors as they are recorded later in time. In contrast, using a model-based controller provided by Cassie's manufacturer (Fig.~\ref{subfig:push_recovery_mpc}), the robot crashes after a similar lateral push.}
\label{fig:walking_perturb}
\end{figure*}

\begin{figure*}[t]
\centering
\begin{subfigure}{0.8\linewidth}
  \centering
  \includegraphics[width=\linewidth]{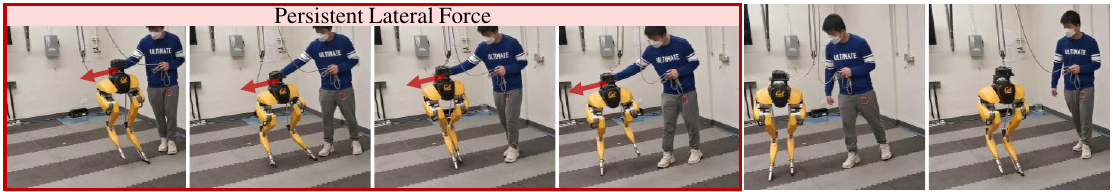}
  \caption{Staying compliant to a persistent lateral force at a normal walking height}\label{subfig:compliance_normalheight}
\end{subfigure}
\begin{subfigure}{0.8\linewidth}
  \centering
  \includegraphics[width=\linewidth]{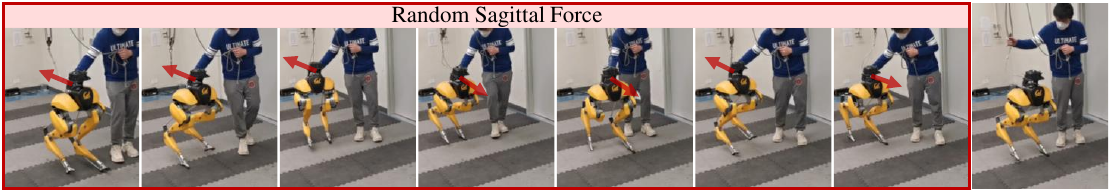}
  \caption{Staying compliant to a persistent and random sagittal force at a low walking height}\label{subfig:compliance_lowerheight}
\end{subfigure} 
\caption{Snapshots of the compliance test using the proposed versitale walking policy in the real world. The red arrows illustrate the direction of the applied force. In Fig.~\ref{subfig:compliance_normalheight} where the robot is walking at a normal height, a human applies a persistent (longer than 3 seconds) lateral force on Cassie's base and drags the robot to its right. In Fig.~\ref{subfig:compliance_lowerheight} where the robot is at a low walking height, a human exerts a persistent but random force on the robot's base and drags the robot back and forth in the sagittal direction. In these tests, the robot demonstrates compliance by walking along the persistent external force direction without losing balance, although being commanded to walk in place. After the force is removed, the robot returns to the commanded task.}
\label{fig:walking_compliance}
\end{figure*}

\subsubsection{Robust Walking Maneuvers}
We further test the robustness of the walking policy, and the proposed policy shows notable robustness in addressing various changes in the environment, as showcased below.

\paragraph{Uneven Terrains (Untrained)} 

Although the walking policy has not been specifically trained for traversing uneven terrain in simulation, it displays considerable robustness to varying elevation changes during deployment.
As depicted in Fig.~\ref{fig:walking_terrain}, the robot can effectively walk \emph{backward} on small stairs or declined slopes. 
It's noteworthy that the robot lacks any terrain elevation sensors. 
Such ability stems from the policy's robustness to the change of contact timing or wrench while walking on varying elevations, facilitated by \emph{not} requiring explicit estimation or control of the contact in the controller.

\paragraph{Robustness to Random Perturbations}
In this test, we evaluate the robustness of the proposed versatile walking policy under two types of external perturbations: (1) impulse perturbation whose elapsed time is less than 1 second; (2) persistent perturbation that lasts more than 3 seconds (longer than the length of controller's I/O history). 

For the impulse perturbation case, we introduced external perturbations over a short timespan from various directions to the robot while walking. 
As an example recorded in Fig.~\ref{subfig:push_recovery_rl}, a substantial lateral perturbation force is applied to the robot while walking in place, causing a significant lateral velocity peak of 0.5 m/s. 
Despite this force, the robot swiftly recovers from the lateral deviation. 
As recorded in Fig.~\ref{subfig:push_recovery_rl}, the robot adeptly moves in the opposite lateral direction, effectively compensating for the perturbation and restoring its stable in-place walking gait.

During the test for persistent perturbation, a human exerts a persistent force on the robot base and drags the robot in random directions while the robot is commanded to walk in place.  
Examples can be seen in Fig.~\ref{subfig:compliance_normalheight}, a persistent lateral dragging force is applied to Cassie's base while the robot is walking at normal height. 
In another test where the robot is walking at a lower height as shown in Fig.~\ref{subfig:compliance_lowerheight}, the robot's base is subject to a persistent force with its direction changing randomly in the sagittal direction. 
During these tests, without losing balance, the robot shows compliance to these external forces by following the directions of these forces, despite being commanded to walk in place. 
These compliance experiments demonstrate the advantages of the proposed RL-based policy in controlling the bipedal robot for potential applications like safe human-robot interaction.

More scenarios, including more impulse perturbations and persisting random perturbations, are recorded in the accompanying Vid. 3 in Table~\ref{tab:video_list}.
In all conducted tests, the robot demonstrates its robustness, being consistent with the results observed in the simulation validation in Sec.~\ref{sec:multi_skill}.
The underlying reason behind this lies in the robot's extensive learning of various tasks such as lateral, forward, backward walking, turning, and more.
This task randomization enables the robot to generalize the knowledge gained while learning from various tasks to recover from deviations occurring in the current commanded task.

\begin{figure*}[t]
\centering
\begin{subfigure}{\linewidth}
  \centering
  \includegraphics[width=\linewidth]{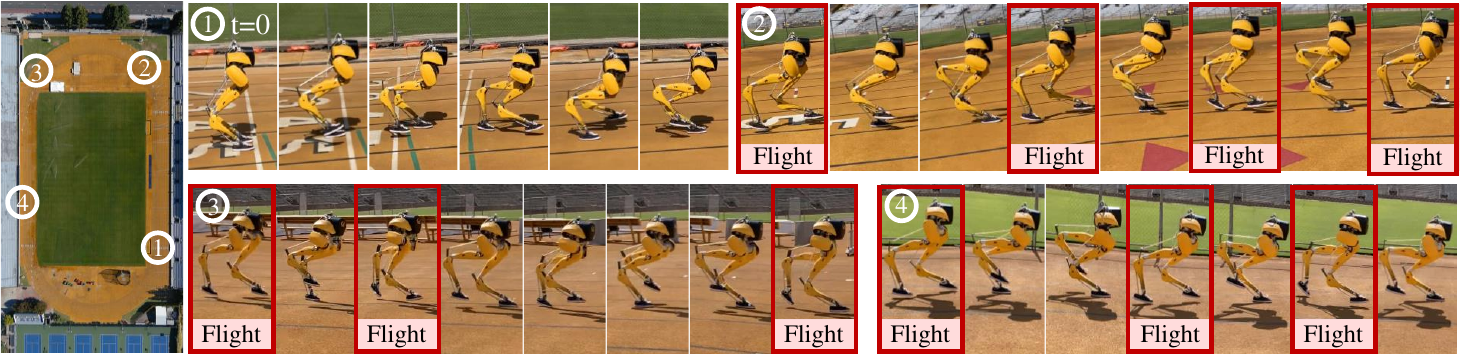}
  \caption{Snapshot of 400-meter dash. The number matches the snapshot of the robot and the corresponding position on the running track.}
  \label{subfig:run_400_snapshot}
\end{subfigure}
\begin{subfigure}{\linewidth}
  \centering
  \includegraphics[width=\linewidth]{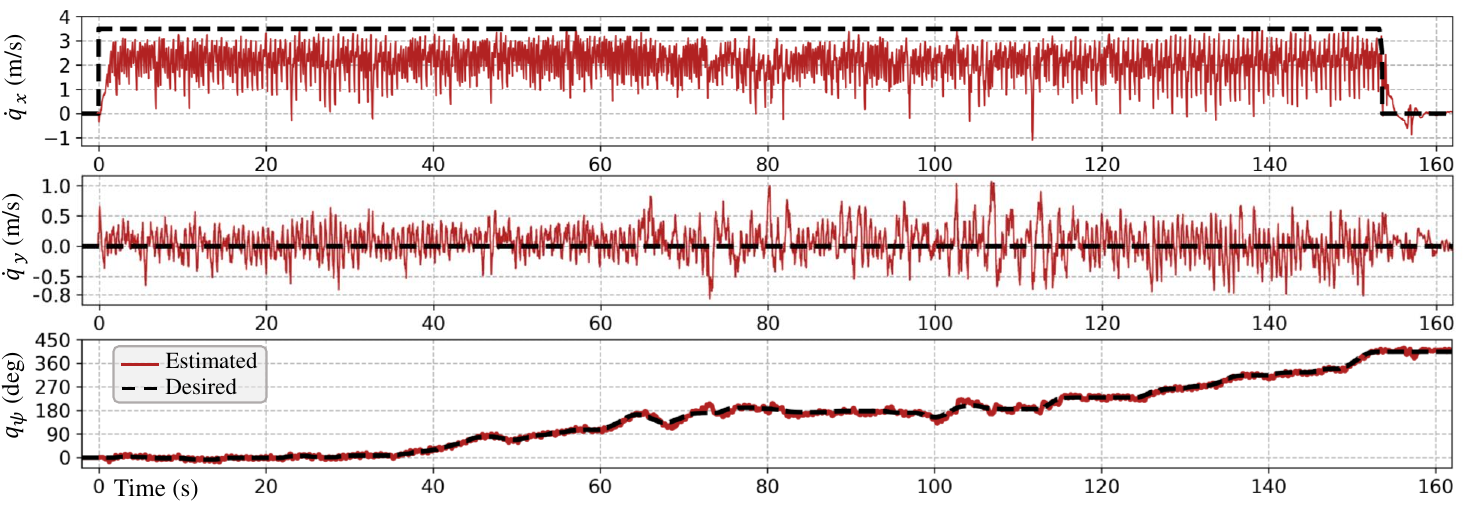}
  \caption{Recorded data during the 400-meter dash. The robot's sagittal velocity $\dot{q}_x$, lateral velocity $\dot{q}_y$ and turning yaw angle $q_\psi$ are shown.}
  \label{subfig:run_400_log}
\end{subfigure}  
\caption{Utilizing a versatile running policy we developed, Cassie successfully completed a 400-meter dash in the bipedal robot regime with a time of 2 minutes and 34 seconds. The dash is completed in the Edwards Stadium at UC Berkeley. This is made possible using a single running policy that enabled the robot to transition from a standing pose to fast running gaits of 2.15 m/s at average and 3.54 m/s at peak. The robot also maintains that speed with an average lateral speed of 0.05 m/s (with zero lateral speed command) throughout the dash. Additionally, the robot is able to accurately follow varying turning commands with an average tracking error (MAE) of 5.95 degrees. This precise turning command tracking was crucial for the successful completion of the 400-meter dash, as it required the robot to reliably turn while running.
During running, we consistently observed noticeable flight phases as examples seen in Fig.~\ref{subfig:run_400_snapshot}. The robot is also able to transit back to standing after it finished the dash, as recorded in the sagittal velocity log (Fig.~\ref{subfig:run_400_log}) and Vid. 2 in Table~\ref{tab:video_list}.}    
\label{fig:400run}
\end{figure*}

\paragraph{Comparison with a Model-based Controller}
We underline the advantages of the proposed walking policy by comparing it with a model-based walking controller, specifically, a model-based controller utilized by the company that manufactured Cassie.
As shown in Fig.~\ref{subfig:push_recovery_mpc}, when the robot is controlled by this model-based controller and subjected to lateral perturbation, this controller fails to maintain control, resulting in a crash. This is because that the model used in the controller does not consider the external perturbation (which is also hard to estimate), and the model-based controller can not deal with such a large modeling error.  
The model-based controller also fails to stabilize the robot during a persistent perturbation test. 
These experiments are recorded in Vid. 3 in Table~\ref{tab:video_list}.
A systematic perturbation test for legged robots was carried out in a very recent work by \cite{van2024revisiting} and it would be interesting to incorporate such tests more widely. 
However, this is beyond the scope of this paper. 
Furthermore, as shown in the video, the model-based controller causes obvious lateral drifts on the robot (before the human operator perturbed it) when it is commanded to track a zero speed without manual gain tuning, while the proposed RL policy shows better tracking performance (that keeps the robot walking in-place) without training using real-world data. However, the model-based controller results in less and favorable stomping force than the RL policy. 
We hope these experiments could help the readers better understand the pros and cons of model-based optimal control and model-free RL methods for bipedal locomotion control.

\subsubsection{Summary of Results}
In summary, the versatile walking policy derived from the proposed method is able to effectively control the bipedal robot Cassie to perform diverse tasks in the real world and be consistent over a long time of usage (more than a year). 
The results show that the robot can track different walking velocities, varying walking heights, turn in different directions, and perform fast walking and transition to and from standing.
The policy also exhibits substantial robustness in the face of challenges such as changes in terrain elevation and external perturbations (including impulse and random but persistent forces).

\begin{figure*}[!htp]
\centering
\begin{subfigure}{0.305\linewidth}
  \centering
  \includegraphics[width=\linewidth]{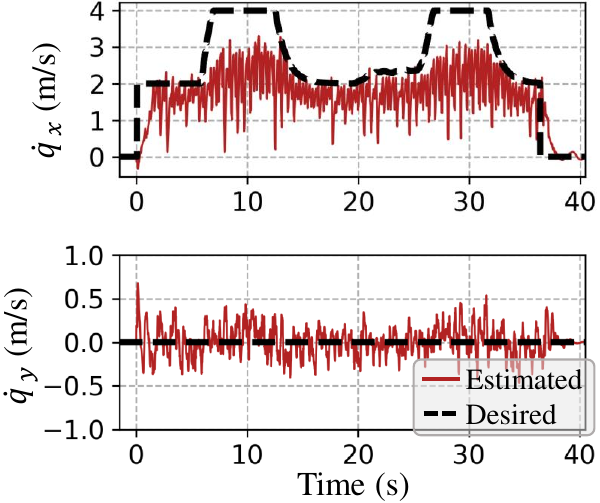}
  \caption{Tracking variable sagittal velocity during running}
  \label{subfig:run_variablevx}
\end{subfigure}
\begin{subfigure}{0.305\linewidth}
  \centering
  \includegraphics[width=\linewidth]{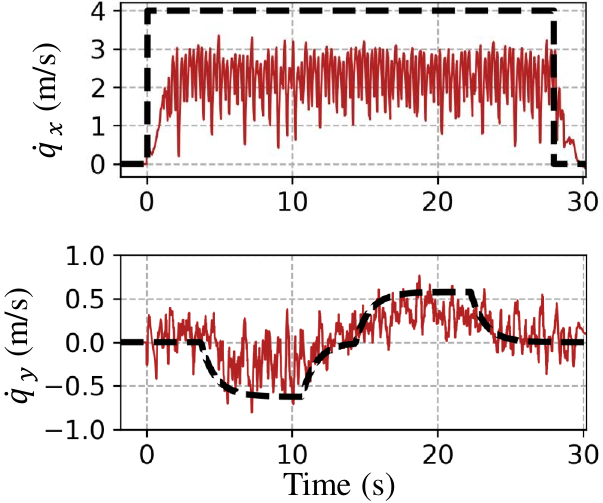}
  \caption{Tracking variable lateral velocity during running}
  \label{subfig:run_variablevy}
\end{subfigure}
\begin{subfigure}{0.375\linewidth}
  \centering
  \includegraphics[width=\linewidth]{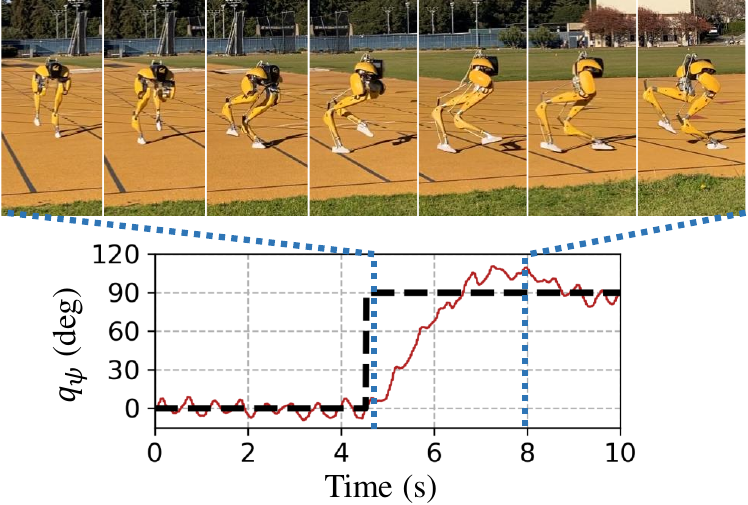}
  \caption{Sharp turn during running. This scenario (tracking nonsmooth turning angle) is not specifically trained.}
  \label{subfig:run_sharpturn}
\end{subfigure}
\caption{Real-world experiments on tracking variable commands while running using the same control policy that completed the 400m-dash in Fig.~\ref{fig:400run}. Using the versatile policy, the robot is able to track varying sagittal velocity $\dot{q}_x$ (Fig.~\ref{subfig:run_variablevx}) and lateral velocity $\dot{q}_y$ (Fig.~\ref{subfig:run_variablevy}). Furthermore, the robot is able to perform a sharp turn while running, as demonstrated in Fig.~\ref{subfig:run_sharpturn}. The robot can turn to 90 degrees within 2 seconds, using 5 steps, as recorded in the logs and corresponding snapshots in Fig.~\ref{subfig:run_sharpturn}. Notably, the robot is not specifically trained for this sharp-turn scenario during training. These experiments are also recorded in Vid. 4 in Table~\ref{tab:video_list}.}
\label{fig:run_variable_cmd}
\end{figure*}

\subsection{Running Experiments}
We now evaluate the versatile running policies developed using the proposed methods in the real world. 
Three specific running policies were obtained: (1) a \emph{general policy} trained on flat ground; (2) a \emph{100-meter-dash finetuned policy}, trained after the convergence of the general policy with a focus on completing a 100-meter dash. This policy incorporated an additional termination condition to end the episode earlier if the robot could not complete the 100-meter dash within 25 seconds; (3) an \emph{uneven-terrain policy}, developed after the convergence of the general policy, integrated with extra terrain randomization as listed in Table~\ref{tab:randomization}.
All the policies have been trained with a variety of running and turning speeds as listed in Table~\ref{tab:command} and transition to and from a standing skill.
As we will see, using the obtained bipedal running policies, we achieved 400-meter dash within 2 min 34 s, 100-meter dash within 27.06 s, running inclines up to $10^\circ$, and more. 

\begin{figure*}[t]
\centering
\begin{subfigure}{\linewidth}
  \centering
  \includegraphics[width=\linewidth]{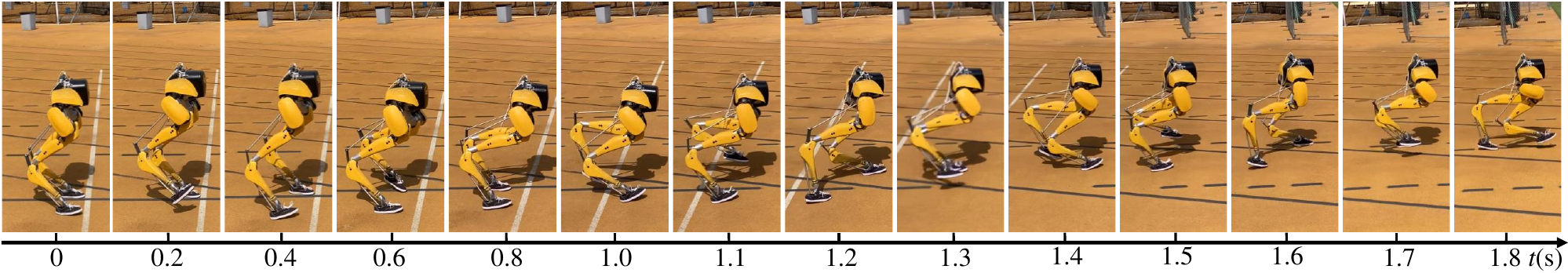}
  \caption{Snapshot of the robot performing transition from standing to running in 100-meter dash, with timestamps indicating the corresponding frames}
  \label{subfig:run_100_init}
\end{subfigure}
\begin{subfigure}{\linewidth}
  \centering
  \includegraphics[width=\linewidth]{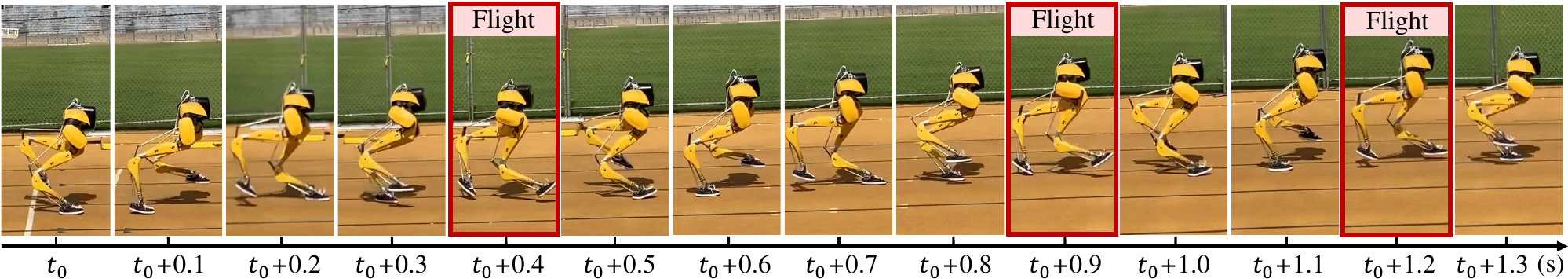}
  \caption{Snapshot of the robot running during 100-meter dash, with timestamps indicating the corresponding frames}
  \label{subfig:run_100_cruise}
\end{subfigure}  
\begin{subfigure}{\linewidth}
  \centering
  \includegraphics[width=\linewidth]{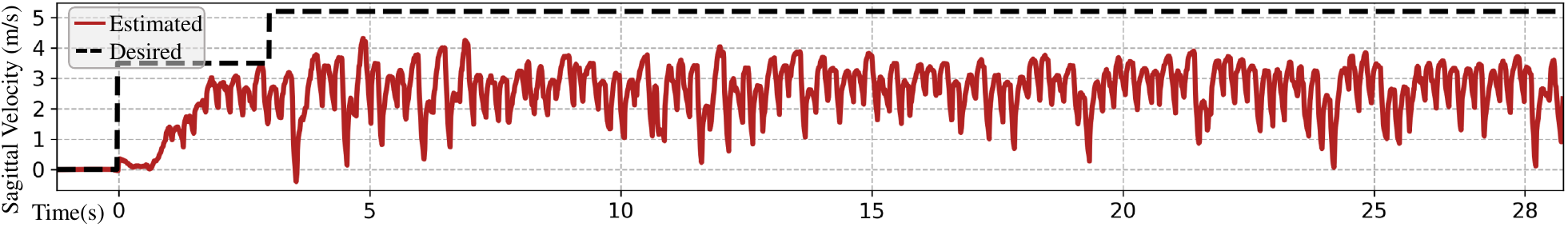}
  \caption{Recorded sagittal velocity $\dot{q}_x$ in the 100-meter dash}
  \label{subfig:run_100_log}
\end{subfigure}  
\caption{Using the finetuned running policy, Cassie finishes the 100-meter dash with a fast running gait. Fig.~\ref{subfig:run_100_init} shows that the robot is able to transit from stationary stance to a rapid running gait, accelerating from 0 m/s to 3 m/s within 2 seconds with aggressive maneuvers. During the cruise phase shown in Fig.~\ref{subfig:run_100_cruise}, the robot maintains a fast yet stable running gait. The corresponding commanded and \emph{estimated} sagittal velocity $\dot{q}_x$ is recorded in Fig.~\ref{subfig:run_100_log}. The robot finished this dash within 28 seconds, with an average speed of 3.57 m/s.}
\label{fig:100run}
\end{figure*}

\subsubsection{Running a 400-meter Dash}
We first evaluate the general running policy to complete a 400-meter dash on a standard outdoor running track, as demonstrated in Fig.~\ref{fig:400run}.
Throughout this test, the robot was commanded to run at a consistent 3.5 m/s while responding to varying turning commands, given by a human operator to navigate the track. 
As shown in Fig.~\ref{subfig:run_400_snapshot} and the accompanying Vid. 2 in Table~\ref{tab:video_list}, the policy is able to smoothly transition from a standing position to a running gait (Fig.~\ref{subfig:run_400_snapshot}\circled{1}). 
The robot managed to accelerate to an average \emph{estimated} running speed of 2.15 m/s\footnote{We note that there is a nontrivial state estimation error on the sagittal velocity when the robot is running at a high speed. The average speed of finishing 400 meter in 154 seconds is 2.6 m/s while the estimated average speed is 2.15 m/s. Such an error is more obvious in the 100-meter dash discussed later. Therefore, we emphasize the logs are estimated values and not true values. We further elaborate this in Appendix~\ref{appendix:estimaion_error}.}, reaching a peak estimated speed of 3.54 m/s, as recorded in Fig.~\ref{subfig:run_400_log}.
The policy successfully maintained the desired speed consistently throughout the entire 400-meter run while accurately adhering to the varying turning commands. 
The average tracking error (MAE) in the turning angle $q_{\psi}$ observed in Fig.~\ref{subfig:run_400_log} is 5.95 degrees.
Throughout the 400m-dash, substantial flight phases were evident, as shown in Fig.~\ref{subfig:run_400_snapshot}\circled{2}-\circled{4}, despite the variable running speeds and turning angles. 
This consistent flight phase distinguishes our running gaits from a fast walking gait and presents significantly more challenges in control. 
However, the controller managed to sustain such a dynamic running gait stably over a significant duration (154 seconds).

Furthermore, during the 400-meter dash, although some lateral drift was detected, as shown in Fig.~\ref{subfig:run_400_log}, the robot can keep an average lateral velocity of 0.05 m/s. 
While the average lateral speed is relatively negligible, the robot did exhibit moments of having higher lateral speeds, exceeding 1 m/s at times, possibly due to irregular contact in the real-world environment. 
However, the policy's training in lateral running skills allowed the robot to correct these deviations, demonstrating the policy's robustness in maintaining stability even in the face of such unexpected lateral movements.

Cassie, controlled by the proposed running policy, successfully finished the 400-meter dash in \emph{2 minutes and 34 seconds}, and was able to transit to a standing pose afterward, as recorded in the video. 
This is a novel capacity of running a full 400-meter lap by a human-sized bipedal robot.

\subsubsection{Tracking Varying Commands while Running}
Using the same versatile policy, the robot is also able to reliably track varying commands, such as sagittal velocity $\dot{q}_x$ shown in Fig.~\ref{subfig:run_variablevx} and lateral velocity $\dot{q}_y$ recorded in Fig.~\ref{subfig:run_variablevy}, while running fast. The corresponding experiments are recorded in Vid. 4 in Table~\ref{tab:video_list}. 
Note that only one dimension of the command is changing during each of the tests conducted in Figs.~\ref{subfig:run_400_log} (turning yaw $q_{\psi}$),~\ref{subfig:run_variablevx} (sagittal velocity),~\ref{subfig:run_variablevy} (lateral velocity). 
By comparing with the recorded logs among these tests, we observed that the command changing in one dimension will not affect the control performance on tracking a constant in other dimensions. 
This indicates that the dimensions on $\dot{q}_x$, $\dot{q}_y$, and $q_{\psi}$ are \emph{decoupled}, which is intriguing as this is realized in fast running control of a highly-nonlinear bipedal robot in the real world.

We further conduct a sharp turning test where the robot is given a step change of the yaw command, from 0 directly to 90 degrees, as recorded in Fig.~\ref{subfig:run_sharpturn}. 
The robot can respond to such a step command and finish a 90-degree sharp turn within 5 steps in 2 seconds, using a natural running gait shown in Fig.~\ref{subfig:run_sharpturn}. 
Remarkably, the robot is not specifically trained for this sharp-turn scenario, as only smooth turning yaw angle command whose changing rate is bounded to 30 deg/s is given during training, as listed in Table~\ref{tab:command}. 
The versatile policy can generalize the learned various turning tasks and direct transfer to accomplish such a sharp-turn test. 
This further highlights the advantage of obtaining a versatile policy as previous work like \cite{yu2022dynamic} requires training a separated policy to perform similar sharp-turn while walking.

\begin{table}[t]
\centering
\caption{Record of the completion time of 100-meter dash in three trials in the real world.}
\label{tab:100run}
\begin{tabular}{cc}
\hline
Trial & Completion Time (s) \\ \hline
1     & 27.06               \\
2     & 27.99               \\
3     & 28.28               \\ \hline
\end{tabular}
\end{table}

\subsubsection{Running a 100-meter Dash}
We now assess the running policy fine-tuned for the 100-meter dash, documented in Fig.~\ref{fig:100run}. We conducted the test 3 times, results of which are detailed in Table~\ref{tab:100run} and available in Vid. 1 and Vid. 4 in Table~\ref{tab:video_list}.
With the deployment of the proposed running policy, the robot accomplished the 100-meter dash with around $28$ seconds, achieving a fastest run time of 27.06 seconds. 
As depicted in Fig.~\ref{subfig:run_100_init}, the robot quickly transitioned from a stationary standing pose to a fast running gait within 1.8 seconds, using the transition maneuvers we commonly observed in human running athletes. 
During the cruising phase (Fig.~\ref{subfig:run_100_cruise}), the robot maintained a rapid running gait and reached a peak estimated speed of 4.2 m/s, showing notable flight phases, as the data recorded in Fig.~\ref{subfig:run_100_log}.

\begin{figure*}[t]
\centering
\begin{subfigure}{0.535\linewidth}
  \centering
  \includegraphics[width=\linewidth]{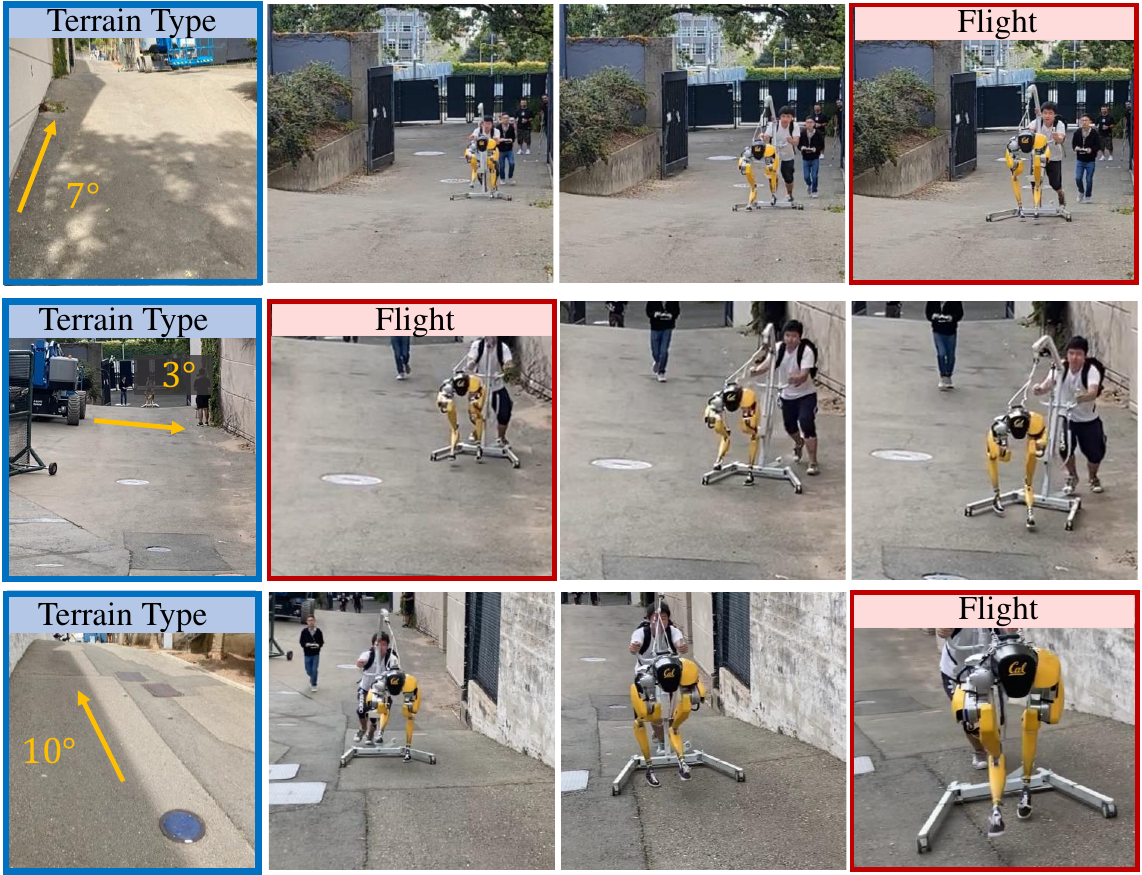}
  \caption{Snapshot of the robot running on varying terrains}
  \label{subfig:run_terrain_snapshot}
\end{subfigure}
\begin{subfigure}{0.41\linewidth}
  \centering
  \includegraphics[width=\linewidth]{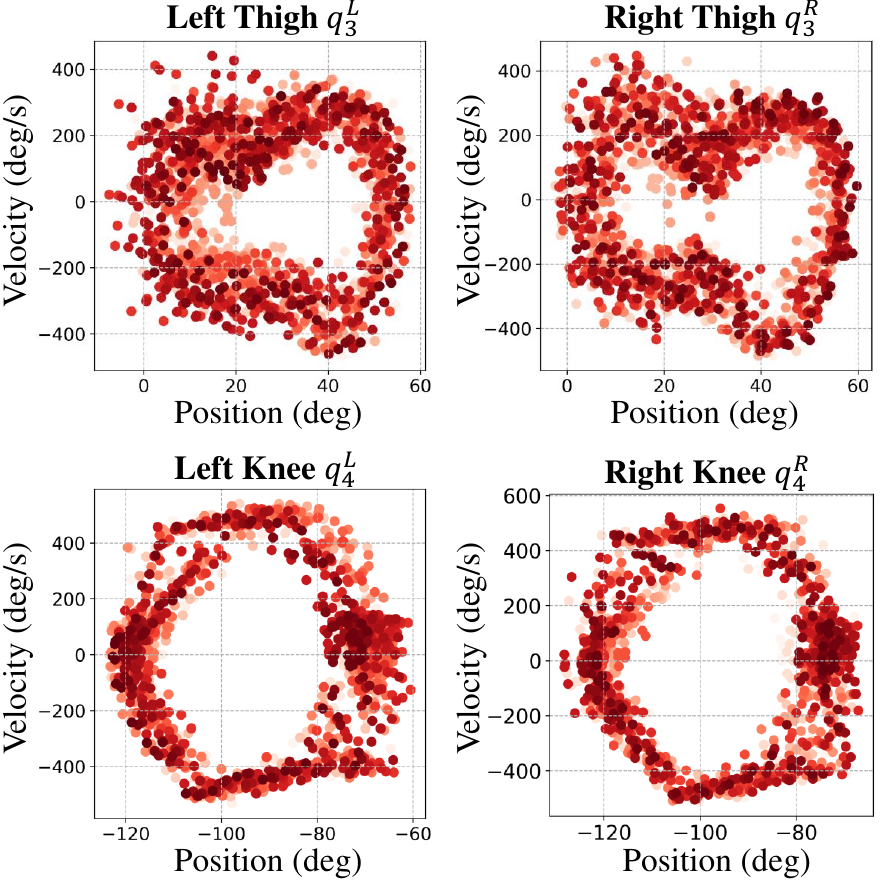}
  \caption{Recorded phase plots of thigh and knee joints}
  \label{subfig:run_terrain_phaseplot}
\end{subfigure}  
\caption{The robot runs over various terrains without any prior knowledge of the terrain specifics or elevation estimates. Fig.~\ref{subfig:run_terrain_snapshot} records the robot running on different types of terrains with different slopes, starting from 7$^\circ$ incline in the sagittal direction, followed by a 3$^\circ$ slope laterally, then a 10$^\circ$ sagittal incline, and finally transitioning to flat ground. The left-most column of the figure showcases these different terrain types. Despite the varying terrain, the robot consistently adapts and maintains a stable running gait with noticeable flight phases, as shown in the right columns of the figure. Fig.~\ref{subfig:run_terrain_phaseplot} records the corresponding phase plots of the robot thigh and knee joints ($q^{L/R}_{3,4}$ versus $\dot{q}^{L/R}_{3,4}$, which play dominant roles in locomotion control), showing converged limit cycles.}    
\label{fig:run_terrain}
\end{figure*}

\begin{figure*}[t]
\centering
\begin{subfigure}{\linewidth}
  \centering
  \includegraphics[width=\linewidth]{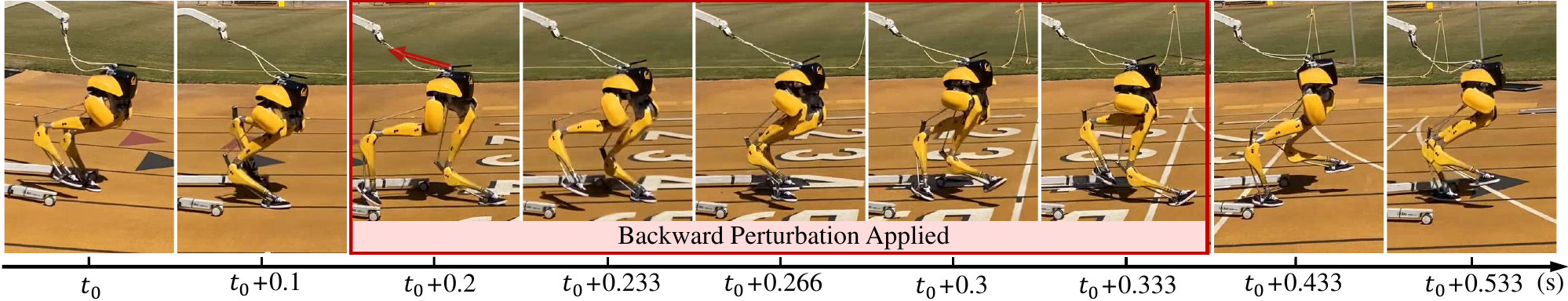}
  \caption{The robot covers from a backward perturbation during running}
  \label{subfig:run_perturb}
\end{subfigure}
\begin{subfigure}{\linewidth}
  \centering
  \includegraphics[width=\linewidth]{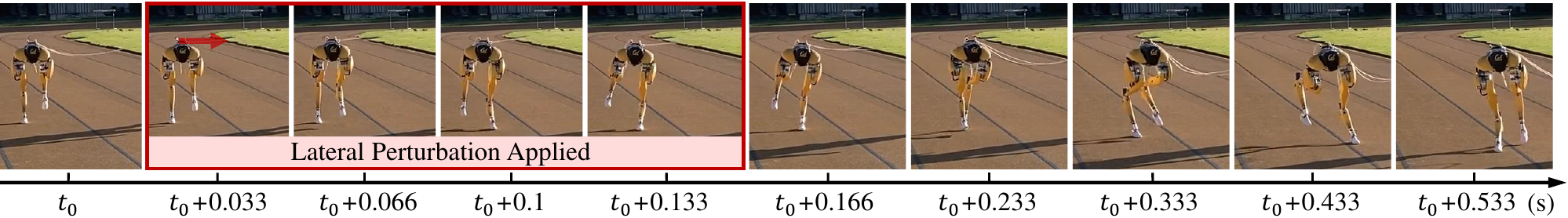}
  \caption{The robot recovers from a lateral perturbation during running}
  \label{subfig:run_lateral_perturb}
\end{subfigure}  
\caption{Snapshot that captures the moment when the robot, while running a 100-meter dash with a fast speed, is unexpectedly perturbed, demonstrating the robustness of the versatile running policy we developed. The frames are aligned with the corresponding timestamps. The robot experiences a sudden decrease in speed and begins to lean laterally along with a significant drift in its base yaw angle, due to being pulled backward by the safety cord. Despite these deviations from its nominal running trajectory, the robustness of the policy, which is enhanced by training with extensive dynamics randomization and versatile tasks (including slower-speed running with lateral and turning commands), enables the robot to maintain stability and correct itself back.}
\label{fig:run_perturb}
\end{figure*}

\subsubsection{Running on Uneven Terrains (Trained)}
We proceed to examine the running policy fine-tuned for uneven terrains. 
As observed in Fig.~\ref{subfig:run_terrain_snapshot}, the robot controlled by the proposed running policy managed to effectively traverse terrains with different slopes, all without the need for explicit terrain height estimation or external sensors.
The terrain tests encompassed challenging variations, beginning with a $7^\circ$ inclined slope in the sagittal plane, followed by a $3^\circ$ slope in the lateral plane, a steeper $10^\circ$ inclined slope in the sagittal plane, and concluding with a flat ground. 
Despite these difficult terrains, the robot maintained a stable running gait, as indicated by the converged limit cycles of the robot's thigh and knee joints ($q^{L/R}_{3,4}$) showcased in Fig.~\ref{subfig:run_terrain_phaseplot} across these varied elevations.
The robot's ability to change its gait to adapt to different terrains is derived from its usage of the robot's input/output history, implicitly encapsulating information about contact events and positioning.
Notably, even while traversing these challenging terrains, the robot retained its flight phase, as showcased in Fig.~\ref{subfig:run_terrain_snapshot}, signifying its ability to maintain a running gait, avoiding a degradation to walking gaits that offer more support but reduced speed.
This is the first implementation of running (with flight phases) over large uneven terrains by a human-sized bipedal robot.

\begin{figure*}[t]
\centering
\begin{subfigure}{0.495\linewidth}
  \centering
  \includegraphics[width=\linewidth]{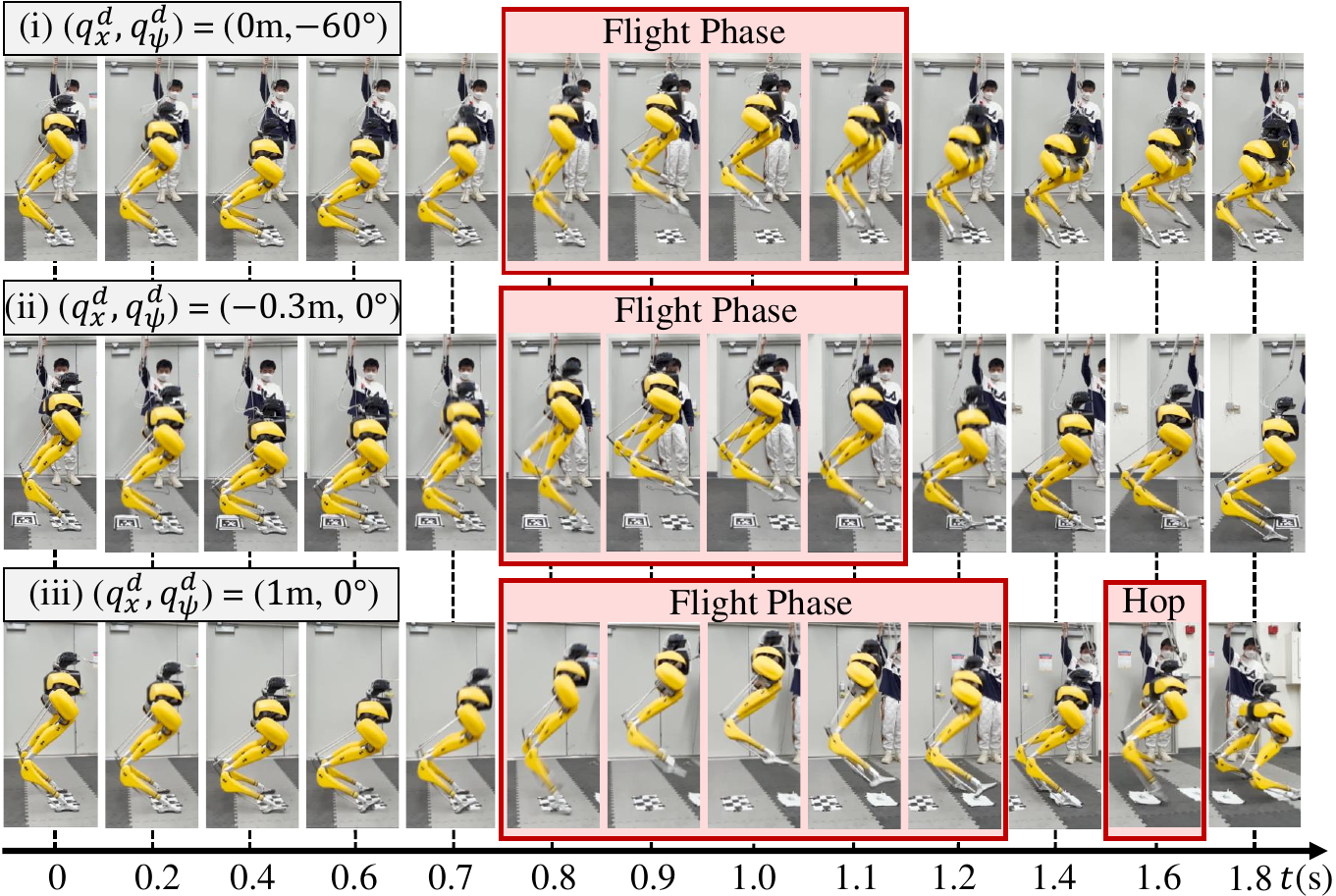}
  \caption{Different jumps using the flat-ground policy}
  \label{subfig:snapshot_flatx3}
\end{subfigure}
\begin{subfigure}{0.495\linewidth}
  \centering
  \includegraphics[width=\linewidth]{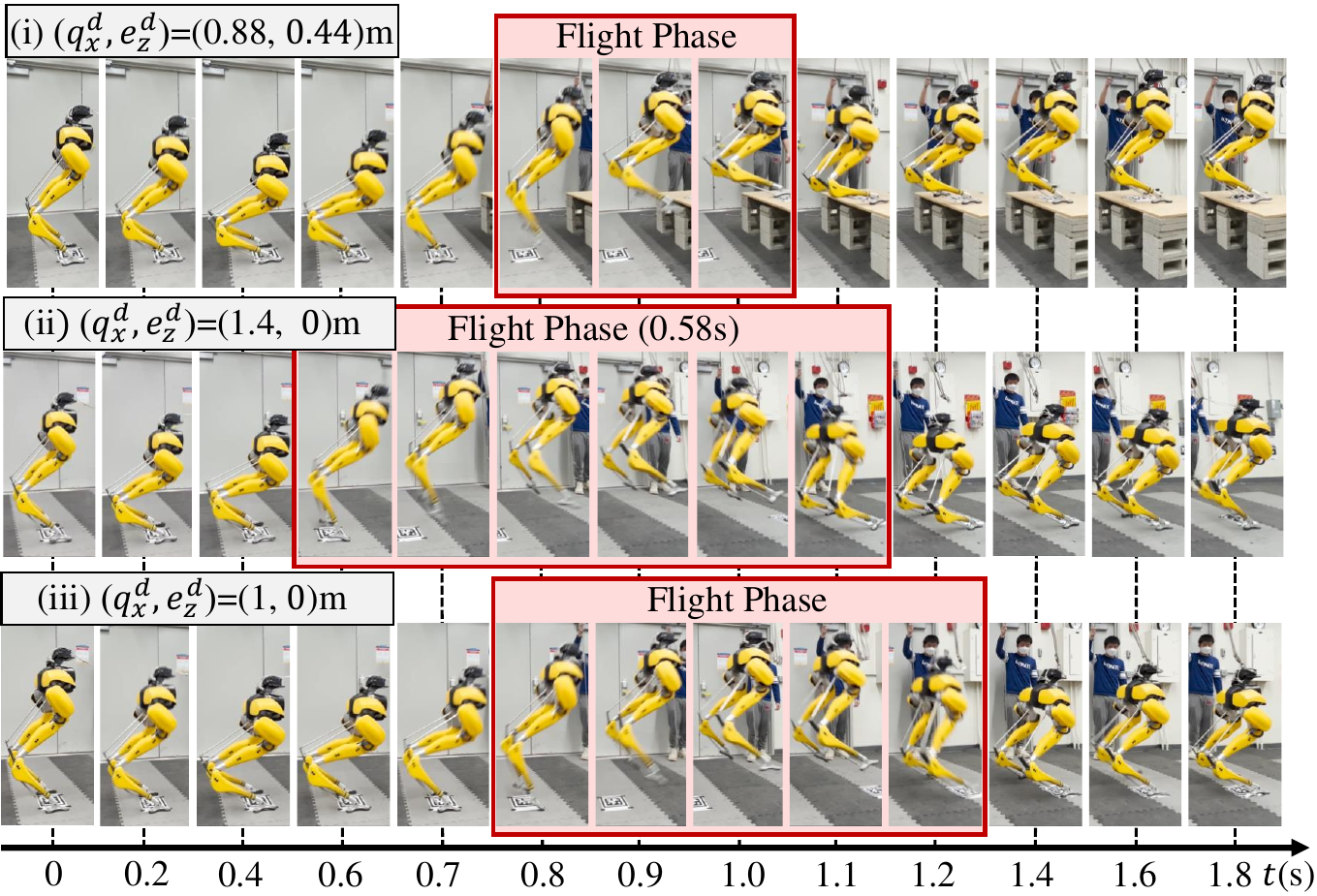}
  \caption{Different jumps using the discrete-terrain policy}
  \label{subfig:snapshot_tablex3}
\end{subfigure}
\caption{Snapshots of the bipedal robot Cassie performing various aggressive jumping maneuvers using two single versatile policies. Each frame is aligned with the corresponding timestamp, and paper tags in the figures represent the desired landing target. In Fig.~\ref{subfig:snapshot_flatx3}, using the \textit{flat-ground} policy, (i) the robot is able to perform in-place jump while turning, (ii) backward jump, and (iii) forward jump, and land accurately on the target (paper tag). During the 1-meter ahead jump, the robot interestingly performs an extra forward hop after its initial landing, reaching the target on this second attempt. This additional maneuver was not observed during simulations that utilized the robot's nominal model. In Fig.~\ref{subfig:snapshot_tablex3}, using the \textit{discrete-terrain} policy, (iii) the robot is also able to perform 1-meter forward jump. Moreover,  (ii) the robot can perform a large standing long jump that lands accurately at 1.4-meter ahead, and (i) jump to a high platform that is elevated 0.44-meter above and 0.88-meter ahead.}
\label{fig:jump_snapshot_timeline}
\end{figure*}

\subsubsection{Robust Running Maneuvers}

During the real-world deployment, the robustness of the developed running policies in handling various perturbation scenarios was observed. 
For example, in Fig.~\ref{subfig:run_perturb}, the robot is controlled by its 100-meter-dash fine-tuned policy when an abrupt impulse perturbation force, produced by the safety cord, causes the robot to suddenly drop speed and lean to the left and twist. 
Despite this unforeseen event, the robot, trained on simulated perturbations and diverse running tasks including turning, is able to maintain stability and quickly recover back to a stable running gait. 
A similar test is also conducted using the general running policy while a lateral perturbation is applied to the robot, as recorded in Fig.~\ref{subfig:run_lateral_perturb}. 
Without losing balance, the robot exerts a lateral running gait to compensate such a lateral perturbation.  
These experiments are recorded in Vid. 4 in Table~\ref{tab:video_list}.
These scenarios underline the robustness of the \emph{versatile} running policies obtained by the proposed \emph{task randomization}, especially beneficial when addressing unexpected disturbances while the robot is running at high speeds in real-world scenarios.

\subsubsection{Summary of Results}
In summary, the running policies derived from the proposed method exhibit effective control over the bipedal robot Cassie in real-world scenarios. They showcase the capacity to manage various running and turning velocities, adapt to changes in terrain based solely on proprioceptive feedback, and seamlessly transition from and to a standing pose. Because of these running policies, Cassie achieves a notable peak speed of 4.2 m/s in the 100-meter dash which was completed within 27.06 seconds, accomplishes a 400-meter dash in 2 minutes 34 seconds, and traverses over uneven terrains, while showing robustness to unexpected distributions.

\subsection{Jumping Experiments}\label{subsec:exp_jumping}
We now evaluate the proposed versatile jumping policies. We obtained two policies: (1) the \textit{flat-ground} policy that specializes in jumping to different locations and executing turns on level ground ($e^d_z=0$); and (2) \textit{discrete-terrain} policy that focuses on leaping to various elevated platforms without executing turns ($q^d_\psi=0$).
The rationale behind having these two separate policies stems from the challenges posed by the dynamic jumping skill. 
In our empirical findings, we found it is difficult to train a robot to perform both jumping onto elevated platforms while turning simultaneously.
As we will see, the proposed jumping policies achieve a large variety of different bipedal jumps including a 1.4-meter long jump and jumping onto a 0.44-meter elevated platform.

\subsubsection{Jump and Turn}

\begin{figure*}[t]
\centering
\begin{subfigure}{0.3225\linewidth}
  \centering
  \includegraphics[width=\linewidth]{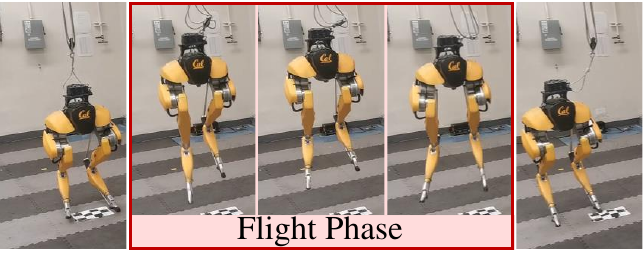}
  \caption{$(q^d_x, q^d_y, q^d_\phi)$ = (0m, -0.3m, 0$^\circ$)}
  \label{subfig:jump_flat_lat03}
\end{subfigure}
\begin{subfigure}{0.67\linewidth}
  \centering
  \includegraphics[width=\linewidth]{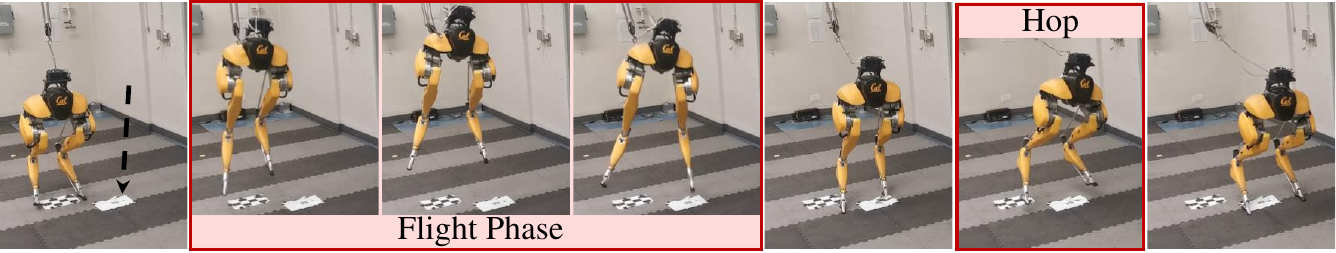}
  \caption{$(q^d_x, q^d_y, q^d_\phi)$ = (0.3m, 0.3m, 0$^\circ$)}
  \label{subfig:jump_flat_lat03sag03}
\end{subfigure}
\begin{subfigure}{\linewidth}
  \centering
  \includegraphics[width=\linewidth]{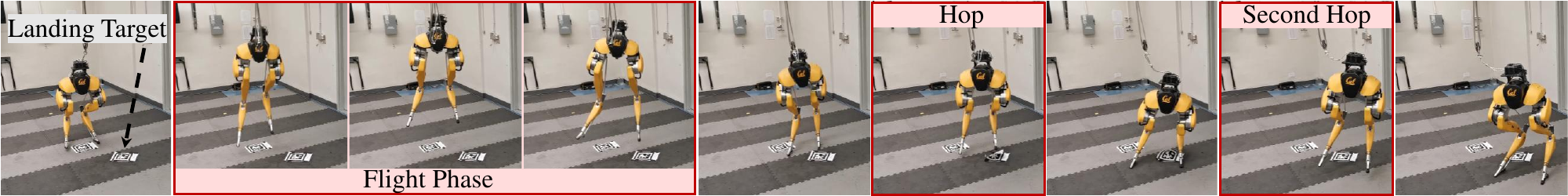}
  \caption{$(q^d_x, q^d_y, q^d_\phi)$ = (0.5m, 0.2m, -45$^\circ$)}
  \label{subfig:jump_flat_multiaxes}
\end{subfigure}
\begin{subfigure}{0.345\linewidth}
  \centering
  \includegraphics[width=\linewidth]{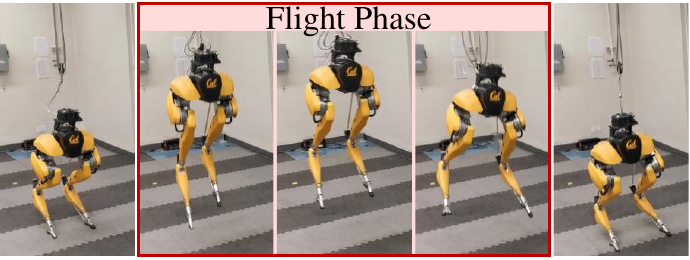}
  \caption{$(q^d_x, q^d_z)$ = (0m, 0m)}
  \label{subfig:jump_table_inplace}
\end{subfigure}
\begin{subfigure}{0.647\linewidth}
  \centering
  \includegraphics[width=\linewidth]{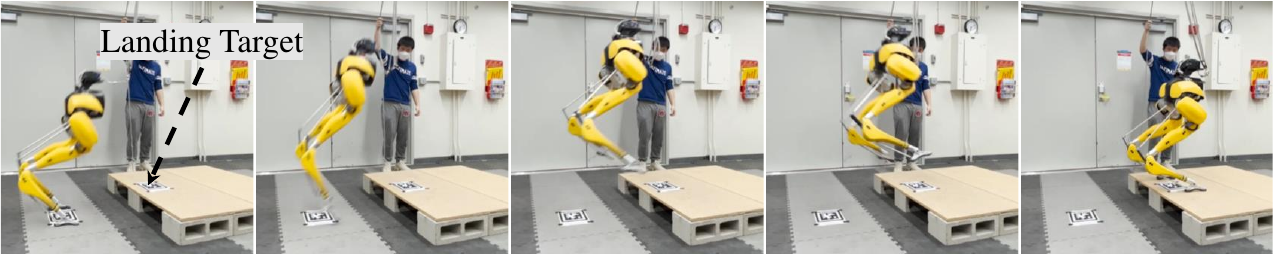}
  \caption{$(q^d_x, q^d_z)$ = (0.88m, 0.17m)}
  \label{subfig:jump_table_088017}
\end{subfigure}
\begin{subfigure}{0.495\linewidth}
  \centering
  \includegraphics[width=\linewidth]{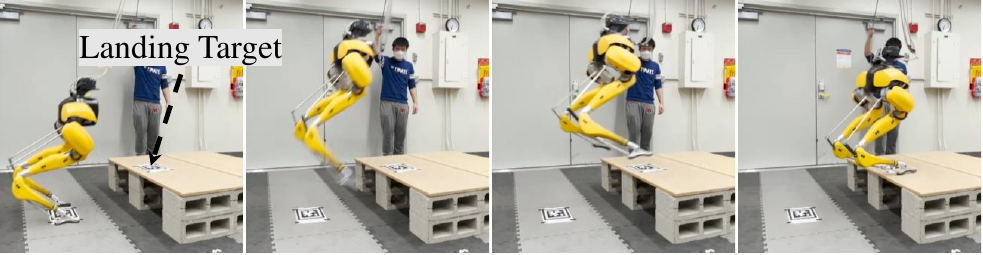}
  \caption{$(q^d_x, q^d_z)$ = (0.88m, 0.32m)}
  \label{subfig:jump_table_088032}
\end{subfigure}
\begin{subfigure}{0.495\linewidth}
  \centering
  \includegraphics[width=\linewidth]{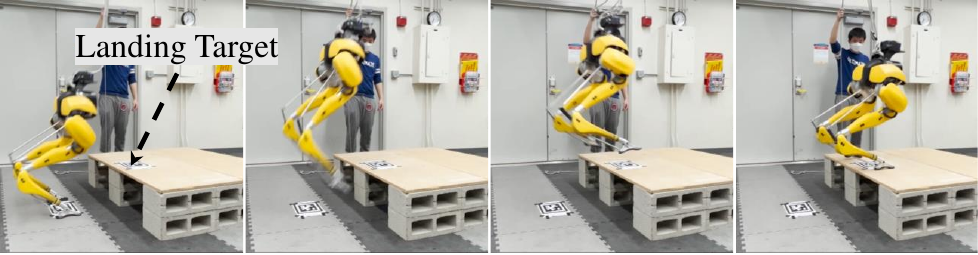}
  \caption{$(q^d_x, q^d_z)$ = (0.64m, 0.32m)}
  \label{subfig:jump_table_064032}
\end{subfigure}
    \caption{Snapshots of more diverse bipedal jumps using the proposed versatile jumping policies. The paper tag on the ground indicates the landing target. Using the same flat-ground policy used in Fig.~\ref{subfig:snapshot_flatx3}, the robot is also able to perform a lateral jump (0.3m to its right in Fig.~\ref{subfig:jump_flat_lat03}), diagonal jump (0.3m ahead and 0.3m to its left in Fig.~\ref{subfig:jump_flat_lat03sag03}), and a complex jumping maneuver that blends forward (0.5m), lateral (0.2m) and turning ($-\text{45}^\circ$) at the same time (Fig.~\ref{subfig:jump_flat_multiaxes}). Using the same discrete-terrain policy used in Fig.~\ref{subfig:snapshot_tablex3}, the robot can also jump in place (Fig.~\ref{subfig:jump_table_inplace}), jump to targets that are placed (Fig.~\ref{subfig:jump_table_088017}) 0.88-meter ahead and 0.17-meter above, (Fig.~\ref{subfig:jump_table_088032}) 0.88-meter ahead and 0.32-meter above, and (Fig.~\ref{subfig:jump_table_064032}) 0.64-meter ahead and 0.32-meter above, respectively.}    
\label{fig:various_jump}
\end{figure*}

We first assess the performance of the \emph{flat-ground} policy, as documented in Fig.~\ref{subfig:snapshot_flatx3}. 
Using this single jumping policy, the robot executes various given target jumps, such as jumping in place while turning $60^\circ$ (Fig.~\ref{subfig:snapshot_flatx3}i), jumping backward to land 0.3-meter behind (Fig.~\ref{subfig:snapshot_flatx3}ii), and jumping to a target placed 1 meter ahead (Fig.~\ref{subfig:snapshot_flatx3}iii), by just changing the command to the policy.
The robot is able to accomplish the task by landing precisely on the target which is marked as a paper tag on the ground. 
Upon closer observation, the robot can adjust its take-off pose in order to follow different commands. 
For instance, it leans backward when required to land on the rear target, and leans forward when executing a forward jump.
The robot is also capable of executing a more diverse set of jumps that combine movements along different axes. 
For instance, Fig.~\ref{subfig:jump_flat_multiaxes} records a representatitve jump where the robot performs a multi-axes jump where it's required to jump forward, laterally, and make a turn at the same time. 
We report more diverse types of jumps (12 different jumps in total) in Appendix~\ref{appendix:other_jumps} and Vid. 5 in Table~\ref{tab:video_list}, and all of these jumps are realized by this single flat-ground jumping policy.

\begin{figure*}[t]
\centering
\begin{subfigure}{0.72\linewidth}
  \centering
  \includegraphics[width=\linewidth]{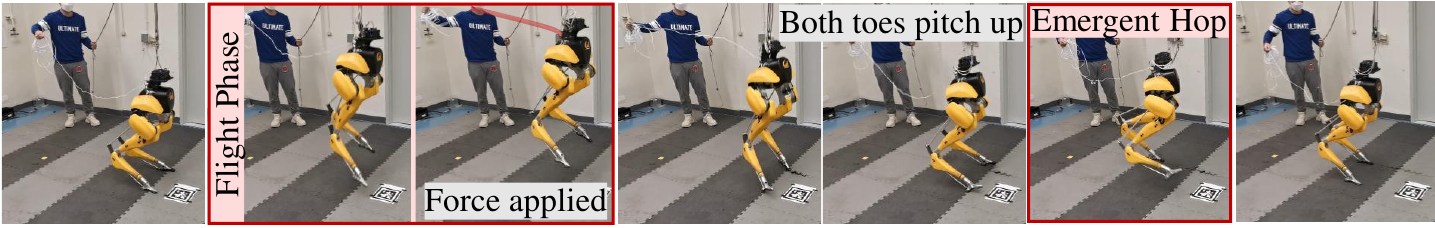}
  \caption{$(q^d_x, q^d_y, q^d_\psi)$ = (0m, 0m, 0$^\circ$), with perturbation force applied during flight.}
  \label{subfig:jump_inplace_perturb}
\end{subfigure}
\begin{subfigure}{0.2735\linewidth}
  \centering
  \includegraphics[width=\linewidth]{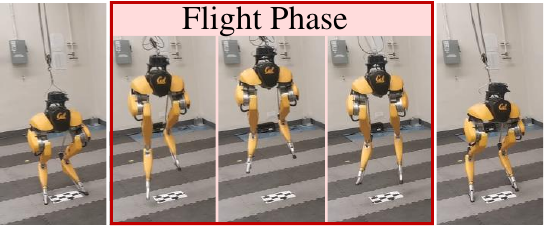}
  \caption{$(q^d_x, q^d_y, q^d_\psi)$ = (0m, 0m, 0$^\circ$)}
  \label{subfig:jump_inplace}
\end{subfigure}
\caption{Snapshots of the robustness test during jumping using the versatile flat-ground policy. In Fig.~\ref{subfig:jump_inplace_perturb}, an unexpected backward perturbation is applied while at its apex height during an in-place jump. This contrasts with Fig.~\ref{subfig:jump_inplace}, where the robot performs the same task without perturbation, serving as a baseline. Despite leaning backward upon landing due to the perturbation, the robot spontaneously executes a backward hop, a maneuver it learned during other tasks like backward jumping, and succeeds to maintain balance after landing. Such a robustness is achieved through \textit{task randomization} alone, as the jumping policy was not trained with simulated perturbation.}    
\label{fig:jump_perturb}
\end{figure*}

\subsubsection{Jump to Elevated Platforms}
We then evaluate the \emph{discrete-terrain} policy to control the robot's jumping ability over various distances and elevations. 
As showcased in Fig.~\ref{subfig:snapshot_tablex3}, the robot exhibits the capability to jump precisely to targets placed at different positions, like 1 meter ahead (Fig.~\ref{subfig:snapshot_tablex3}iii) or 1.4 meters ahead (Fig.~\ref{subfig:snapshot_tablex3}ii), and at varying elevations, including 0.44 meters elevated (Fig.~\ref{subfig:snapshot_tablex3}i, considering the robot height is only 1.1 meter). 
Using the single discrete-terrain policy, the robot is able to adjust its take-off maneuvers for different jumping targets and efficiently manages its angular momentum after the large impact upon landing to maintain stability. 
In addition to standing long jumps (1.4 meters forward) and standing high jumps (0.44 meters elevated), we conducted a range of varied jumping tasks within this range, including jumping in place (with foot clearance around 0.4 meters at its apex height) and jumping over different sagittal distances onto various elevated platforms. 
These tasks were all conducted using the same policy, as shown in Fig.~\ref{fig:various_jump}.
Notably, the standing long jump over 1.4 meters and standing high jump to a 0.44-meter elevated platform (while using the same controller) are the novel locomotion capacities in the human-sized bipedal robot regime.

\subsubsection{Robust Jumping Maneuvers}
We illustrate the robustness of the jumping policy obtained from the proposed method with an example shown in Fig.~\ref{fig:jump_perturb}. Employing the \emph{flat-ground} policy, we commanded the robot to perform an in-place jump. 
At the apex of the jump, we applied a backward impulse perturbation, as depicted in Fig.~\ref{subfig:jump_inplace_perturb}.
This significant perturbation caused the robot to deviate from its nominal in-place jumping trajectory (as seen in Fig.~\ref{subfig:jump_inplace}), resulting in a significant change in its pose upon landing. Consequently, the robot landed with a backward lean and its toes pitched up trying to stabilize the robot. 
This pose presented a significant challenge, given that the robot became underactuated and was nearly losing balance.
However, since the jumping policy has been trained to execute backward jumps, as previously showcased in Fig.~\ref{subfig:snapshot_flatx3}ii, the robot rapidly adjusted its intended landing trajectory and executed a backward hop. By this adjustment, the robot is able to better correct its pose in the air to achieve a more favorable configuration upon its next landing. 

Note that the jumping policies were never trained explicitly to handle perturbations. 
Therefore, this experiment highlights the advantage of a versatile jumping policy. 
The robot demonstrates the ability to generalize its learned diverse tasks to devise a better maneuver and contact plan rather than strictly adhering to the given task during real-world deployment. 
This serves as a detailed report of successful recovery from a perturbation while a bipedal robot is executing a jump in the real world, illustrating the importance of the \textit{task randomization}.

\begin{remark}
Using versatile policies for walking, running, and jumping, the robot exhibits the capability to develop its own contact strategy online, deviating from the contact plans (implicitly) provided by reference motions. This ability enhances stability and robustness. For instance, in jumping experiments, the robot often breaks contact after landing and executes small hops to achieve a better landing configuration, as seen in Fig.~\ref{fig:jump_perturb}.
This capacity is also evident in other skills, such as standing (Fig.~\ref{subfig:bm_multi_walking}), walking (with significant double-support phases contrary to the single-support in reference motions, as in Fig.~\ref{subfig:walking_slope}), and running (where the robot does not strictly follow the contact sequence of the periodic running reference motion, as seen in Fig.~\ref{fig:100run}).
This emergent capability aligns with what contact-implicit optimization, like in \cite{posa2014direct,drnach2021robust,landry2022bilevel}, aims to achieve. While these piror methods have only achieved such optimization offline for bipedal robots, our work realizes this online on a real bipedal robot.
\end{remark}

\begin{remark}
We note that during the jumping experiments, the robot occasionally oscillates while standing after a large jump. 
This is due to the challenges of learning a single RL policy to learn both dynamic aperiodic jumping skill and stationary standing skill. 
The features learned during a jump may favor high-acceleration motion and cannot easily alter the policy's behavior for perfectly stationary standing. 
This also provides an insight into the challenges of obtaining a single unified policy to learn all different locomotion skills (to combine dynamic and stationary skills). 
But we want to note that, the jumping policy is still able to damp out such an oscillation during standing, though with a longer noticeable time, as demonstrated in the supplementary video (Vid. 5). 
\end{remark}

\subsubsection{Summary of Results}
We demonstrated 19 distinct bipedal jumps in the real world, encompassing a range of landing locations, turns, and elevations. 
These jumps were conducted using just two versatile policies: the \emph{flat-ground} policy and the \emph{discrete-terrain} policy.
Through these extensive real-world experiments, we illustrated two key aspects of the proposed method, using the jumping skill as an exemplary case. 
First, we showcase the \emph{adaptivity} of the proposed policy. 
The robot exhibits the capability to land precisely on a designated target after a flight phase, all without requiring global position feedback. 
To achieve this, the robot must generate a precise amount of momentum at take-off. 
By accomplishing these diverse jumping tasks, the RL-based control policy showcases its adaptivity to the robot's hardware dynamics, mainly by leveraging the long-term I/O history, to control the robot and achieve accurate take-off translational velocities to land at the target.
Second, we highlight the \emph{robustness} of the policy. 
Despite being trained without perturbation, the robot demonstrates an ability to respond to unexpected perturbations during jumping by employing learned tasks for agile recovery.
\section{How to Train Your Bipedal Robot: A Discussion}\label{sec:discussion}
After detailing our methodology, analysis, and experiments on robust, dynamic, and versatile control policies for a spectrum of bipedal locomotion skills via RL, we now turn to a discussion of the key lessons learned throughout this development. 
Our objective is to provide valuable and generally applicable insights that could steer future research in the use of RL for locomotion control in complex dynamic systems, particularly in bipedal and humanoid robots.

The following section is structured as follows: (1) We first discuss the importance of utilizing the robot's I/O history, and how to effectively use it (by using a dual-history policy to learn direct adaptive control) in Sec.~\ref{subsec:diss_history}. (2) In Sec.~\ref{subsec:diss_robustness}, we show that while properly using robot's I/O history empowers the controller with adaptivity, the robustness of the controller can be improved by task randomization. (3) In Sec.~\ref{subsec:diss_general}, we provide several “bonuses" that RL can bring forth, such as motion inference and optimization, as well as contact planning. We wrap up this discussion with a small debate on general versus task-specific control policies. 

\subsection{How to Use the Robot's History?}\label{subsec:diss_history}
\paragraph{Robot's Long I/O History is Important}
In existing RL-based locomotion control literature, there is no convergence on how to use the robot's history: short or long, history of only state feedback or of state and action pair. 
Our research demonstrates the clear benefits of providing a long history of both robot's inputs and outputs, especially in the case of high-dimensional highly nonlinear systems like bipedal robots.  
This approach is grounded in the robot's full-order dynamics \eqref{eq:full_order_dynamics}, suggesting that a sequence of I/O history can be more informative in identifying the dynamic system to control.
This assertion is supported by our ablation study (Fig.~\ref{fig:dynrand_three}), where policies incorporating a longer I/O history outperformed those with shorter history or states-only long history in handling randomized dynamics parameters in realizing different locomotion skills. 
Additionally, analyzing the latent embedding from the long I/O history encoder (Fig.~\ref{fig:latent_vis}) revealed that it could implicitly capture information such as contact events, external forces, and changes in system dynamics parameters, and more.
This finding is intriguing, as it shows that optimization by model-free RL, with a proper utilization of robot's I/O, can learn to extract vital information for dynamic control.

\paragraph{Short History Complements Long History}\label{subsec:dual_history_advantages}
While a long I/O history can improve control performance, its effectiveness is limited without proper formulation. 
Our ablation study (Fig.~\ref{fig:dynrand_three}) shows that using only a long I/O history does not outperform policies with a short history, aligned with the report from \cite{singh2023learning}.
We discover that to fully exploit the benefits of long history, it's effective to provide a short I/O history directly to the base MLP, bypassing the long history encoder, termed as \emph{dual history} approach.
This modification markedly improves learning performance across various locomotion skills (Fig.~\ref{fig:dynrand_three}) and outperforms other approaches in real-world experiments (Fig.~\ref{fig:bm_realworld}). 
While long history has been introduced in RL-based locomotion control in prior work, the complementary use of short history is a novel approach. 
Many previous efforts lack this addition to the base policy, potentially overlooking a critical element that is the \emph{recent} I/O history for controlling complex robots in \textit{real time}. 
When only using long I/O history, recent events may become obfuscated in the latent embedding after the long-history encoding, and this can be addressed by incorporating an \emph{explicit} short I/O history alongside the long-history encoder. 
This explicit recent I/O feedback can also help with the non-Markovian property when we opt to use a long robot I/O history.
Furthermore, while the long history is better for system identification and state estimation, the short-term features are better for denoising the high-frequency estimates from the long-history encoder.
Although this approach might not be essential for simpler systems like stationary robotic arms~\citep{peng2018sim} or quadrupedal robots~\citep{lee2020learning} as observed in prior studies, for more complex dynamic systems like the bipedal robots in our work, the benefits of this dual history approach are clearly pronounced.
Additionally, while the dual-history approach could boost the learning using non-recurrent policy that is trained to leverage the history sequence with an explicit length, we found it is hard to aid the training of recurrent policy in terms of the bipedal locomotion control. Details are presented in Appendix~\ref{appendix:lstm_tcn}.

\paragraph{Are We Learning A Robust or Adaptive Controller?}
Adaptivity in RL is not a new concept, and there are notable efforts from the control theory community to bridge adaptive control with RL, like~\cite{annaswamy2023adaptive}. 
On the other hand, many studies on RL-based locomotion control have demonstrated robustness in real-world applications, attributed by some to the robustness of the policy trained with dynamics randomization and by others to the adaptivity in RL.  
Many in this community are curious about which attribute actually plays a critical role. 
This curiosity motivated us to conduct an extensive empirical study in this work to investigate both the adaptivity and robustness of the RL policy. 
We show that in the RL-based controller, these two aspects can originate from different sources: adaptivity can arise from the proper use of the robot's I/O history, while robustness can be introduced by task randomization. 
Dynamics randomization occupies a middle ground: in order to be robust to different dynamics, the policy can be trained with randomized dynamics to learn to effectively utilize the history (if provided) in adapting to the dynamics shifts.

\paragraph{What Do We Want to Learn, Direct or Indirect Adaptive Control?}
Recognizing the value of properly employing long I/O history, we provide a discussion about an alternative method: using the history to estimate selected environmental parameters for control. 
This approach, known as the Teacher-Student (TS) training strategy~\citep{lee2020learning} or RMA~\citep{kumar2021rma}, can be seen as an RL-version of \emph{indirect} adaptive control. 
However, for real-time dynamic locomotion control, our findings suggest that a \emph{direct} adaptive control method, which integrates long I/O history directly into the controller without explicitly estimating model parameters, yields better performance, supported by the results in Sec.~\ref{sec:policy_structure}.
The underlying reason behind this becomes clearer when considering the capabilities of the history encoder (Fig.~\ref{fig:latent_vis}) obtained by our proposed method. 
It not only adapts to time-invariant modeling parameters but also captures crucial information deemed important by the robot, like time-variant contact events. 
This contrasts with TS or RMA methods, which are limited to estimating modeling parameters and might encounter estimation errors or failures in challenging tasks like bipedal running (Fig.~\ref{fig:dynrand_three}). 
Compared to TS and RMA which separate training for the base control policy and history encoder, end-to-end training without expert supervision gives the robot more freedom to explore and exploit useful information.

\subsection{Versatility Improves Robustness}\label{subsec:diss_robustness}
Besides the adaptivity brought by the proper use of the robot's long I/O history, another key benefit of using RL for locomotion control is robustness. 
When policies are trained with \emph{dynamics randomization}, they can stay robust to environmental changes, like during the sim-to-real transfer. 
However, our research reveals that the robustness of RL policies extends beyond this, with \emph{task randomization} emerging as another key strategy.

Our ablation study (Fig.~\ref{fig:bm_single_sim}) supports this. 
While dynamics randomization expands the range of trajectories a robot is trained on within a specific task, it doesn't dramatically alter the training distribution. 
For instance, extensive randomization in a standing policy doesn't equip the robot with new locomotion skills like walking or jumping for recovery (Figs.~\ref{fig:robust_standing},~\ref{fig:robust_standing_run_jump}). 
Task randomization, however, imparts robustness differently. 
By training across diverse tasks, such as varying walking speeds or jumping targets, the robot learns to generalize and exploit these learned tasks, aiding recovery even without specific dynamics randomization, as demonstrated in Fig.~\ref{fig:bm_single_sim}.
Interestingly, the robot's response to disturbances differs based on the source of robustness. 
With extensive dynamics randomization, it tends to adhere to the commanded task, whereas using a versatile policy trained with task randomization, it shows more \emph{compliance} to external disturbances, deviating from its given command.

Although some previous RL-based locomotion research unintentionally achieved versatile policies and attributed their success of robustness primarily to dynamics randomization, our study distinguishes the sources of robustness. 
We recommend additional task randomization in future RL applications for robust locomotion control.

\subsection{Exploring What RL Can Enable for Legged Locomotion}\label{subsec:diss_general}
In this subsection, we aim to offer broader insights into the potential (and limitations) of RL for controlling complex dynamic systems.

\paragraph{Trajectory Optimization and Motion Inference}
By using RL for controlling high-dimensional bipedal robots, reference motions are often used to achieve natural gaits, as demonstrated in our work. 
While there are prior attempts using trajectory optimization based on robot dynamics model to obtain a reference motion, like~\cite{bogdanovic2022model} and the walking policy in this work, we also show that such a “pre-optimization" could be unnecessary for RL-based control.
For example, our work with bipedal running and jumping, using only kinematically feasible retargeted human motion and animation, shows that robots can learn to stay close to reference motions and maintain dynamic stability through trial and error, as long as the action space is not corrupted by the added reference motion (as the residual approach in Fig.~\ref{fig:dynrand_three}).
This indicates the ability of RL to jointly learn trajectory optimization and real-time control. 
Furthermore, our work demonstrates that RL also enables robots to infer motions beyond the provided references. 
For example, the robot can learn to vary running or jumping maneuvers from single reference motions, achieved by task randomization and goal-conditioned policy.
However, using a single reference motion has its limits. 
For instance, if we only provide a reference forward jumping motion, it becomes challenging for the robot to learn a backward jump based solely on task completion reward. 
Therefore, selecting an effective single reference motion may require it to be relatively unbiased, such as the jumping-in-place motion used in this work.

\paragraph{Motion (and Contact) Planning}
In our real-world experiments (Sec.~\ref{sec:experiments}), we observed numerous instances where the robot demonstrated the ability to perform a sequence of varied motions with varying contact sequences for recovery. 
For example, when perturbed from a standing position, the robot exerted a variety of different walking gaits in order to return to a stand, all without requiring an online motion or contact sequence scheduler (Fig.~\ref{subfig:stand_recover_by_walk}). 
This reflects a longer-horizon motion planning capability, distinct from its real-time locomotion control occurring at 33 Hz.
This is further supported by jumping experiments like Figs.~\ref{subfig:jump_flat_multiaxes},~\ref{subfig:jump_inplace_perturb}. 
These observations indicate that RL policies can autonomously develop contact sequences for different tasks during deployment when provided with the unvaried reference motion.
This capability also suggests that specifying a contact sequence to the policy can be less advantageous, as the (legged) robot will have less freedom to explore more optimal contact patterns for enhanced stability and robustness.
\emph{In short, the advantage of utilizing RL for legged locomotion control where contact is crucial is in the ability to control without explicitly considering contact.}

\paragraph{The Possibility of Learning a Unified Control Policy} 
Given that all the policies developed in this work use the same control architecture and training procedure, it is possible to obtain an unified policy for different skills.
This possibility is further supported given that in all dynamic walking, running, and jumping policies, we have combined them with a distinguished skill: the stationary standing. 
For example, the policy that realized the transition among running and standing skills shows the potential to learn different skills (such as slower speed walking) within this spectrum.   
Adding the standing skill with the other is realized by using a carefully designed MDP and an additional substage for the new skill training as introduced in Appendix~\ref{appendix:transition_to_stand}. 
However, if using this relatively simple method, it is challenging to keep adding different new skills as the robot may suffer from the catastrophic forgetting problem. 

Using an adversarial motion prior (AMP) as in~\cite{peng2021amp} could potentially help to obtain a unified control policy for diverse locomotion skills without specifically tuning motion tracking rewards. However, applying AMP to aggressive real-world bipedal locomotion remains a challenge. GAN-styled methods are prone to mode-collapse, and can struggle to imitate aggressive motions that occur over brief periods. The actor policy can “cheat" the discriminator by mimicking a parts of a jump, without performing the motion in its entirety. This is further compounded by the large sim-to-real gap for bipedal robots.  

Another potential way to develop such a unified policy can be done by continual RL to keep learning new skills like \cite{liu2023continual} or imitation learning from offline datasets collected by skill-specific policies like \cite{huang2024diffuseloco}. 
However, increasing the robustness and overcoming the sim-to-real gap using these methods for bipedal robots is still an open question. A practical way could be learning to transition among different pre-trained skill-specific policies, such as \cite{lee2018composing}. 
In either way, this proposed method for skill-specific policies in this work serves as a solid starting point.

\paragraph{Generalization versus Precision}
In our research, we have showcased many examples of RL's generalization capabilities in dynamic locomotion control, spanning various tasks and dynamics parameters. 
While control precision has not been a primary focus, it becomes evident in skills like bipedal jumping. 
In these tasks, the robot successfully jumps to specified targets (Sec.~\ref{subsec:exp_jumping}), requiring precise translational velocity at take-off for accurate landing. 
However, attaining perfect precise control using one policy that handles a wide variety of tasks and dynamics variations remains an open question. For example, it is challenging to track a specific sagittal velocity with minor errors in fast-running tasks in the real world (Fig.~\ref{subfig:run_100_log}).
Yet, the benefits of a generalized locomotion control policy, like foundation models, lie in providing a solid starting point for further fine-tuning on specific downstream tasks, like precision control.
On the flip side, a controller optimized for precision might be restricted to the finetuned task.
We present these trade-offs for readers to consider, highlighting the balance between generalization and precision in RL-based dynamic locomotion control.
\section{Conclusion and Future Work}\label{sec:conclusion}
In conclusion, this work presents a comprehensive study on using deep reinforcement learning to develop versatile, robust, and dynamic locomotion controllers for bipedal robots. 
This work introduces a dual-history approach that integrates the robot's input/output (I/O) history into RL-based controllers and highlights its significance. 
We demonstrate the adaptivity and robustness of the proposed RL-based controller, particularly underscoring how a well-designed long I/O history encoder can adapt to both time-invariant dynamics changes and time-variant events such as varying contacts. 
Additionally, task randomization, which encourages the robot to explore a broader range of scenarios by accomplishing different tasks, significantly enhances robustness, complementing the robustness achieved through traditional dynamics randomization.
The proposed method is validated thoroughly on the bipedal robot Cassie, successfully realizing versatile and robust walking, running, and jumping skills in the real world. 
These experiments include several state-of-the-art results, including walking control with consistent performance over a long timespan (459 days), versatile running capabilities demonstrated in a 400-meter dash and across challenging terrains, and a large repertoire of jumping tasks including the furthest 1.4-meter forward jump and 0.44-meter high jump.

In the future, we hope this method can be extended to humanoid robots that can also leverage upper-body motions for agility and stability. 
Additionally, integrating depth vision directly into the locomotion controller can be realized in a straightforward way where an additional depth encoder can work alongside the robot's I/O history encoder in our proposed control architecture.  
We encourage the readers to also try this setup for visual control on their bipedal robots. 
With the advancements made in this work, we think a large portion of practical problems in realizing effective locomotion control for human-sized bipedal robots can be addressed. 
For the bipedal robotics community, an exciting direction would be the combination of bipedal locomotion and bimanual manipulation to tackle long-horizon loco-manipulation tasks, opening new possibilities in the field.

\section*{Acknowledgments}
This work was in part supported by NSF Grant CMMI-1944722, The AI Institute, and Canadian Institute for Advanced Research (CIFAR).
The authors thank Xuxin Cheng for the help in the initial development of this work. The authors thank Daniel Wong for the design of running shoes for Cassie. The authors would also like to thank Bike Zhang, Jiaming Chen, Lizhi Yang, Xiaoyu Huang, Yiming Ni, Yufeng Chi, Akshay Thirugnanam, Yuanzhuo Li, Dr. Jun Zeng, and Dr. Ayush Agrawal for their help with the experiments. 

\balance
\bibliographystyle{IEEEtran}
\bibliography{references}
\pagebreak

\appendix{

\subsection{Advantages of Using an Action Filter}\label{appendix:lpf}
We used a Low Pass Filter (LPF) after the policy output, acting as an action filter. The LPF used in this work is a Butterworth low pass filter with a cut-off frequency of 4 Hz. LPF has been widely applied in control engineering including model-based optimal control methods~\citep{gong2022zero} and model-free RL methods~\citep{peng2020learning,li2021reinforcement,smith2022legged,escontrela2022adversarial} for legged locomotion. 
LPF can further smooth out the action from the RL-based policy, which further complements the smoothing rewards listed in Table~\ref{tab:reward}.
We conducted an ablation study by training a jumping-in-place policy (single-task training) \emph{without} LPF, which still results in jumping but with worse learning performance (converged return) due to the resulting jittering motion, as shown in Fig.~\ref{subfig:lpf_benchmark}. 
If we take out the LPF, we will need to increasing the weight on the smoothing term to highlight the importance of the smoothness of the entire motion. However, if the smoothing term takes too much weight in the entire reward composition, the robot may easily learn a suboptimal behavior by being stationary without exploring dynamic skills such as jumping. Therefore, LPF can help with reducing such reward tuning burden.
Furthermore, since LPF only helps to damp out the high-frequency changes from the action (policy's output), the entire system may still have oscillation (\textit{e.g.}, due to a large joint velocity), having the smoothing reward is still necessary to smooth the entire closed-loop system. These two parts complement each other.

\begin{figure}[!]
\centering
\includegraphics[width=0.75\linewidth]{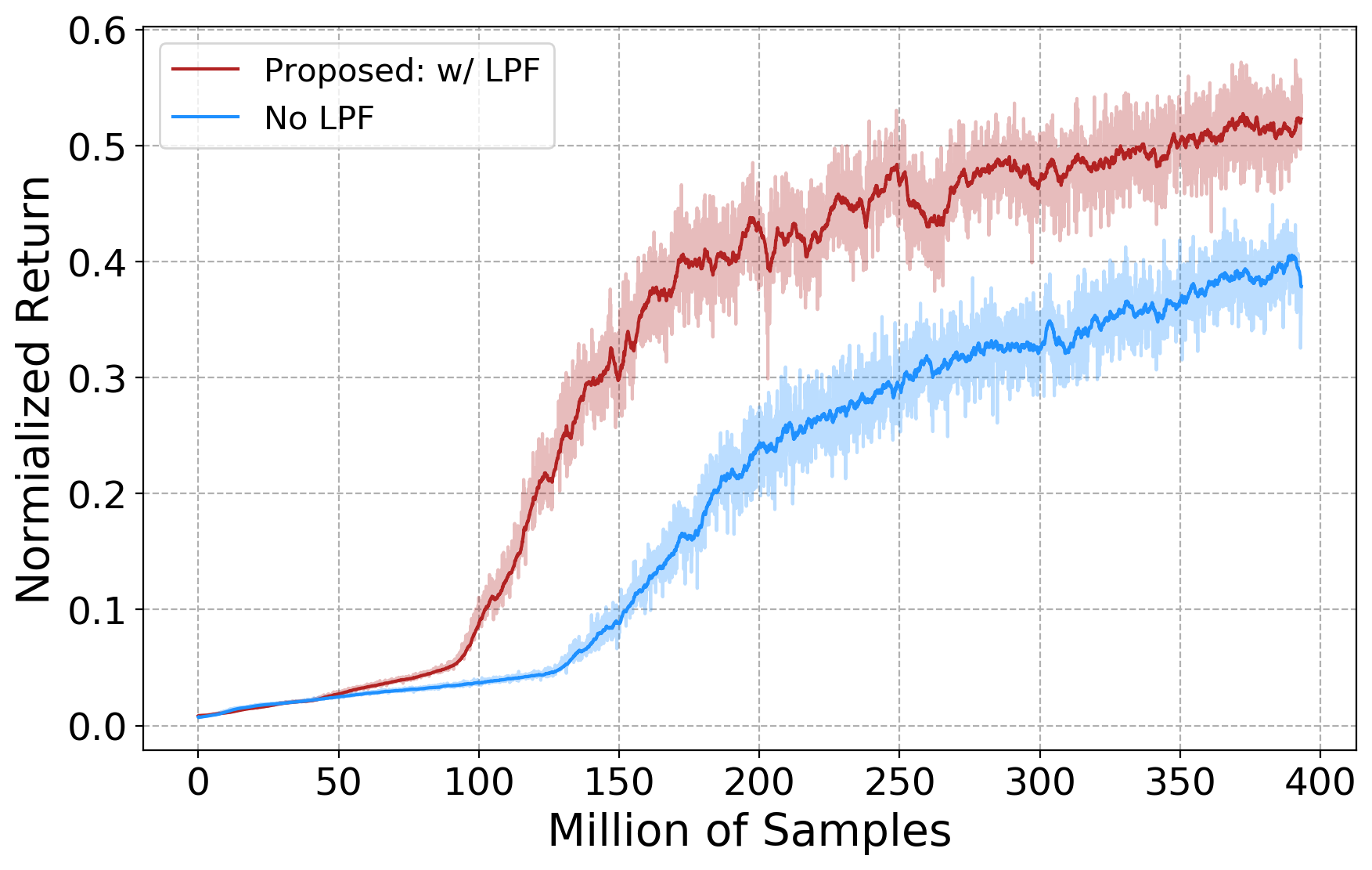}
\caption{Ablation study on the use of Low Pass Filter (LPF) as an action filter for the training of the in-place jumping skill from scratch. Without using the LPF, the training return (blue curve) is much lower than the one using LPF (ours, red curve). These two policies are obtained by using the exact hyperparameters and training settings. The underlying reason is that it is harder for the RL-based policy to damp out high-frequency jittering motion without the use of LPF.}\label{subfig:lpf_benchmark}
\end{figure}

\subsection{Learning to Combine a Standing Skill}\label{appendix:transition_to_stand}
Enabling the robot to learn standing along with other locomotion skills is useful for real-world applications, as it is essential to retrieve the physical robot hardware (while the robot is not moving). 

For the walking skill, after the robot has mastered the diverse walking tasks after the task randomization stage, we added a sub-stage. 
In the episode of this sub-stage of training, a standing skill is commanded after a random timespan of walking, and \emph{will last until the end of episode}. The robot is informed to perform standing by two changes: (1) the reference motion input to the control policy is changed to a nominal standing motion (the preview of reference motion is now a stack of three same standing motor positions), and (2) the reward is formulated to encourage performing stationary standing skill. 
The reward for standing has the same formulation used for training dynamic locomotion skills listed in Table~\ref{tab:reward}, and the weights of smoothing terms, such as motor velocity and change of action, are increased with other terms remaining unchanged.
The increased smoothing weight is to encourage the robot to maintain a stationary standing pose. 
After the robot has acquired the skills of walking, standing, and the transition from an arbitrary state from walking to standing, we can move on to dynamics randomization. 
Starting from this stage, the standing skill will be only introduced in the middle of the training of walking and last for a random finite timespan. In this way, the robot can also learn the transition back from standing to walking.  

For the running skill, the POMDP design for this sub-stage is similar to the one for walking. However, since transiting from fast running to standing is more challenging, we provide an additional reference motion of transition from running to standing, retargeted from human mocap. This could facilitate the training during the transition phase. 
For the jumping skill, since it is an aperiodic skill and the robot is required to maintain a standing pose after landing, we do not need such an additional sub-stage of training.       

This method may also be useful to combine different locomotion skills (like combining all walking, running, and jumping) using one single policy. This could potentially solve the problem of developing multi-skill policy for highly-dynamic legged locomotion control. 
We note this as a possible extension of this work.

\subsection{Command Range for Different Skills}\label{appendix:command}

\begin{table}[!ht]
\centering
\scriptsize
\caption{Command range for different skills.}
\label{tab:command}
\begin{tabular}{|cc|}
\hline
\multicolumn{1}{|c|}{\textbf{Task Parameters}}          & \textbf{Range}        \\ \hline
\multicolumn{2}{|c|}{\textbf{Walking}}                                          \\ \hline
\multicolumn{1}{|c|}{Sagittal Velocity $\dot{q}^d_x$}   & {[}-1.5, 1.5{]} m/s   \\ \hline
\multicolumn{1}{|c|}{Lateral Velocity $\dot{q}^d_y$}    & {[}-0.6, 0.6{]} m/s   \\ \hline
\multicolumn{1}{|c|}{Turning Velocity $\dot{q}^d_\psi$} & {[}-45, 45{]} deg/s   \\ \hline
\multicolumn{1}{|c|}{Walking Height $q^d_z$}            & {[}0.65, 1.0{]} m     \\ \hline
\multicolumn{2}{|c|}{\textbf{Running}}                                          \\ \hline
\multicolumn{1}{|c|}{Sagittal Velocity $\dot{q}^d_x$}   & {[}2.0, 5.0{]} m/s    \\ \hline
\multicolumn{1}{|c|}{Lateral Velocity $\dot{q}^d_y$}    & {[}-0.75, 0.75{]} m/s \\ \hline
\multicolumn{1}{|c|}{Turning Velocity $\dot{q}^d_\psi$} & {[}-30, 30{]} deg/s   \\ \hline
\multicolumn{2}{|c|}{\textbf{Jumping}}                                          \\ \hline
\multicolumn{1}{|c|}{Sagittal Landing Location $q^d_x$} & {[}-0.5, 1.5{]} m     \\ \hline
\multicolumn{1}{|c|}{Lateral Landing Location $q^d_y$}  & {[}-1.0 1.0{]} m      \\ \hline
\multicolumn{1}{|c|}{Turning Direction at Landing $q^d_\psi$} & {[}-100, 100{]} deg \\ \hline
\multicolumn{1}{|c|}{Change of Elevation $e^d_z$}       & {[}-0.5, 0.5{]} m     \\ \hline
\end{tabular}
\end{table}

The training ranges for the command of three locomotion skills (walking, running, and jumping) developed in this work are detailed in Table ~\ref{tab:command}. 
During a training episode, the given command is drawn uniformly from the listed range.  

\subsection{Hyperparameters Used in Training}\label{appendix:hyperparameters}

\begin{table*}[!]
\scriptsize
\centering
\caption{Number of Training Iterations used for Different Locomotion Skills. Each iteration collects a batch of 65536 samples.}
\label{tab:num_iter}
\begin{tabular}{|ccccc|}
\hline
\multicolumn{5}{|c|}{\textbf{Walking}} \\ \hline
\multicolumn{1}{|c|}{Single-Task} & \multicolumn{1}{c|}{Task Randomization} & \multicolumn{1}{c|}{Combining Standing} & \multicolumn{1}{c|}{Dynamics Randomization} & Added Perturbation Training \\ \hline
\multicolumn{1}{|c|}{6000} & \multicolumn{1}{c|}{8000} & \multicolumn{1}{c|}{2000} & \multicolumn{1}{c|}{8000} & 5000 \\ \hline
\multicolumn{5}{|c|}{\textbf{Running}} \\ \hline
\multicolumn{1}{|c|}{Single-Task} & \multicolumn{1}{c|}{Task Randomization} & \multicolumn{1}{c|}{Combining Standing} & \multicolumn{1}{c|}{Dynamics Randomization} & Added Perturbation Training \\ \hline
\multicolumn{1}{|c|}{6000} & \multicolumn{1}{c|}{18000} & \multicolumn{1}{c|}{5000} & \multicolumn{1}{c|}{15000} & 5000 \\ \hline
\multicolumn{5}{|c|}{\textbf{Jumping}} \\ \hline
\multicolumn{1}{|c|}{Single-Task} & \multicolumn{1}{c|}{Task Randomization} & \multicolumn{1}{c|}{} & \multicolumn{1}{c|}{Dynamics Randomization} &  \\ \hline
\multicolumn{1}{|c|}{6000} & \multicolumn{1}{c|}{12000} & \multicolumn{1}{c|}{} & \multicolumn{1}{c|}{20000} &  \\ \hline
\end{tabular}
\end{table*}
\begin{table}[!]
\scriptsize
\centering
\caption{Hyperparameters used in PPO Training. These are consistent among different locomotion skills.}
\label{tab:hyperparam_ppo}
\begin{tabular}{|l|l|}
\hline
\textbf{Hyperparameter} & \textbf{Value} \\ \hline
PPO iteration batch size & 65536 \\ \hline
PPO clip rate & 0.2 \\ \hline
Optimization step size (both actor and critic) & $1e^{-4}$ \\ \hline
Optimization batch size & 8192 \\ \hline
Optimization epochs & 2 \\ \hline
Discount factor ($\gamma$) & 0.98 \\ \hline
GAE smoothing factor ($\lambda$) & 0.95 \\ \hline
\end{tabular}
\end{table}

The number of training (PPO) iterations used to develop different locomotion skills is detailed in Table~\ref{tab:num_iter}.
Furthermore, the hyperparameters of PPO are reported in Table~\ref{tab:hyperparam_ppo}, and they are consistent over different locomotion skills.

\subsection{Comparison of Use of Different History Lengths}\label{appendix:history_length}

\begin{figure}[ht]
\centering
\includegraphics[width=0.75\linewidth]{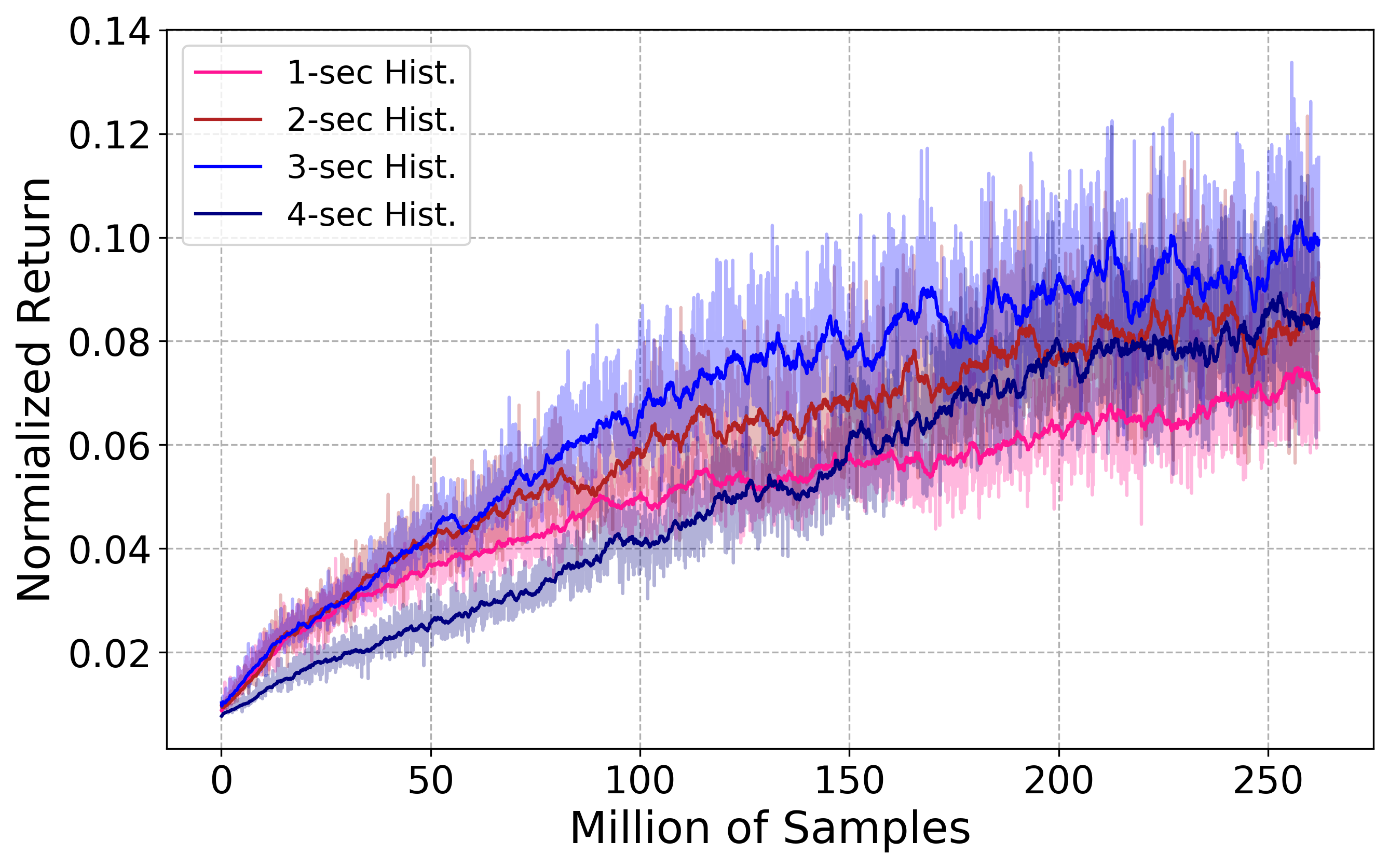}
\caption{Learning performance using different lengths of the robot's I/O history when training a single-task running policy with dynamics randomization. All of these policies used the proposed dual-history-based policy. When increasing the explicit length of robot history from 1 second (pink curve), 2 seconds (red curve, as used in this work), and 3 seconds (blue curve), we observe an increase in learning performance. However, if we keep increasing the history length, such as to 4 seconds (dark blue curve), the improvement of the learning performance may get saturated.}
\label{fig:hist_compare}
\end{figure}

\begin{figure}[ht]
\centering
\begin{subfigure}{0.475\linewidth}
  \centering
  \includegraphics[width=\linewidth]{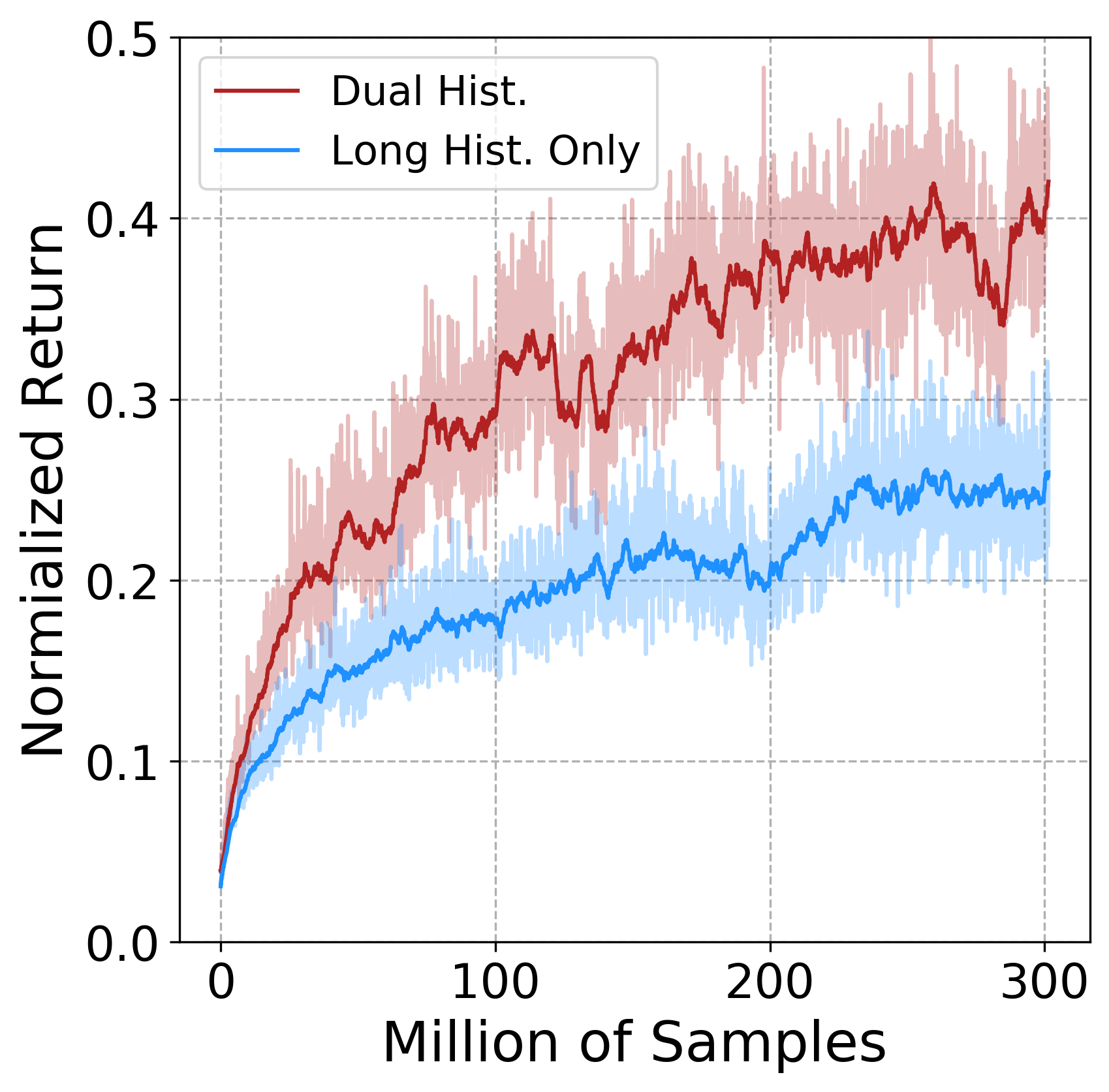}
  \caption{TCN with dual history approach (Dual Hist.) and the TCN only (Long Hist. Only).}
  \label{subfig:tcn_compare}
\end{subfigure}
\begin{subfigure}{0.475\linewidth}
  \centering
  \includegraphics[width=\linewidth]{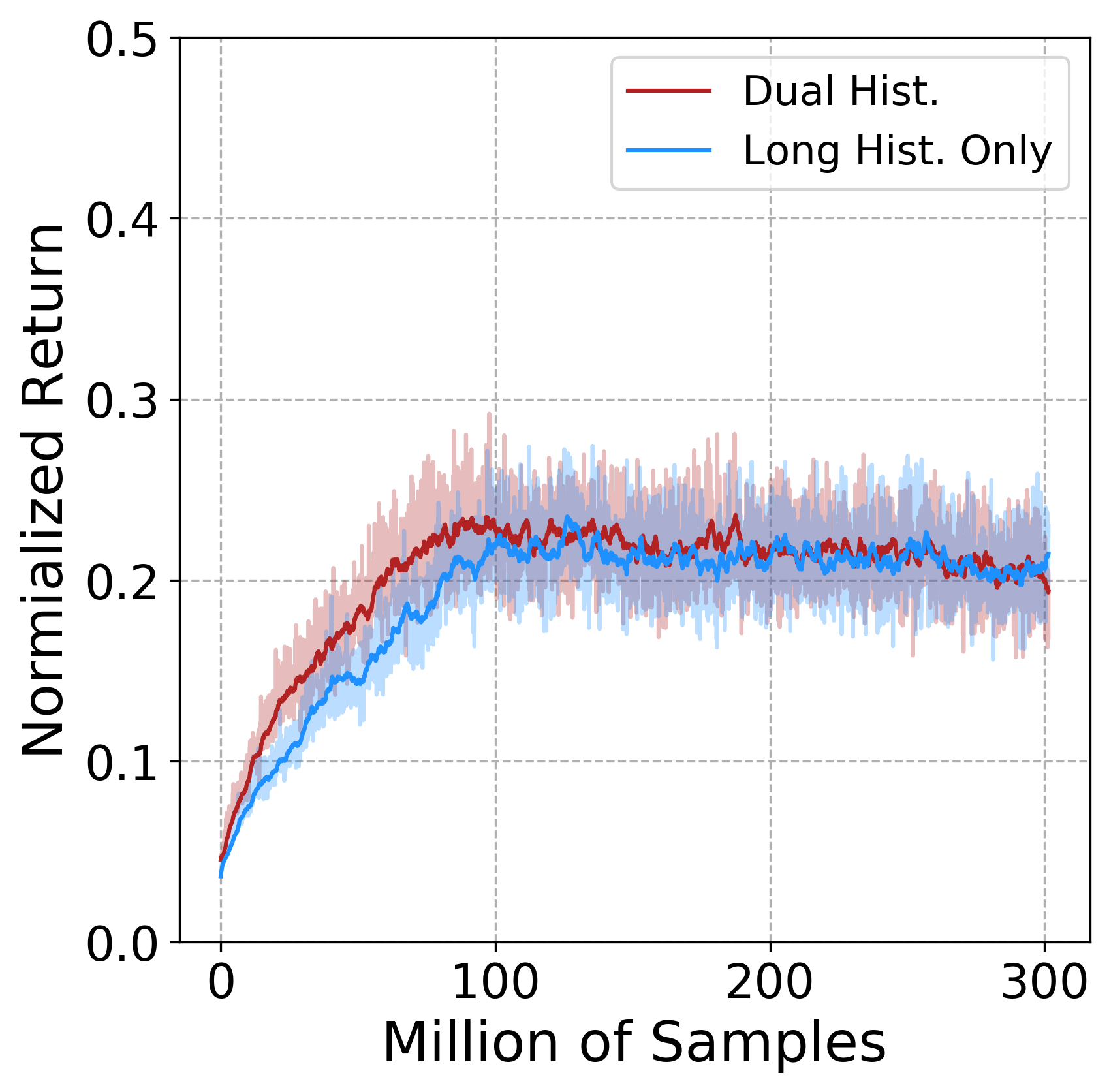}
  \caption{LSTM with dual history approach (Dual Hist.) and the LSTM only (Long Hist. Only).}
  \label{subfig:lstm_compare}
\end{subfigure}
\caption{Learning performance using different neural network architectures to encode the robot's I/O history when training a single-task walking policy with dynamics randomization (walking forward at a fixed speed, after finishing single-stage training). For both Long Hist. Only methods, we still provide explicit immediate state feedback alongside the temporal encoder. As shown in Fig.~\ref{subfig:tcn_compare}, using the proposed dual-history approach by providing an explicit short I/O history alongside the TCN encoder, the learning performance is much better than the TCN only. The TCN encodes 2-second robot I/O history and has 3 layers with filter sizes of [34,34,34], a kernel size of 5, a dilation base of 2, and a stride size of 2, with ReLU activation, as suggested in~\cite{lee2020learning}. Fig.~\ref{subfig:lstm_compare} shows that the dual-history approach will not help with the LSTM-based policy. The LSTM encoder has 1 layer of 128 units. However, both TCN with dual-history approach and Long Hist. Only perform better than the LSTM-based policy while using the hyperparameters tuned for LSTM. It suggests that LSTM may only learn to leverage a recent short history and converge to a more suboptimal policy.}
\label{fig:dual_hist_compare}
\end{figure}

\begin{figure}[ht]
\centering
\includegraphics[width=0.75\linewidth]{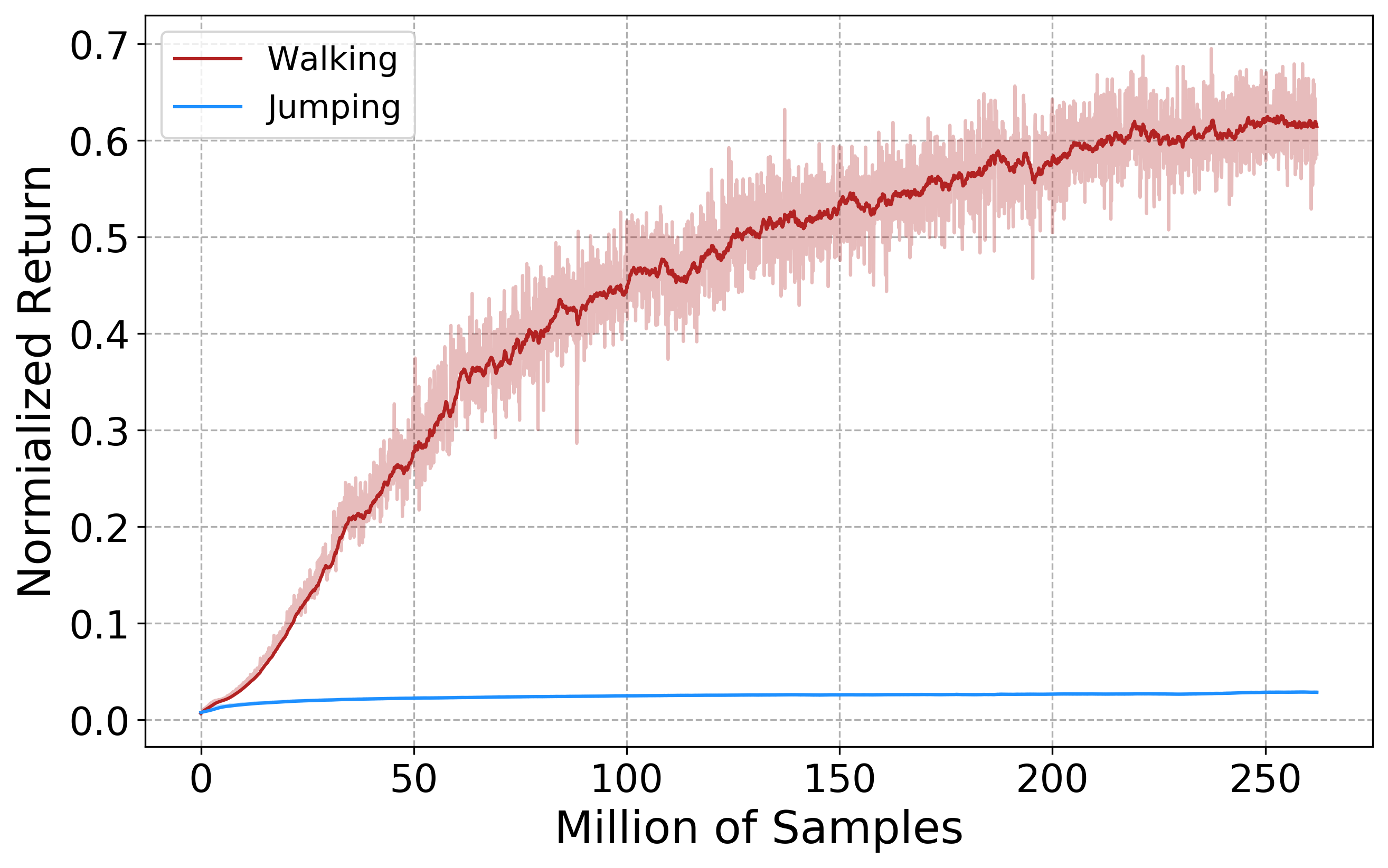}
\caption{Learning performance for training walking (red curve) and jumping (blue curve) skills \emph{from scratch} using the same neural network including the LSTM architecture with the same hyperparamters used in~\cite{siekmann2020learning}. It shows that while the LSTM can learn the walking skill, it may struggle to learn highly dynamic locomotion skills like jumping (shown as a flat curve). This is not to suggest that LSTM cannot learn aperiodic jumping skills, but rather to highlight its sensitivity to hyperparameter tuning across different MDPs (different locomotion skills). This contrasts with the non-recurrent-based policy like the 1D CNN used in this work, which employs a unified set of hyperparameters for all different skills due to the ease of training.}
\label{fig:jump_vs_walk_lstm}
\end{figure}

\begin{figure*}[htp]
\centering
\begin{subfigure}{0.66\linewidth}
  \centering
  \includegraphics[width=\linewidth]{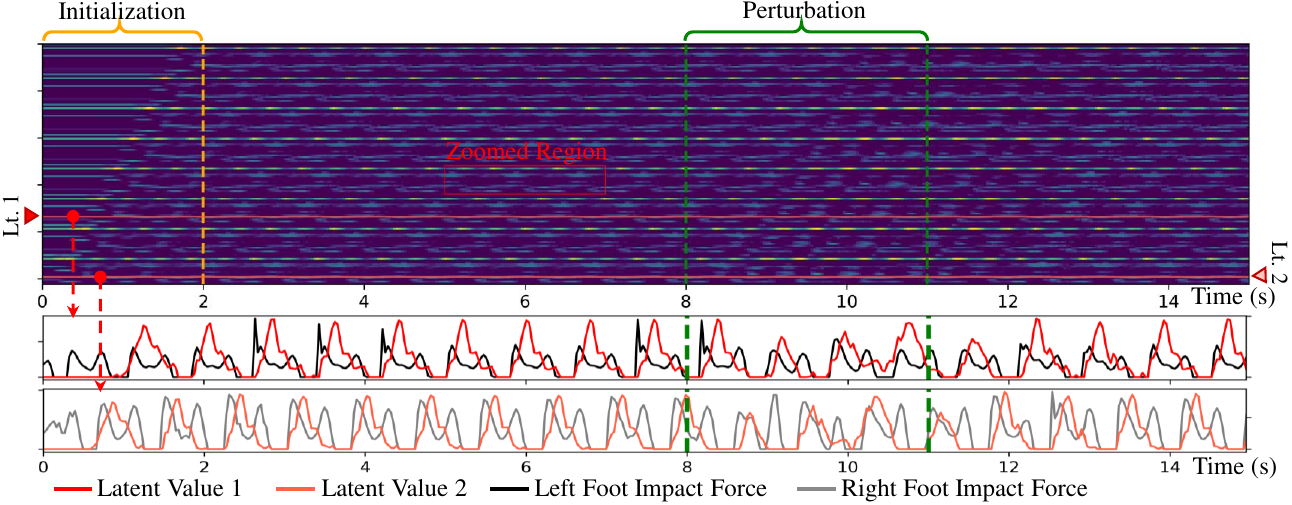}
  \caption{Recorded latent representation after long-term I/O history encoder during walking. The figure below compares two selected dimensions (marked as red lines) with recorded impact forces on each of the robot's feet.}
  \label{subfig:latent_walk_force}
\end{subfigure}
\begin{subfigure}{0.33\linewidth}
  \centering
  \includegraphics[width=\linewidth]{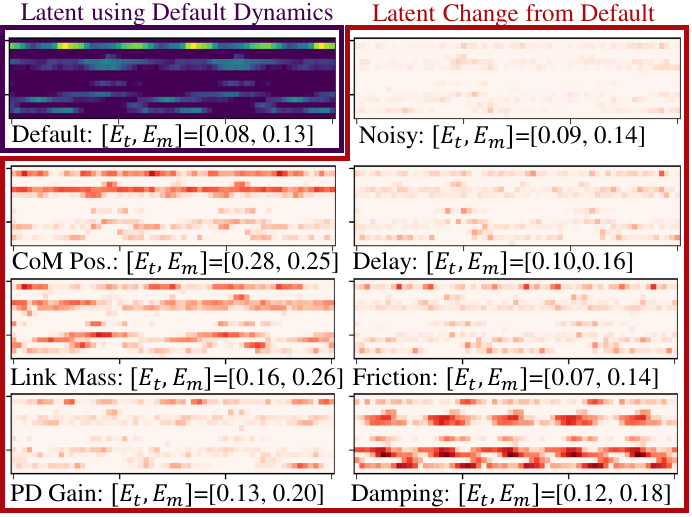}
  \caption{The blue plot shows the robot's latent representation with default dynamics parameters during walking. The red plots indicate changes in the same region under different dynamics.}
  \label{subfig:latent_walk_zoomed}
\end{subfigure}
\caption{Adaptivity test on the obtained walking policy in simulation (MuJoCo). The robot is commanded to walk forward at 0.6 m/s with no lateral or turning movement at normal height (0.95 m). Fig.~\ref{subfig:latent_walk_force} records the latent embedding after the policy's long I/O history encoder over 15 seconds. An external backward perturbation force of 30 N is applied on the robot base from 8 to 11 seconds. We can observe the changes in the latent embedding with the existence of the perturbations. The image below shows a strong correlationship between the two selected latent dimensions with the robot's impact force or contact event. Fig.~\ref{subfig:latent_walk_zoomed} shows the change of the latent embedding (the same zoomed region marked as the red bounding box in Fig.~\ref{subfig:latent_walk_force}) with respect to the change of different dynamics parameters during the same walking task. These ablation studies are the same as running conducted in Fig.~\ref{subfig:latent_run_zoomed}. The control performance metrics, tracking error $E_t$ and motion tracking error $E_m$, show small changes with a large change in the dynamics parameters.}\label{fig:latent_vis_walking}
\end{figure*}

\begin{figure*}[htp]
\centering
\begin{subfigure}{0.49\linewidth}
  \centering
  \includegraphics[width=\linewidth]{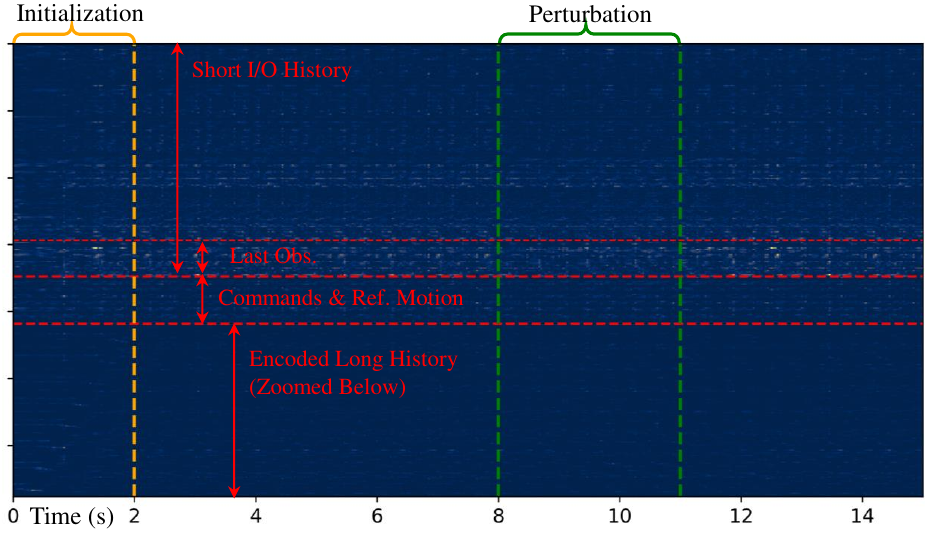}
  \caption{Running}
  \label{subfig:run_saliency_all}
\end{subfigure}
\begin{subfigure}{0.49\linewidth}
  \centering
  \includegraphics[width=\linewidth]{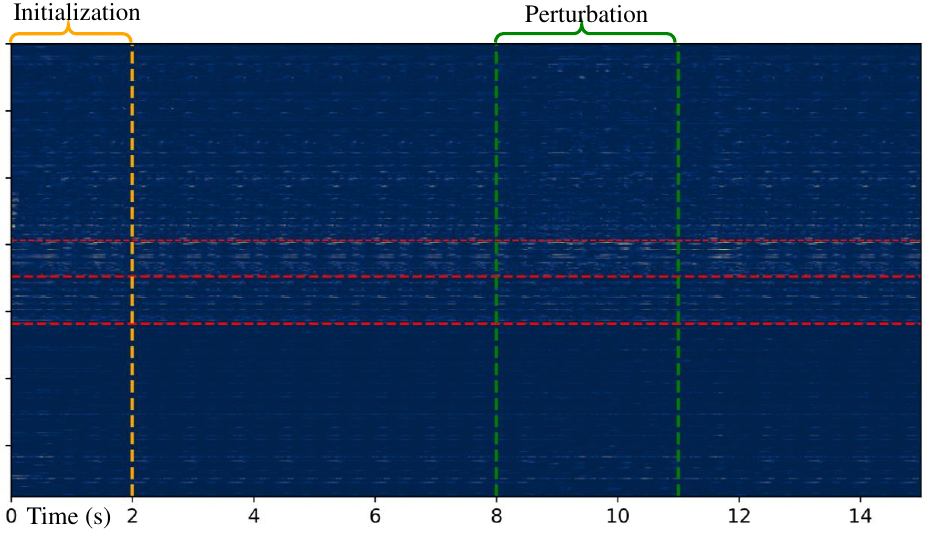}
  \caption{Walking}
  \label{subfig:walk_saliency_all}
\end{subfigure}
\begin{subfigure}{0.485\linewidth}
  \centering
  \includegraphics[width=\linewidth]{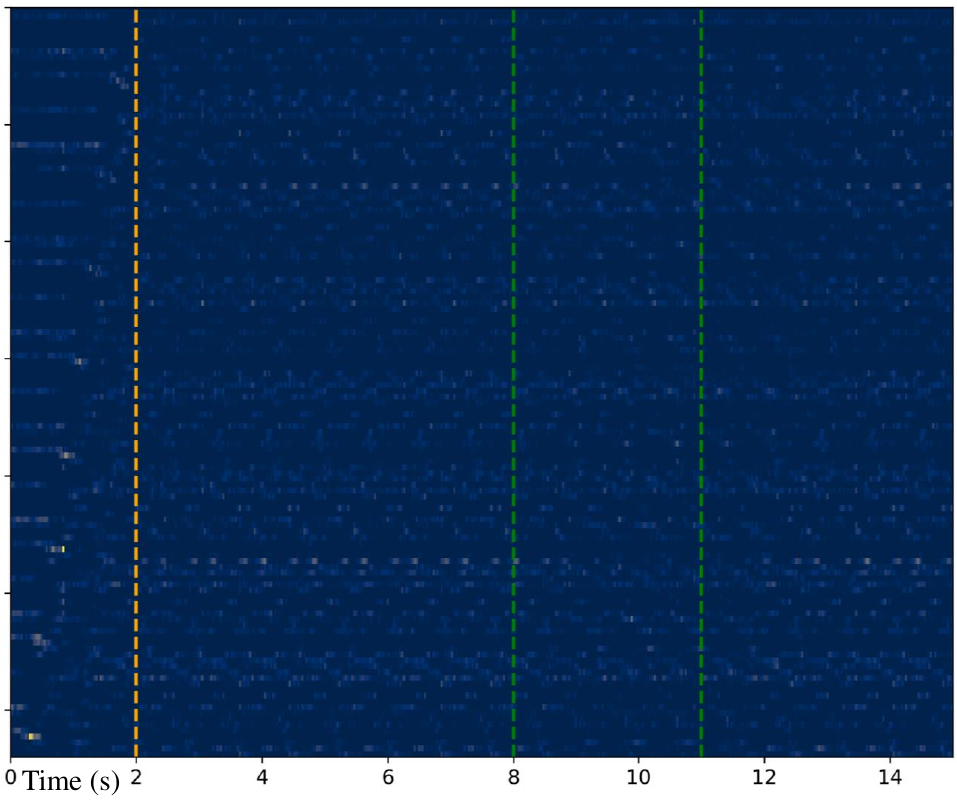}
  \caption{Zoomed Region (Encoded Long History) of Running}
  \label{subfig:run_saliency}
\end{subfigure}
\begin{subfigure}{0.49\linewidth}
  \centering
  \includegraphics[width=\linewidth]{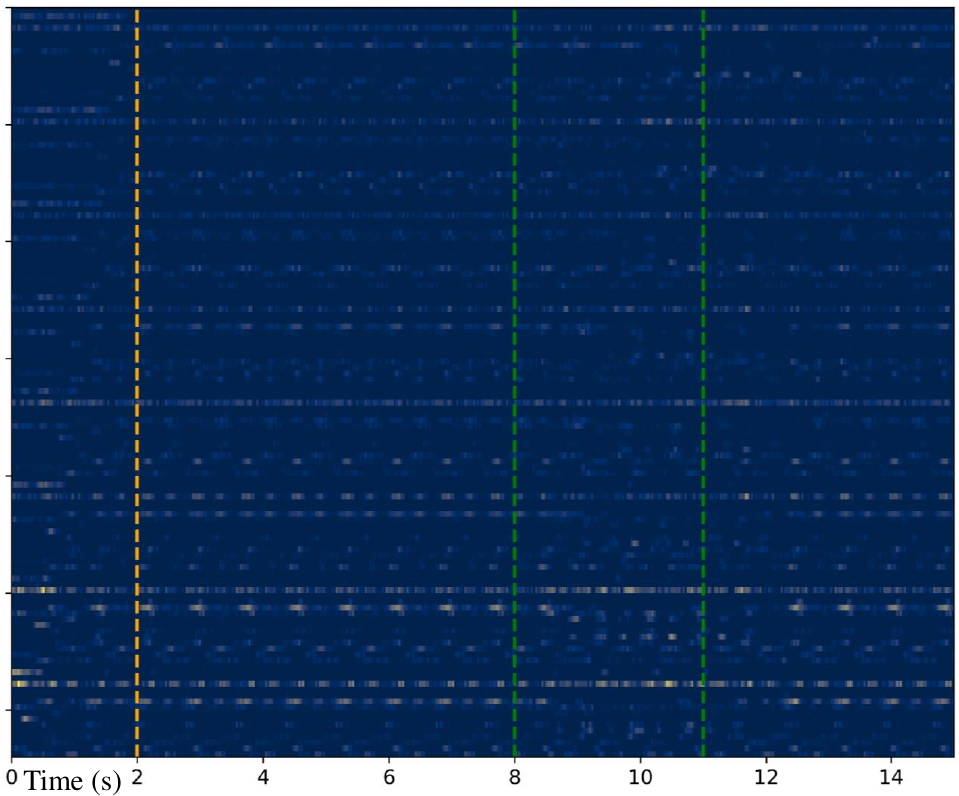}
  \caption{Zoomed Region (Encoded Long History) of Walking}
  \label{subfig:walk_saliency}
\end{subfigure}
\caption{The saliency map of the MLP base with respect to the entire input while the robot is (a) running or (b) walking with the existence of a perturbation from 8 to 11 seconds in a 15-second test. The zoomed regions of the saliency map w.r.t. latent embedding from the long history encoder are shown in detail in (c) for running and (d) for walking. These are the same tests reported in Fig.~\ref{fig:latent_vis} and Fig.~\ref{fig:latent_vis_walking}. The pixels with a lighter color represent the MLP base focus more on this value (higher saliency). These maps show that the MLP base, which produces the policy action, focuses more on the short I/O history, especially the latest observation, highlighting the importance of short I/O history in the input. Moreover, if we take a close look at the encoded long I/O history (c,d), the robot can also focus on different regions of the latent embedding to deal with the change in environments and therefore adjust its output. Therefore, it indicates that the inclusion of long I/O history is also beneficial, which also provides insights on the benchmark conducted in Fig.~\ref{fig:dynrand_three}.}\label{fig:saliency_map}
\end{figure*}

We conducted an ablation study to investigate the effects of varying history lengths on a non-recurrent policy that explicitly requires a specified history length. Using bipedal running training as an example, Fig.~\ref{fig:hist_compare} illustrates the learning performance of training a single-task running policy (which has completed the first-stage training) with extensive dynamics parameters and varying lengths of robot I/O history.

As suggested by Fig.~\ref{fig:hist_compare}, increasing the history length, such as from 1 second to 2 seconds, or from 2 seconds to 3 seconds, consistently enhances learning performance. This improvement occurs because a longer I/O history provides more information on the robot's dynamics parameters, aiding in better state estimation. However, continually extending the history length may not always be beneficial, as it can introduce redundant information that the robot must filter out. This is evidenced by a drop in learning performance when history length increases from 3 seconds to 4 seconds, as shown in Fig.~\ref{fig:hist_compare}. 
In bipedal locomotion control, where remembering events that happened further earlier is less crucial, a longer history could lead more training samples for the robot to distill useful information. 
Therefore, we recommend using a history length that spans a relatively short timespan, such as the 2 seconds used in this work, for more stable training and relatively good performance in learning bipedal locomotion control. For readers interested in adopting this method for their robots, finding the optimal history length could be beneficial for specific tasks, but we suggest starting with a 2-second history, which has been tested extensively in this work.

}

\subsection{Comparison of Use of Different Temporal Encoders}\label{appendix:lstm_tcn}

To explore the effects of different neural network structures encoding the robot's I/O long history, we conducted an ablation study and benchmarked policies based on TCN and LSTM. TCN, a non-recurrent structure that encodes temporal information, is used in~\cite{lee2020learning}, while LSTM, a recurrent structure, is utilized in other bipedal locomotion control works like~\cite{siekmann2020learning}. To eliminate potential confounding factors such as poor implementation, we adopted the training algorithms (PPO and recurrent PPO) and training hyperparameters open-sourced from~\cite{siekmann2020learning}, which underpin many subsequent studies such as~\cite{siekmann2021blind, siekmann2021sim, crowley2023optimizing}. The MDP design (training environment) employed the walking and jumping scenarios developed in this work.

As shown in Fig.~\ref{subfig:tcn_compare}, providing the base policy with a short 4-timestep I/O history alongside the TCN encoder significantly improves learning performance compared to using TCN alone (while still providing the base policy with immediate last state feedback, which is the form used in \cite{lee2020learning}). This suggests that the dual-history approach can consistently enhance learning performance for non-recurrent structures that encode an explicit history length. Note that in this comparison, we use the same TCN implementation as detailed in Table S5 in~\cite{lee2020learning}.

Conversely, as depicted in Fig.~\ref{subfig:lstm_compare}, the LSTM-based policy shows no significant improvement whether a short history is provided or not. Additionally, it tends to converge to a lower return plateau compared to the TCN-based policy, even though the TCN used hyperparameters tuned for LSTM. This might indicate that LSTM encoders may learn to focus more on recent short history and pay less attention to older history. Although LSTM hidden states are only reset at episode ends, they may quickly converge to a suboptimal policy focusing primarily on recent history, explaining why providing additional short history does not enhance performance and why LSTM underperforms compared to TCN that was explicitly trained to utilize a long history. This observation aligns with findings reported in~\cite{lee2020learning} and \cite{singh2023learning}.

Moreover, extending the LSTM-based policy to learn jumping skills resulted in training failure, even using the same architecture and hyperparameters from~\cite{siekmann2020learning} that reproduced the same walking skill developed in this work, as illustrated in Fig.~\ref{fig:jump_vs_walk_lstm}. Please note that we are not asserting that it is impossible for LSTM to learn aperiodic skills like jumping, but it is difficult without carefully tuning hyperparameters during training. 
In contrast, our method provides a straightforward and general approach for learning various skills, using unified hyperparameters across all tasks. After finalizing the training for walking, we did not adjust any hyperparameters for jumping or running.

In summary, the dual history approach consistently helps learning performance for policies recording explicit robot history (non-recurrent policies like TCN and 1D CNN), but does not improve recurrent policies like LSTM. Moreover, recurrent policies are sensitive to hyperparameter tuning across different MDPs (such as locomotion control) and may more readily converge to suboptimal policies. Therefore, dual-history-based non-recurrent policy could be more favorable for bipedal locomotion control cases.

\subsection{Latent Visualization of Walking Policy}\label{appendix:walking_latent}
The results of adaptivity test on the obtained walking policy by the proposed method are presented in Fig.~\ref{fig:latent_vis_walking}. 
The adaptivity test is similar to the ones conducted on the running policy discussed in Sec.~\ref{subsec:adaptivity}. 
It evaluates the change of latent embedding after the long I/O history encoder when the robot is commanded to walk at a constant forward speed of 0.6 m/s. 
The findings are consistent with the ones observed from the tests on other skills reported in Sec.~\ref{subsec:adaptivity}: we show that the latent embedding is able to capture the time-variant changes like external perturbation and contact events (Fig.~\ref{subfig:latent_walk_force}) and time-invariant dynamics shifts (Fig.~\ref{subfig:latent_walk_zoomed}). 
During the ablation study on the dynamics shifts, the latent embedding differs when dynamics parameters change, but the resulting control performance shows little change. This indicates the adaptivity of the walking policy.

\subsection{Saliency Map}\label{appendix:saliency}
To further understand how the policy adjusts its action based on the environment changes, we visualize the saliency map in Fig.~\ref{fig:saliency_map} as used in~\cite{lee2020learning}. The saliency map reflects how important each input dimension is to the neural network output (\textit{i.e.}, where does the policy focus).

As shown in the saliency map of the MLP base, which produces the final policy action on the entire input, the robot focuses more on the short I/O history, particularly the most recent observation, highlighting the importance of the short I/O history, during both running (Fig.~\ref{subfig:run_saliency_all}) and walking (Fig.~\ref{subfig:walk_saliency_all}).

We can take a close look at the zoomed regions (encoded long history) of the entire saliency map in the same test of the running (Fig.~\ref{subfig:latent_run_force}) and walking (Fig.~\ref{subfig:latent_walk_force}). The saliency maps with respect to the encoded long history have different focuses on different parts of the embeddings from the long-history encoder with the existence of the external perturbation during running (Fig.~\ref{subfig:run_saliency}) or walking (Fig.~\ref{subfig:walk_saliency}). 
This further suggests that the proposed RL policy could utilize the long history information to adjust the action for different control scenarios, \textit{i.e.}, the long I/O history is useful.

From the extensive ablation study in Fig.~\ref{fig:dynrand_three}, we observed differences in learning performance and sim-to-real transfer depending on the use of history.
It will result in degradation by either removing the short I/O history, even keeping the most recent observation that drew the most attention, which is Long History Only in Fig.~\ref{fig:benchmark}d, or removing the long I/O history (Short History Only, Fig.~\ref{fig:benchmark}e). These saliency maps on the entire input space provide insights into the reason behind this.

\subsection{Errors of Estimator in High-Speed Running}\label{appendix:estimaion_error}

\begin{figure}[!]
\centering
\includegraphics[width=0.8\linewidth]{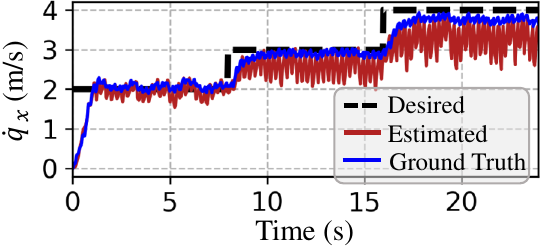}
\caption{The large estimation error using the robot onboard velocity estimator (based on EKF) during high-speed running \emph{in simulation}. In this figure, the robot is controlled by the proposed running policy to track variable commands (black dashed line) in simulation, the estimated sagittal velocity $\dot{q}_x$ is recorded as the red line while the robot's actual running speed is recorded as the blue line. \emph{The ground-truth speed is obtained from the simulator}. Although showing accurate results under slow speed (such as 2 m/s), the estimated velocity shows a significant error in the high-speed region (above 3 m/s) compared to the ground-truth speed. The robot's actual speed tends to be the upper envelope of the estimated speed. In real-world experiments, we only have access to the state estimator whose result we can report, such as the running speed tracking results in Fig.~\ref{subfig:run_400_log}. This comparison shows that the robot's actual running speed in the real world is faster than the reported estimated value and closer to the command.}\label{fig:state_est_err}
\end{figure}

In this section, we present the estimation errors of the state estimator (EKF) we used in the high-speed running. In Fig.~\ref{fig:state_est_err}, we compare the robot ground-truth sagittal velocity $\dot{q}_x$ during running obtained from simulation and the corresponding estimated velocity from the estimator. During high-speed running (larger than 3 m/s), we observe a large estimation error from the estimator to which we only have access during real-world experiments. This suggests that the robot's actual running speed in the real world is faster than the recorded estimated speed, and may be the upper envelope of the estimation. This large estimation error is further supported when comparing the estimated speed and the robot's actual average speed (traveled distance/elapsed time). These suggest a smaller actual tracking error during high-speed running than the experiment logs suggested. This also suggests the necessity of including this estimator during training rollout to let the robot train with the observation data produced by this inaccurate estimator. But still, the robot tends to have a noticeable tracking error during running in the high-speed region due to the sim-to-real gap and robot hardware limitation. We note that developing a reliable state estimator for dynamic bipedal locomotion skills using RL could be an interesting future work.

\begin{figure}[!htp]
\centering
\begin{subfigure}{0.455\linewidth}
  \centering
  \includegraphics[width=\linewidth]{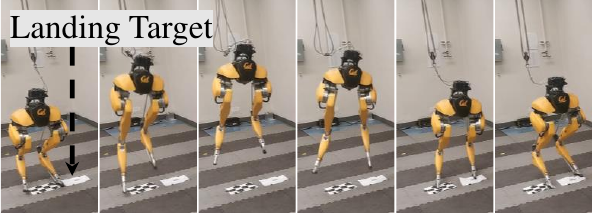}
  \caption{$(q^d_x, q^d_y, q^d_\phi)$ = (0m, 0.3m, 0$^\circ$)}
  \label{subfig:jump_flat_lateral03_appx}
\end{subfigure}
\begin{subfigure}{0.53\linewidth}
  \centering
  \includegraphics[width=\linewidth]{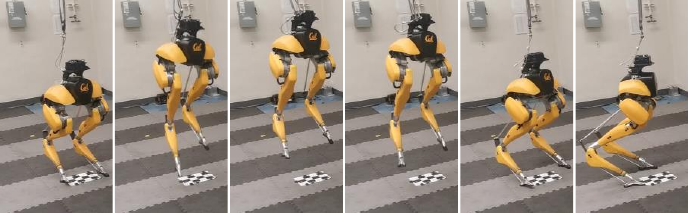}
  \caption{$(q^d_x, q^d_y, q^d_\phi)$ = (0m, 0m, 60$^\circ$)}
  \label{subfig:jump_flat_turn60_appx}
\end{subfigure}
\begin{subfigure}{\linewidth}
  \centering
  \includegraphics[width=\linewidth]{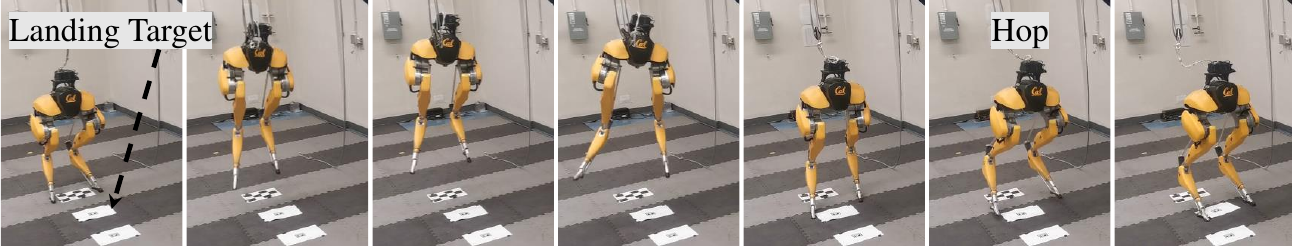}
  \caption{$(q^d_x, q^d_y, q^d_\phi)$ = (0.5m, 0m, 0$^\circ$)}
  \label{subfig:jump_flat_forward05_appx}
\end{subfigure}
\begin{subfigure}{\linewidth}
  \centering
  \includegraphics[width=\linewidth]{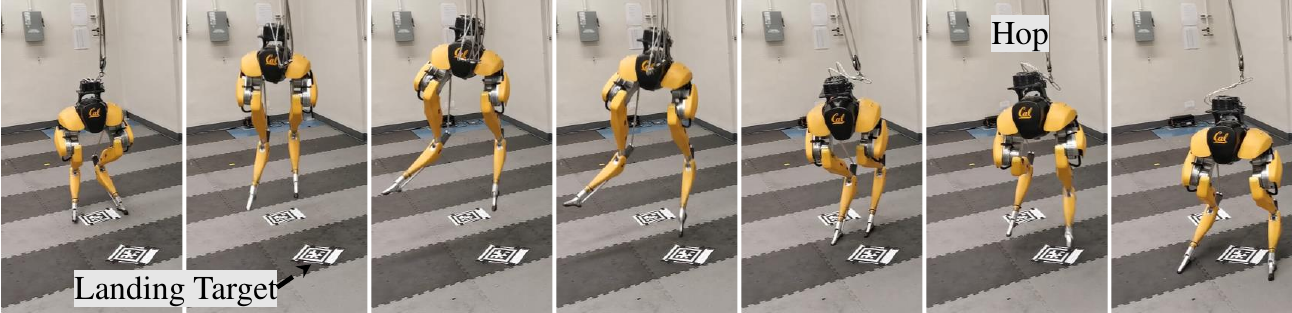}
  \caption{$(q^d_x, q^d_y, q^d_\phi)$ = (0.7m, 0m, -45$^\circ$)}
  \label{subfig:jump_flat_foward07turn45_appx}
\end{subfigure}
\caption{More bipedal jumps using the same flat-ground jumping policy. The paper tag on the ground indicates the jumping target.}\label{fig:more_jumps}
\end{figure}

\subsection{Diverse Bipedal Jumps}~\label{appendix:other_jumps}
Using the same flat-ground jumping policy that realized various bipedal jumps demonstrated in Sec.~\ref{subsec:exp_jumping}, we further evaluate the capacity to jump to other different targets in the real world, as showcased in Fig.~\ref{fig:more_jumps}.

\end{document}